\documentclass[final,3p,times,authoryear]{elsarticle}

\usepackage{amssymb}
\usepackage{amsmath}
\usepackage{framed,multirow}
\usepackage{latexsym}
\usepackage{multirow}
\usepackage{svg}
\usepackage{xcolor}
\usepackage{subfiles}
\usepackage{multicol}
\usepackage{color, colortbl}
\usepackage{graphicx,wrapfig,lipsum}
\usepackage{subcaption}
\usepackage{url}
\usepackage{hyperref}
\usepackage{makecell}
\usepackage{booktabs}
\usepackage{array}
\usepackage{multirow}
\usepackage{xcolor}
\usepackage{siunitx}

\journal{Medical Image Analysis}

\begin{document}

\begin{frontmatter}



\title{RACR-MIL: Rank-aware contextual reasoning for weakly supervised grading of squamous cell carcinoma using whole slide images}%


\author[cor1]{Anirudh Choudhary}
\author[cor1]{Mosbah Aouad}
\author[cor1]{Krishnakant Saboo}
\author[cor2]{Angelina Hwang}
\author[cor2]{Jacob Kechter}
\author[cor2]{Blake Bordeaux}
\author[cor2]{Puneet Bhullar}
\author[cor2]{David DiCaudo}
\author[cor2]{Steven Nelson}
\author[cor3]{Nneka Comfere}
\author[cor3]{Emma Johnson}
\author[cor4]{Olayemi Sokumbi}
\author[cor2]{Jason Sluzevich}
\author[cor2]{Leah Swanson}
\author[cor3]{Dennis Murphree}
\author[cor2]{Aaron Mangold}
\author[cor1]{Ravishankar Iyer}

\cortext[cor1]{Corresponding author. \textit{E-mail address}: \href{ac67@illinois.edu}{ac67@illinois.edu}, \textit{Phone}: +1-6462514328 (Anirudh Choudhary)}

\affiliation[cor1]{organization={University of Illinois Urbana-Champaign},
            city={Champaign},
            postcode={61820},
            state={IL},
            country={USA}}

\affiliation[cor2]{organization={Mayo Clinic},
            addressline={13400 E. Shea Blvd. Scottsdale},
            city={Scottsdale},
            postcode={85259},
            state={AZ},
            country={USA}}

\affiliation[cor3]{organization={Mayo Clinic},
            addressline={200 1st St SW, Rochester},
            city={Rochester},
            postcode={55905},
            state={MN},
            country={USA}}

\affiliation[cor4]{organization={Mayo Clinic},
            addressline={4500 San Pablo Rd S},
            city={Jacksonville},
            postcode={32224},
            state={FL},
            country={USA}}

\begin{abstract}
Squamous cell carcinoma (SCC) is one of the most common cancer subtype, with an increasing incidence and a significant impact on cancer-related mortality. SCC grading using whole slide images is inherently challenging due to the lack of a standardized grading protocol and substantial tissue heterogeneity. We propose RACR-MIL, a weakly-supervised SCC grading approach that achieves robust generalization across multiple anatomies (skin, head \& neck, lung). RACR-MIL is an attention-based multiple-instance learning framework that introduces two key innovations for learning grade-specific contextual representations: (1) a hybrid WSI graph that captures both local tissue context and non-local phenotypic dependencies between tumor regions, and (2) rank-ordering constraints on the attention mechanism that encourages consistent prioritization of higher-grade tumor regions and improves region-level grade confidence, aligning with pathologists’ diagnostic process. Our model achieves state-of-the-art performance across multiple SCC datasets, achieving 3-9\% improvements over existing methods and up to 10\% improvement in tumor localization.  In a pilot study, pathologists reported that RACR-MIL improved grading efficiency in 60\% of cases, underscoring its potential as a clinically viable cancer diagnosis and grading assistant.
\end{abstract}




\end{frontmatter}



\section{Introduction}
Squamous cell carcinoma (SCC) is the most pervasive cancer subtype and remains a leading cause of cancer-related mortality (\cite{tokez2020incidence}). SCC arises across multiple anatomical sites, most notably the skin, head and neck, and lung, with cutaneous SCC alone accounting for approximately one million cases annually (\cite{tokez2020incidence, rogers2015incidence}). SCC prognosis and clinical management critically rely on accurate histological tumor grading from whole slide images (WSI). Despite the prevalence of SCC, accurate WSI-based grading remains challenging due to significant intratumoral heterogeneity, subtle morphological differences between adjacent grades, and the lack of a reliable and standardized grading protocol (\cite{diede2024grading}). In clinical practice, pathologists typically assess SCC grade by integrating cellular differentiation and localized tumor morphology with broader tissue context, such as tumor growth pattern, keratinization, and characteristics of the surrounding tumor microenvironment.

\begin{figure*}[!th]
    \centering
    \includegraphics[width=\linewidth]{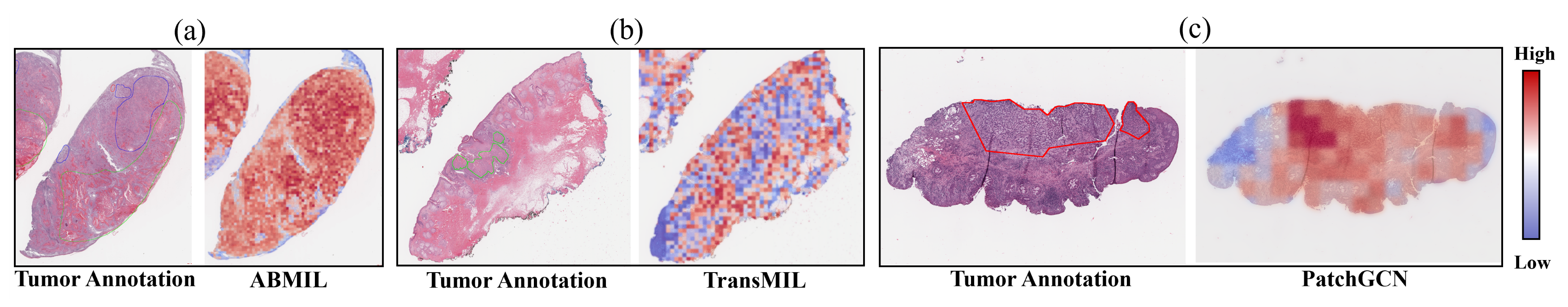}
    \caption {Attention heatmaps for existing MIL methods with corresponding tumor annotations (\textcolor{red}{Red}: Poor, \textcolor{blue}{Blue}: Moderate, \textcolor{green}{Green}: Well): (a) ABMIL without ranking constraint fails to consistently rank tumor regions assigning similar importance to Moderately-differentiated and Well-differentiated regions, (B) Transformer-based approach (TransMIL) is susceptible to tissue noise leading to false positives and misses the relevant tumor region; (c) Graph convolutional network (PatchGCN) suffers from oversmoothing and assigns similar importance to tumor and non-tumor regions, failing to prioritize relevant local information.}
    \label{fig:intro_image_var}
\end{figure*}

 Weakly-supervised learning, which leverages WSI-level labels instead of pixel-level annotations, has emerged as a promising paradigm for automated cancer grading (\cite{dominguez2024systematic}). However, most weakly-supervised methods do not explicitly model the long- and short-range tissue dependencies that clinicians rely on (e.g., keratinization, stromal reaction). Moreover, they lack a mechanism to consistently prioritize the diagnostically critical worse-grade tumor regions. As a result, slides containing varying grades are treated equivalently, which can dilute attention to the most informative regions (Figure \ref{fig:intro_image_var}). In this paper, our goal is to develop a weakly supervised SCC grading approach to classify WSIs into one of four grading classes: normal (no tumor), well-differentiated, moderately-differentiated, and poorly-differentiated. To the best of our knowledge, this work is the first AI-based SCC grading framework designed to achieve robust generalization across multiple anatomical sites (skin, lung, head, and neck). Our framework incorporates a novel dual-graph representation learning architecture to encode clinically relevant tissue context, and handles tumor heterogeneity through a ranking-based formulation, prioritizing diagnostically critical regions. 

We formulate SCC grading using attention-based multiple-instance learning (MIL), which has demonstrated strong performance in prior cancer grading studies (\cite{mun2021yet,silva2021self,otalora2021combining}). However, its application to SCC introduces three key challenges. First, SCC WSIs often contain heterogeneous tumor regions, and pathologists implicitly prioritize more aggressive tumor components while assigning a grade (\cite{prezzano2021concordance}). Standard attention-based MIL imposes no explicit ordering on instance importance and can consequently overemphasize lower-grade regions or less informative regions, limiting grading accuracy and tumor localization (\cite{dominguez2024systematic}). Second, SCC grading depends on tissue context as multiple scales, such as tumor growth patterns and tumor microenvironmental structure (e.g., keratinization and inflammatory response (\cite{diede2024grading}). Existing methods that model patches independently fail to capture these contextual dependencies (\cite{silva2021self, mun2021yet}), whereas transformer- and graph-based approaches (\cite{behzadi2024weakly}) introduce distinct modeling limitations. Transformer-based methods (e.g., TransMIL (\cite{shao2021transmil}) capture long-range context through dense self-attention, but are prone to assigning non-zero importance to biologically irrelevant regions, thereby introducing tissue noise and diluting fine-grained grade-specific morphological differences (Figure \ref{fig:intro_image_var}). Graph-based approaches provide a sparse and efficient alternative, but existing approaches typically model only a subset of relevant relationships. Spatial graphs, such as PatchGCN (\cite{chen2021whole}), have limited ability to capture distant phenotypic dependencies with shallow message passing, while latent graphs (\cite{guan2022node}) fail to model the local tumor-microenvironment context and can oversmooth subtle differences between adjacent grades (Figure \ref{fig:intro_image_var}). Third, SCC training cohorts are often limited in size and imbalanced, with substantially fewer higher-grade cases. This further increases the risk of overfitting in the presence of tumor heterogeneity. To address these challenges, we propose a rank-aware contextual reasoning framework, RACR-MIL, that explicitly models complementary local and non-local tissue relationships while emulating the ordinal grading protocol used by pathologists. 

Our model, RACR-MIL, introduces two key innovations. First, we propose a rank-aware attention mechanism with two distinct ordering constraints: (i) an \textit{inter-grade} constraint, which prioritizes patches associated with more severe grades (e.g., poorly differentiated) than those associated with lower grades, and (ii) an \textit{intra-grade} constraint, which prioritizes confident patches with higher grade-specific probability within the same grade. Rank ordering enables our model to consistently assign higher importance to the most diagnostically relevant tumor regions, thereby aligning WSI feature aggregation and prediction with the overall label and improving region-level tumor localization.

Second, we represent each WSI using a hybrid dual-graph that captures complementary short- and long-range tissue dependencies relevant to SCC grading. The dual graph models distinct structural priors: (i) a \textit{latent graph} that sparsely connects phenotypically similar regions (e.g., scattered tumor nests) based on feature similarity, to model non-local patterns associated with tumor growth; and (ii) a \textit{spatial graph} that links physically adjacent regions to capture local tumor microenvironment context, such as keratinization and immune infiltration. We refine the latent graph using graph diffusion to capture higher-order relationships, and apply transformer-based self-attention to adaptively weight relations (edges) in both graphs. A diversity constraint further encourages the two branches to learn distinct and complementary relational patterns. Overall, the dual-graph design preserves fine-grained morphology while integrating local and global tissue context.

We comprehensively evaluate RACR-MIL across three SCC cohorts spanning skin SCC, head \& neck SCC, and lung SCC, and compare it with state-of-the-art MIL methods. RACR-MIL consistently outperforms existing methods, achieving a 3-9\% improvement in grade classification and 5-20\% higher accuracy for challenging higher-grade tumors. It also reliably localizes diagnostically important tumor regions, achieving up to 10\% higher sensitivity than prior methods. Qualitative analysis of attention heatmaps in skin SCC further demonstrates greater alignment with clinicians' annotations and more consistent prioritization of relevant tumor regions. To summarize, our contributions are as follows:
\begin{enumerate}
    \item We propose RACR-MIL, the first weakly supervised framework for multiclass SCC grading that jointly models local spatial context and non-local phenotypic relationships through a hybrid dual-graph representation. A diversity constraint encourages the two graphs to capture complementary grading-relevant information. 
    \item We introduce a rank-aware attention mechanism with inter-grade and intra-grade constraints that explicitly prioritizes regions exhibiting stronger evidence of higher-grade tumor, mimicking the ordinal nature of SCC grading and improving the localization of diagnostically relevant tumor regions.
    \item We evaluate RACR-MIL across three SCC cohorts—skin, head and neck, and lung—and benchmark it against a broader set of weakly supervised baselines. RACR-MIL achieves state-of-the-art grading performance and improved alignment with pathologists' annotations, both qualitatively and quantitatively. Additionally, we conduct a pilot study with pathologists to assess RACR-MIL's efficacy as a grading assistant, supporting its potential for clinical translation.
    \item  We demonstrate strong generalization and enhanced robustness across diverse SCC cohorts, including greater resilience to class imbalance and reduced performance variability across evaluation folds.
\end{enumerate}
\section{Related Work}
\textbf{Attention-based MIL}: Attention-based MIL (ABMIL) (\cite{ilse2018attention}) assigns learnable importance weights to patch features before aggregating them into a slide-level representation. Subsequent methods improve instance discrimination and aggregation. CLAM (\cite{lu2021data}) incorporates an instance-level clustering objective to enhance separation between high- and low-attention patches, while its multiclass variant (CLAM-MB) uses class-specific attention heads. \cite{li2021dual} propose a dual-stream architecture that jointly optimizes instance- and embedding-level classifiers and uses self-attention to model latent patch dependencies. Despite their effectiveness, conventional ABMIL methods primarily optimize patch aggregation and do not explicitly model tissue relationships or enforce the ordinal importance of tumor regions required for grading.\\
\textbf{Contextual MIL}: Recent work incorporates tissue context by modeling relationships among image patches using graph neural networks or self-attention. Graph-based approaches represent WSI patches as nodes and define edges using spatial proximity or latent feature similarity. PatchGCN (\cite{chen2021whole}) models local tissue structure by connecting spatially adjacent patches, while GCN-MIL (\cite{xiang2023automatic}) constructs proximity-based graphs for prostate cancer grading. Dynamic graph variants such as WiKG (\cite{li2024dynamic}) adapt spatial edge definitions using learned patch embeddings and perform directed message passing. Beyond local relations, graph methods account for tissue heterogeneity and inter-tissue interactions for survival modeling. ProtoSurv (\cite{wu2024leveraging}) further incorporates tissue-category information and clustering into spatial graphs to distinguish heterogeneous tissue structures. However, a key limitation of purely spatial graphs is their restricted receptive field, limiting their ability to model distant phenotypic relationships. Latent graphs address this limitation by connecting morphologically similar regions. \cite{behzadi2024weakly} use attention-based MIL to identify salient patches followed by graph convolution on latent graph. Node-GCN (\cite{guan2022node}) constructs latent graphs over patch clusters for cross-slide feature alignment using representative nodes. However, cluster-level representations might be suboptimal for grading, where small, spatially localized tumor regions and subtle differences between adjacent grades can determine the final diagnosis. Clustering may dilute these regions, and conventional graph convolution can blur fine-grained morphological distinctions. HEAT (\cite{chan2023histopathology}) extends latent graph modeling through heterogeneous tissue-type nodes with semantic edge attributes and self-attention based message passing. However, latent graph-based methods suboptimally utilize the local spatial neighbourhood context around tumor. 

Transformer-based methods model broader patch relations using self-attention. Long-MIL (\cite{li2023longmil}) introduces locality constraints to improve attention efficiency for large WSIs, while TransMIL (\cite{shao2021transmil}) incorporates spatial priors into self-attention using convolutional positional encoding. Hybrid architectures combine graph-based local reasoning with transformer attention to model multiscale context. Graph Transformer (\cite{zheng2022graph}) applies self-attention over contextual patch features derived using spatial graph convolution, whereas CAMIL (\cite{fourkioti2023camil}) combines global self-attention with local neighbor attention guided by feature similarity. Dense self-attention can capture long-range dependencies but may assign non-negligible weights to biologically irrelevant regions, increasing sensitivity to tissue noise and diluting subtle grade-specific morphology. Moreover, these approaches often rely on approximations such as Nyström attention (\cite{xiong2021nystromformer}), which can degrade performance. Consistent with this, our experiments suggest that dense self-attention mechanism underperforms on fine-grained tumor grading. Existing contextual methods thus tend to emphasize either local spatial context, non-local phenotypic similarity, or dense global interactions. RACR-MIL instead constructs WSI-specific spatial and latent graphs and encourages them to model complementary relations via diversity regularization and graph-specific attention. The spatial graph preserves local tumor-microenvironment context, while the enhanced latent graph connects phenotypically similar patches beyond one-hop neighbors without collapsing them into cluster representations. 

\textbf{Weakly-supervised grading}: Most weakly-supervised grading methods have focused on non-SCC cancers, particularly prostate cancer. These approaches generally use either weak tissue-level annotations along with slide-level grades (\cite{kanavati2020weakly, otalora2021combining}), or rely solely on WSI-level Gleason or ISUP grades (\cite{mun2021yet,silva2021self}). \cite{otalora2021combining} combine limited patch-level supervision with large-scale weak labels using transfer learning-based MIL. \cite{silva2021self} use max-pooling MIL to infer pseudo-labels from WSI-level annotations, and subsequently train a patch-level classifier. \cite{mun2021yet} adopt a two-stage non-contextual approach that first distinguishes cancerous from benign patches, followed by attention-based grading using pre-extracted features. \cite{zhang2021joint} further incorporate patches sampled at multiple magnifications into non-contextual attention-based MIL. Recent studies have explored graph-based contextual modeling for grading using spatial graphs (\cite{xiang2023automatic}) or latent graph on selected patches based solely on predicted tumor content (\cite{behzadi2024weakly}). \cite{myronenko2021accounting} use transformer-based self-attention to model patch dependencies and introduce a pseudo-label-based patch-level loss. However, comparative studies (\cite{dominguez2024systematic}) report only marginal improvements of contextual approaches over non-contextual methods such as CLAM-SB, highlighting that effective contextual modeling for grading remains challenging.  
Weakly-supervised methods have also been investigated for skin cancers such as basal cell carcinoma (BCC). \cite{yacob2023weakly} combine spatial graphs with transformer self-attention to capture both local and global context, but show limited performance in differentiating higher-grade lesions. Similarly, \cite{GEIJS2024103063} proposed StreamingCLAM for BCC risk stratification, although higher-grade categories were merged into a binary risk classification problem, reducing clinical granularity. In comparison, weakly supervised SCC grading remains underexplored and, to our knowledge, has not been addressed in a multiclass setting. \\

\section{Methodology}
Each WSI is represented as a bag $b$ of patches $X_b = \{x_{n}\}_{n=1}^N$. $N$ denotes the number of non-overlapping patches in a WSI, and $b \in \{1,2...B\}$, where $B$ is the number of training samples. The patch-level labels $y_n$ are unknown, and we have access only to the bag label $Y_b \in $ \{normal, well, moderate, poor\}. 

\subsection{Tissue Feature Encoding} \label{feature_extraction}
Tissue feature extraction consists of local feature extraction using self-supervised learning followed by contextual feature extraction using an attention-based dual graph convolution network.

\subsubsection{Local feature extraction} 
We leverage self-supervised learning (SSL) to pre-train the patch feature extractor using unlabeled patches extracted from WSIs. To capture fine-grained pathological features (e.g., nuclei details, cell distribution, tumor microenvironment), we extract $448 \times 448$ sized non-overlapping patches at $20X$ magnification. The patch size and magnification were chosen to preserve high-resolution morphological details while ensure grading-relevant histological context within a single patch. Existing studies (\cite{alfasly2024rotation}, \cite{lu2024visual}) have shown that using a larger patch-level context can capture semantic tissue cues better and improve performance on morphology-driven tasks. In our setting, a larger patch also allows use of higher batch size during training leading to stable optimization under high class imbalance and limited data size. We use Nest-S (\cite{zhang2022nested}), a hierarchical transformer architecture, for efficient pretraining with larger-sized 448-dimensional images. We pretrain the feature extractor using DINO, a knowledge distillation-based SSL framework (\cite{caron2021emerging}). 

After pretraining, the transformer network is used as an offline feature extractor to derive $384$-dimensional feature $f_n \in \mathcal{R}^d$ for each patch $x_n$, leading to a WSI representation of $\{f_n\}_{n=1}^N$. Since the pre-trained features $f_n$ are agnostic to the downstream task, we project them into a lower-dimensional space using a multi-layer perceptron (MLP) with nonlinear activation to get local patch features $h_n^0 \in \mathcal{R}^d$. The projection reduces noise and emphasizes grading-specific features, allowing the model to refine and adapt the pre-trained features. We jointly train the MLP along with the graph neural network, and the rank-aware grade classifier. Thus, the resulting local patch feature $h_n^0$ captures grading-specific information and de-emphasizes information irrelevant to the downstream task. 

\begin{figure}[!t]
    \centering
    \includegraphics[width=\textwidth]{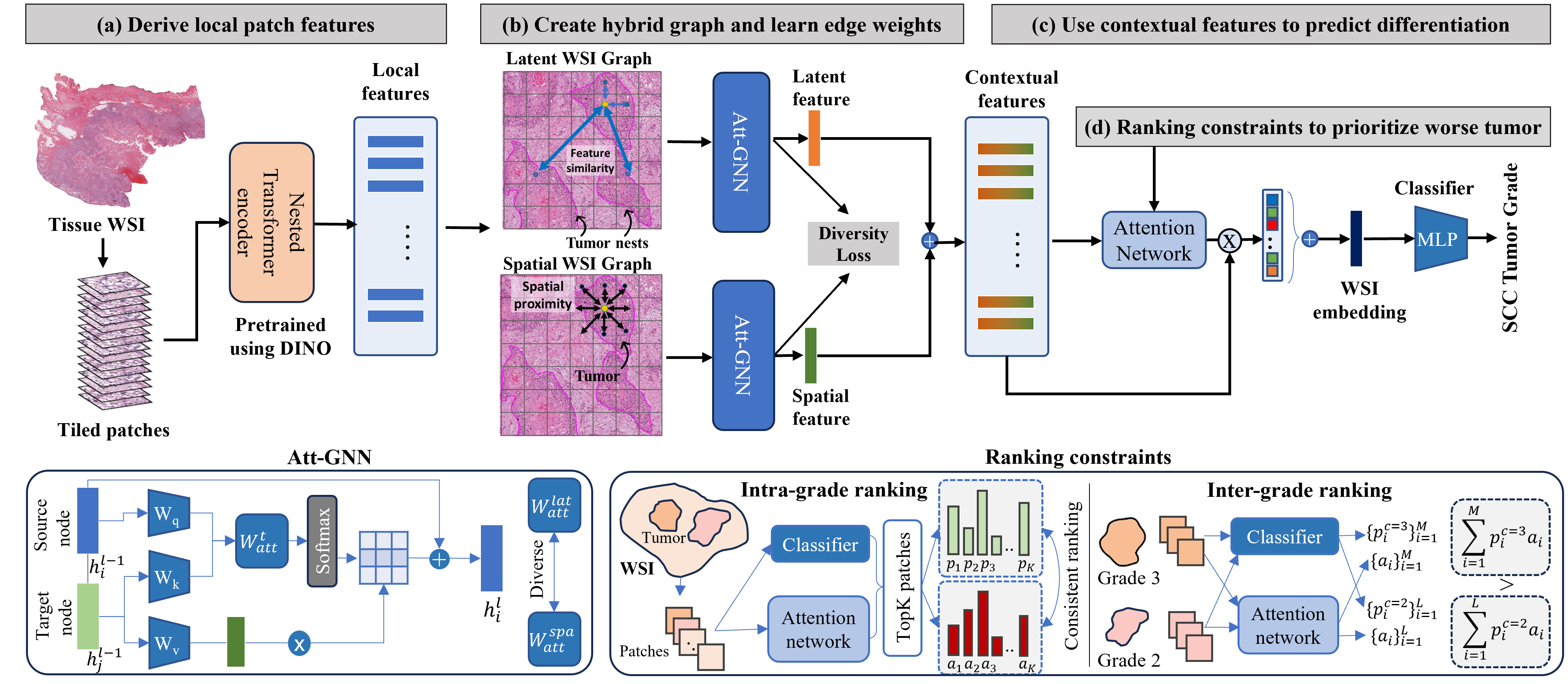}
    \caption{The proposed model. (a) Tissue tiling and local feature extraction. (b) Derivation of contextual patch features using self-attention-based graph convolution, taking latent and spatial semantic dependencies into account. (c) Prediction of grade using multiscale contextual features imposing a rank-order constraint on the attention network.}
    \label{fig:architecture}
\end{figure}

\subsubsection{Contextual feature extraction} 
\begin{figure}[t]
    \centering
    \includegraphics[width=\linewidth]{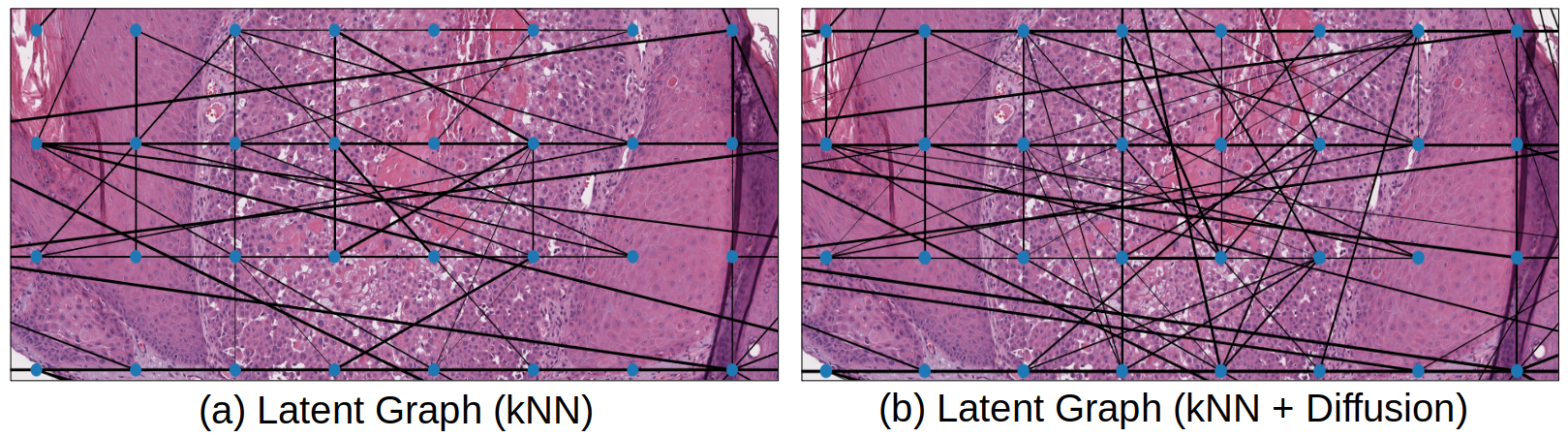}
    \caption{Latent graph refinement using graph diffusion: (a) initial graph defined using 8-nearest neighbours based on feature similarity; (b) refined graph with denser edges. For interpretability, we only highlight the edges with cosine similarity $ \ge 0.5 $}
    \label{fig:diffusion_image}
\end{figure}

Since WSI patches capture localized tissue morphology, explicit modeling is needed to capture grade-relevant spatial relationships and long-range dependencies across tissue regions. To incorporate this contextual information, we represent each WSI using a hybrid graph that captures both phenotypic similarity and spatial context within tumor microenvironment.
\textbf{\\Hybrid graph construction}: To derive contextual patch features, we construct an undirected hybrid graph for each WSI $G = (V, A)$ with patches as nodes $V$ and edges between them defined using adjacency matrix $A \in \mathcal{R}^{N \times N}$. The graph captures two complementary dependencies which are incorporated in $A$: (i) a latent graph $A^{lat}$, linking morphologically similar patches that may be spatially distant; and (ii) a spatial graph $A^{spa}$, connecting neighboring patches in the tumor microenvironment which captures local spatial context. Modeling these relations separately enables the network to capture distinct phenotypic dependencies and local dependencies while reducing oversmoothing due to redundant message passing and diffuse attention. Both graphs are sparsely constructed using k-nearest neighbors (kNN) relations. \\
\textbf{Latent graph for phenotypic similarity}: We construct the latent graph by computing pairwise cosine distance between patch features: 
\begin{equation}
d^{lat}_{ij} = 1-\frac{f_i \cdot f_j}{||f_i||_2||f_j||_2}
\end{equation}
where $f_i$ and $f_j$ represent features of patches $i$ and $j$.  We connect each patch to its k-nearest neighbors in the feature space, i.e. $A^{lat}_{ij} = d^{lat}_{ij}$. A standard kNN graph may introduce noisy edges due to task-agnostic pretrained features and is limited to one-hop neighbors in the feature space. To address these issues, we construct the graph in two stages: 
\begin{enumerate}
    \item \textbf{Initial graph construction}: We first identify the eight nearest neighbors of each patch in the feature space.  To improve robustness, we retain only $k$-reciprocal nearest neighbors $r_{i}(k)$ (\cite{qin2011hello}), such that two patches are connected only if each lies among the other’s $k$-nearest neighbors:
    \begin{equation}
    r_i(k) = j \in \mathcal{N}_k(i) \ \wedge \ i \in \mathcal{N}_k(j) \nonumber
    \end{equation}
    where $\mathcal{N}_k(i)$ denotes the $k$-nearest neighbors in the feature space. Reciprocal kNN reduces spurious connections by retaining edges only between mutually similar patches, improving robustness to outliers in the pretrained feature space. 
    \item \textbf{Graph diffusion}: To capture higher-order relationships beyond one-hop neighbors in the latent graph, we apply kernel-based graph diffusion. We use personalized PageRank \cite{gasteiger2019diffusion} with random walk formulation to derive the diffused adjacency matrix:
\begin{equation}
    \hat{A}^{lat} = \sum^{\infty}_{k=0} \theta_k T^k
\end{equation}
where $\theta_k = \alpha(1-\alpha)^k$ with $\alpha$ being the probability of random jump to parent node ($\alpha = 0.25$),  $T=A^{lat}D^{-1}$ is the normalized adjacency matrix where $D$ is the diagonal matrix, $D = \sum_{i} A^{lat}_{ii}$. Graph diffusion results in a dense matrix $\hat{A}_{ij}$ wherein edge weights represent the cumulative probability of reaching a target node based on feature similarity. Finally, we retain $K=5$ neighbors with highest probabilities per node and truncate smaller values of $\hat{A}_{ij}$, recovering a sparse latent graph $\tilde{A}^{lat}_{ij} = \hat{A}^{lat}_{ij} > 0.02$.  As shown in Figure \ref{fig:diffusion_image}, diffusion strengthens meaningful short- and long-range connections while suppressing weak links between morphologically dissimilar regions.
\end{enumerate}
\textbf{Spatial graph for local tissue context}: We construct the spatial graph by connecting each patch to its eight nearest spatial neighbors. We construct the spatial graph by connecting each patch to its eight nearest spatial neighbors. The distance between patches $i$ and $j$, with coordinates $(s_i,t_i)$ and $(s_j,t_j)$, is
\begin{equation}
    d^{spa}_{ij} = \sqrt{(s_i-s_j)^2 + (t_i-t_j)^2}.
\end{equation}
The spatial graph captures local tissue organization and tumor-microenvironment interactions, including keratinization, cellular organization, and stromal interactions.\\
\textbf{Contextual feature derivation}: We derive the contextual patch features ($h_n^1$) from local patch features ($h_n^0$) using an attention-based graph neural network (\textit{Att-GNN}). Each graph layer performs self-attention-based message-passing over the connected neighbors $j$ of each node $i$ ($A_{ij} > 0$), updating node features via the following operation:
\begin{equation}
H^{l+1} = GConv(H^l,A;W^l) 
\end{equation}
where $H^l = [h^l_1, h^l_2...,h^l_N]$ denotes node representations at layer $l$ and $W^l$ are the trainable parameters. The adjacency matrix $A$ corresponds to the underlying graph, which can be either the spatial graph ($A^{spa}$) or the refined latent graph ($\tilde{A}^{lat}$).

To model heterogeneous tissue interactions in WSI, we incorporate self-attention within each graph leveraging the graph transformer architecture (\cite{hu2020heterogeneous}). Each graph uses single-head self-attention module with a distinct projection head, allowing the model to learn graph-specific contextual relationships. In each graph, node embeddings (target nodes) are projected into $d$-dimensional key vectors using the projection matrix $W_{k} \in \mathbf{R}^{d \times d}$. Neighbouring nodes (source nodes) are independently projected into $d$-dimensional query vectors using the projection matrix $W_q \in R^{d \times d}$. We use distinct matrices for latent ($W_{att}^{lat}$) and spatial ($W_{att}^{spa}$) graphs to prioritize different semantic relations. The attention scores are computed between the node $i$ and its neighbours $j$ using key and query vectors: 
\begin{equation}
\alpha^{t(l)}_{ij} = \frac{(h^{(l)}_i W^{(l)}_k) \cdot W^t_{att} \cdot (h^{(l)}_j W^{l}_q)}{\sqrt{d}}; t \in \{lat, spa\}
\end{equation}
The edge attention scores are normalized over the neighbours of node \( j \in \mathcal{N}(i) \):
\begin{equation}
\beta^{t(l)}_{ij} = \frac{\exp(\alpha^{t(l)}_{ij})}{\sum_{k \in \mathcal{N}(i)} \exp(\alpha^{t(l)}_{ik})}
\end{equation}
Next, we compute the value vector for each neighboring node using the projection matrix $W_v \in R^{d \times d}$. The contextual representation of node $i$ is obtained by attention-weighted aggregation of neighboring value vectors:
\begin{equation}
h^{t(l+1)}_i = \sum_{j \in \mathcal{N}(i)} \beta^{t(l)}_{ij} (h^{(l)}_j W^{(l)}_v)
\end{equation}
Separate attention parameters for the latent and spatial graphs allow the model to capture complementary phenotypic and spatial relationships. Because the two graphs may contain overlapping edges and learn redundant interactions, we introduce a diversity regularizer that minimizes redundancy by reducing the correlation between their attention weight matrices. This is achieved using a cosine similarity-based loss:
\begin{equation}
\mathcal{L}_{\text{diversity}} = \frac{W_{att}^{lat} \cdot W_{att}^{spa}}{\| W_{att}^{lat} \| \| W_{att}^{spa} \|}
\end{equation}
Minimizing this loss encourages the two graph branches to learn complementary contextual information. This enhances the spatial graph's ability to encode local tissue context (e.g., heterogeneity in tumor microenvironment) beyond the phenotypical similarity captured by the latent graph.
The latent and spatial contextual features are then combined, normalized and passed through a non-linear activation (ReLU). A residual connection preserves the original local patch representation, yielding the final contextual feature $h^1_n$:  
\begin{align}
    h^1_n &= h^0_n + ReLU[norm(h^1_{combine})] \nonumber \\
    h^1_{combine} &= GConv^{latent}(h^0_n) + GConv^{spatial}(h^0_n)  
\end{align}
Layer normalization stabilizes graph-network training, while the residual connection preserves fine-grained patch morphology along with the learned contextual information (\cite{li2020deepergcn}).

\subsection{Grade Classification}
We aggregate contextual patch features using attention to emphasize diagnostically relevant regions within each WSI. By assigning higher weights to informative patches, the model can focus on tumor regions that are most relevant for grading in the presence of substantial tissue heterogeneity. Using the 1-hop contextual embeddings $h^1_n$ obtained from the graph network, we compute normalized patch attention weights following \cite{ilse2018attention}:
\begin{equation}
    \label{eqn:diff_attn}
    w_n = \frac{\exp [a^T \tanh(U \cdot h^1_{n})]}{\sum_{n \in \Omega_b} \exp [a^T \tanh(U \cdot h^1_{n})]}
\end{equation}
where $a \in \mathcal{R}^d$ and $U \in \mathcal{R}^{d \times d}$ are learnable parameters. Instead of applying attention pooling directly over patch features, we leverage conjunctive pooling (\cite{early2024inherently}), which aggregates patch-level class scores using the learned attention weights. Patch classification is performed using a temperature-scaled cosine classifier. Specifically, each contextual embedding $h_n^1$ is $l_2$-normalized and compared with a learnable class prototype $z_c$ to derive class logits: 
\begin{equation}
s_n^c = \frac{1}{t} \bigg(\frac{h_n^1}{||h_n^1||} \cdot \frac{z_c}{||z_c||} \bigg)
\end{equation}
where $t$ is temperature parameter, $c \in \{0,1,2,3\}$ represents \{normal, well, moderate, poor\}, $\cdot$ represents dot product. Feature and prototype normalization prevents the classifier from relying on feature magnitude, improving robustness to class imbalance. Finally, the WSI-level likelihood for class $c$ is computed as:
\begin{equation}
    p_b^c = \frac{\exp(s_b^c)}{\sum_{c'}\exp(s_b^{c'})}, s_b^c = \sum_{n=1}^N w_n s_n^c
\end{equation}
This formulation decouples patch importance from patch classification, allowing diagnostically relevant regions to contribute more strongly to the WSI prediction while preserving patch-level grade class confidence required for ordinal ranking.
To further address class imbalance due to a lower proportion of higher-grade cases, we use class-balanced sampling (\cite{cui2019class}) during training. We use the ground-truth WSI grade label $Y_b$ to compute the grade classification loss:
\begin{equation}
\mathcal{L}_{grade} = -\frac{1}{B}\sum_{b=1}^B \sum_{c} \mathbf{1}(Y_b=c) \log(p_b^c)
\end{equation}

\subsubsection{Ordinal Ranking of Patches}
We introduce two ranking constraints to encourage the attention network to reflect the ordinal nature of SCC grading: (i) an inter-grade constraint to impose higher attention values for higher-grade patches, and (ii) an intra-grade constraint to impose higher attention values for more likely patches within the same grade class.

1) \textbf{Inter-grade ranking}:
SCC grading is driven by the most aggressive tumor regions, with pathologists implicitly ranking tumor regions and assigning greater importance to higher-grade tumors. We therefore introduce an \textit{inter-class ranking constraint} to assign higher importance to regions associated with progressively higher grades. We enforce the constraint using pairwise inequality relations such that $w_n^{normal} < w_n^{well} < w_n^{moderate} < w_n^{poor}$.
Since patch-level grade annotations are unavailable, we use the model's cosine softmax classifier to estimate the patch-level class likelihood for grade $c$:
\begin{equation}
p_n^c = p (y_{n}=c|h^1_n) = \frac{\exp(s_{n}^c)}{\sum_{c'}\exp(s_{n}^{c'})}
\end{equation}
To reduce the effect of noisy patch predictions, we retain only high-confidence patches with $p_n^c > \tau$ ($\tau = 0.6)$. For each WSI $b$, let $\Omega_b^c$ denote the set of confident retained patches for grade $c$. We then compute a grade-specific attention score for the WSI as a probability-weighted average of the unnormalized patch attention scores:
\begin{equation}
A_b^c = \frac{\sum_{n \in \Omega_b^c} p_{n}^c a_n}{\sum_{n \in \Omega_b^c} p_n^c}
\end{equation}
where $a_{n} = a^T \tanh(U \cdot h_n^1)$ is unnormalized attention score for patch $n$ (Eq. \ref{eqn:diff_attn}). 
The weighted aggregation reflects the relative grade evidence among the selected patches and suppresses ambiguous patches, leading to a more stable WSI-level estimate of grade-specific importance under weak supervision. Using unnormalized attention avoids competition across patches and preserves the contribution of multiple grade-relevant regions. Normalized probability weights further reduce sensitivity to the number of selected patches, yielding a more stable estimate of grade-specific importance. To prioritize higher-grade tumor regions, we define a set of pairwise comparisons $(i, j)$ between WSIs from adjacent grades which is sufficient to encode the overall ordinal relationship: $Z$ = $\{(i,j); Y_{i} = c, Y_{j} = c+1\}$. We then enforce an ordinal relationship between their grade-specific importance using a pairwise ranking loss (\cite{10.1145/1102351.1102363}) :
\begin{equation}
    \mathcal{L}_{inter} = \sum_{i,j \in Z} \log[1 + \exp(A^c_i - A^{c+1}_j)]
\end{equation}

2) \textbf{Intra-grade ranking}: We further introduce an \emph{intra-grade ranking} constraint to encourage attention network to prioritize the more confident tumor regions within each grade. Ranking all patches within a WSI is noisy and computationally expensive, so we select a candidate set comprising patches that either have high attention \(a_n\) or high predicted probability \(p_n^c\) for grade $c$:
\[
S_b = \{n \mid n \in \mathrm{Top}_K(a_n) \cup \mathrm{Top}_K(p^c_{n})\},
\]
We retain only patches with $p_n^c > \tau (\tau = 0.6)$. Instead of ranking all possible pairs of patches, we group patch probabilities into coarse confidence levels using intervals of width $0.2$ and construct ordered patch pairs between patches belonging to different discrete bins
\[
\tilde{Z}
=
\left\{(i,j)\;\middle|\; q_i^c < q_j^c,\; i,j \in S_b\right\}.
\]
where $q_n^c = \left\lfloor \frac{p_n^c}{0.2} \right\rfloor$.
We apply the pairwise ranking loss (\cite{10.1145/1102351.1102363}) to align attention scores with the relative class confidence:
\[
\mathcal{L}_{\text{intra}} = \sum_{(i, j) \in \tilde{Z}} \log (1 + \exp(a_i - a_j))
\]
This encourages patches with stronger grade evidence within a WSI to receive higher attention, thereby improving agreement between patch-level predictions and slide-level aggregation.

\textbf{Overall Loss}: The final objective combines grade classification, inter-grade ranking, intra-grade ranking and graph diversity:
\begin{equation}
    \mathcal{L}_{total} = \mathcal{L}_{grade} + \lambda_1(\mathcal{L}_{inter} + \mathcal{L}_{intra}) + \lambda_2 \mathcal{L}_{diversity}\\     
\end{equation}
where $\lambda_1$ and $\lambda_2$ are used to balance the losses and determined using hyperparameter tuning.

\section{Experimental Setup}
\subsection{Datasets} 
To validate the effectiveness of our proposed framework, we assess multi-class SCC grading on public and private datasets: two publicly available datasets for lung SCC and head and neck SCC collated from the Cancer Genome Atlas (TCGA) (\cite{tomczak2015review}) and Clinical Proteomic Tumor Analysis Consortium (CPTAC) (\cite{edwards2015cptac}), along with our own collected histology dataset for skin SCC. Each WSI is associated with a single grade label annotated by pathologists, and no pixel-level annotation was used for training. Table \ref{tab1:data_distribution} provides a high-level grade distribution for each dataset.\\
\underline{\textit{Private Dataset (Skin SCC)}}: We curated a dataset of 815 formalin-fixed paraffin-embedded hematoxylin and eosin (H\&E)-stained WSIs, each from a unique patient, collected at the Mayo Clinic, USA. The dataset was collected between 2017 and 2022, and includes 247 normal, 383 well-differentiated, 108 moderately-differentiated, and 77 poorly-differentiated cases. The majority of patients were elderly white individuals (mean age: 82 years). WSIs were scanned using an Aperio AT2 scanner at $40X$ magnification ($0.25 \mu \text{m/pixel}$) and processed at $20X$ magnification for patch feature extraction. Each slide was independently graded by two expert pathologists based on cellular differentiation, followed by review and confirmation by a supervising dermatologist. In instances of disagreement, the re-reviewing pathologist and dermatologist discussed to arrive at a consensus and excluded cases in which no consensus was achieved. The dataset exhibits morphological heterogeneity across grade classes, including stain variations, tumor growth patterns, and tissue processing-related differences (e.g., biopsy type and sectioning effects, with higher-grade tumors often yielding larger specimens). Figure \ref{fig:cscc_variability} (Appendix) includes representative cases illustrating the diversity across normal, well-, moderately-, and poorly-differentiated SCCs. A subset of WSIs was annotated at the pixel level to delineate tumor regions for evaluating model localization during inference only.\\
\underline{\textit{Public Datasets}}:  To highlight the strong performance of our model across diverse SCC tumors, we utilized the publicly available cancer datasets from TCGA and CPTAC. \\
\textit{Lung SCC}: For lung cancer, we use 289 diagnostic FFPE H\&E WSIs from the CPTAC-LSCC study, representing 109 patients. The dataset comprises 161 normal, 72 moderately-differentiated, and 56 poorly-differentiated cases. Due to the lack of well-differentiated tumors, lung SCC grading was formulated as a 3-class classification task. All WSIs were scanned at $20X$ magnification ($0.5 \mu \text{m/pixel}$), and grade labels were assigned at the patient level. \\
\textit{Head and Neck SCC}: The head and neck SCC dataset is assembled from 310 WSIs from TCGA-HNSC and 362 WSIs from CPTAC-HNSCC, totaling 672 WSIs. It includes 105 normal, 79 well-differentiated, 356 moderately-differentiated, and 132 poorly-differentiated tumors. All diagnostic WSIs for each patient were included in the analysis. To address the scarcity of well-differentiated samples in individual datasets and leverage normal WSIs available in CPTAC-HNSCC, both sources were merged. CPTAC WSIs were scanned at $20X$ magnification ($0.5 \mu \text{m/pixel}$) while TCGA  WSIs were acquired at $40X$ magnification ($0.25 \mu \text{m/pixel}$). For TCGA, grade labels were derived from pathology reports available via cBioPortal (\cite{gao2013integrative}).

\begin{table}[t!]
\centering
\caption{Number of WSIs by grade for each dataset and evaluation setting. ID denotes in-distribution evaluation and OOD denotes out-of-distribution evaluation.}
\label{tab1:data_distribution}
\resizebox{0.78\textwidth}{!}{
\begin{tabular}{|l|l|c|ccccc|}
\hline
\multirow{2}{*}{\textbf{Dataset}} &
\multirow{2}{*}{\textbf{Source}} &
\multirow{2}{*}{\textbf{Usage}} &
\multicolumn{5}{c|}{\textbf{Number of WSIs}} \\ \cline{4-8}
& & &
\textbf{Normal} &
\textbf{Well} &
\textbf{Moderate} &
\textbf{Poor} &
\textbf{Total} \\ \hline

Skin
& Private
& Train, Test
& 247 & 383 & 108 & 77 & 815 \\ \hline

Lung
& CPTAC
& Train, Test
& 161 & -- & 72 & 56 & 289 \\ \hline

Head \& Neck (ID)
& CPTAC, TCGA
& Train, Test
& 105 & 79 & 356 & 132 & 672 \\ \hline

\multirow{2}{*}{Head \& Neck (OOD)}
& TCGA
& Train
& -- & 59 & 277 & 120 & 456 \\ \cline{2-8}

& CPTAC
& Test
& -- & 20 & 79 & 12 & 111 \\ \hline

\end{tabular}
}
\end{table}

\subsection{Pathologists Review}
We assess our trained model to determine its effectiveness in improving pathologists' performance in diagnosing skin SCC tumor grades. Two board-certified pathologists analyzed 12 WSIs, each diagnosing 6 cases with and without model assistance, separated by a cooling period of three months between diagnoses. During diagnosis, pathologists reviewed the WSI with tumor-grade heatmaps overlaid on the image. They were required to provide feedback on the impact of model assistance, focusing on the localization of grade-relevant tumor regions, identification of false positives and false negatives, and the model's impact on diagnosis efficiency.

\subsection{Training Details}
To robustly evaluate RACR-MIL across different feature backbones, we use three patch-level encoders: (i) SCC-DINO, our SCC-specific NesT-S encoder \cite{zhang2022nested} pretrained using DINO, (ii) PathGen-CLIP (\cite{ICLR2025_ebf8764e}), a pathology vision–language foundation model pretrained on 1.6M histopathology image–text pairs, and (iii) CONCHv1.5 (\cite{lu2024visual}) which was trained with 1.26 M image-text pairs using CoCa objective. SCC-DINO is pretrained separately for each SCC dataset, whereas PathGen-CLIP is pretrained on pan-cancer and provides transferable representations that capture both morphological patterns and semantic context. CONCHv1.5 is trained on histology images spanning diverse stains and tissue types and generates robust region-level features. Using features derived from each extractor, RACR-MIL is trained to classify WSIs into four classes (Normal, Well-, Moderately-, and Poorly-differentiated) for skin SCC and head and neck SCC, and three classes (Normal, Moderately-, and Poorly-differentiated) for lung SCC. The graph module and grade classifier are jointly optimized using AdamW with a learning rate of $10^{-4}$. The optimal hyperparameters for RACR-MIL training are reported in Table~\ref{tab:hyperparameters} (Appendix). For each feature extractor, we use the same ranking-loss weights across all SCC datasets. For SCC-DINO features, we use a 64-dimensional embedding in our model, whereas for the 768-dimensional PathGen-CLIP and CONCHv1.5 features, we use a 128-dimensional embedding. Model evaluation is performed using five-fold cross-validation with patient-level stratification to address class imbalance. Each test fold comprises 20\% of patients, while the remaining 80\% are divided into training (64\%) and validation (16\%) sets. We use seven evaluation metrics averaged over the folds to assess the model's performance: Macro-averaged Precision, Recall and F1 score; Quadratic-weighted $\kappa$; Class-wise accuracy; Area Under Curve (AUC) for low- vs high-risk (well- vs moderate/poor) grades; and Matthews Correlation Coefficient (MCC). Results are reported as mean $\pm$ standard deviation.\\
\textit{Skin SCC and Lung SCC}: We evaluate all three feature extractors for skin and lung SCC cohorts. We report in-distribution evaluation for both datasets without applying stain normalization. \\
\textit{Head and Neck SCC}: For HNSC, we account for cohort-specific batch effects arising from merging TCGA and CPTAC. We report (i) an in-distribution (ID) evaluation on the merged cohort and (ii) an out-of-distribution (OOD) evaluation where we train on TCGA and test on CPTAC. For ID evaluation, we leverage the SCC-DINO encoder and apply stain normalization before patch feature extraction to better align feature distributions across the two cohorts (Figure \ref{fig:umap_hnsc}, Appendix). For the OOD setting, we restrict evaluation to three grade classes (Well, Moderate and Poor) and train on TCGA cases and evaluate on CPTAC. Here, we use PathGen-CLIP feature encoder due to its stronger cross-domain transferability and extract features from the original (non-normalized) patches.

\subsection{Tissue Extraction} We preprocess each WSI using Otsu thresholding followed by morphological operations to remove background regions and irrelevant tissue sections ($area < 20$). Slides acquired at $40X$ are downsampled to $20X$, while CPTAC slides are processed at their native $20X$ resolution. We tile each WSI into non-overlapping patches, using $448 \times 448$ patch size for SCC-DINO and CONCHv1.5 encoders and $336\times336$ for PathGen-CLIP encoder. Finally, we remove low-texture patches using an image gradient-based entropy threshold of $4$. 

\subsection{Baselines}
We compare RACR-MIL against a broad set of state-of-the-art MIL baselines, with particular emphasis on \emph{contextual} methods that model inter-patch dependencies. Baselines are grouped into two categories: 1) non-contextual methods that treat patches independently, including ABMIL (\cite{ilse2018attention}), Gated ABMIL (\cite{ilse2018attention}), CLAM-MB (\cite{lu2021data})), Mean Pooling and Max Pooling; 2) contextual methods that account for patch dependencies, including PatchGCN (\cite{chen2021whole}), DSMIL (\cite{li2021dual}), TransMIL (\cite{shao2021transmil}), WiKG (\cite{li2024dynamic}), CAMIL (\cite{fourkioti2023camil}), LongMIL (\cite{li2023longmil}), HEAT(\cite{chan2023histopathology})  and Graph Transformer (GTP) (\cite{zheng2022graph}). All baselines, except Mean Pooling and Max Pooling, leverage attention-based feature pooling. 
Among non-contextual baselines, CLAM-MB extends ABMIL by adding a clustering mechanism on individual patches to enhance discriminability. Contextual baselines span graph- and transformer-based designs: PatchGCN applies graph convolution over a spatial kNN graph while WiKG dynamically constructs head-tail relations from instance embeddings and applies directed message passing with a knowledge-aware attention mechanism. HEAT constructs a heterogeneous graph over tissue node types and incorporates semantic edge attributes during message-passing. Transformer-based methods include TransMIL which models short- and long-range patch dependencies via self-attention, and LongMIL, which scales to large patch sets using a local-global hybrid transformer with locality-based attention masking. GTP combines graph-derived contextual patch features with transformer for global aggregation, and CAMIL integrates global self-attention with prior-guided spatial neighborhood attention. DSMIL employs dual instance- and bag-level classifiers coupled with self-attention to model patch interactions.

Additionally, we compare variants of our RACR model that incorporate only subsets of the proposed innovations (e.g., spatial or latent contextual features, attention ranking) to evaluate their contributions to classification performance. We also evaluate the tumor localization capability of our approach on the skin SCC dataset. For a fair comparison, all approaches use the same pre-trained feature extractor. Baseline hyperparameters (e.g., learning rate, dropout rate, number of layers, embedding dimension) are tuned on the corresponding validation sets, while batch size and optimizer weight decay are held consistent across methods to ensure a fair comparison.

\section{Results}
\subsection{Grade Classification}

\begin{table*}[t!]
\centering
\small
\setlength{\tabcolsep}{4pt}
\caption{Comparison of RACR-MIL with state-of-the-art MIL approaches for SCC grading using multiple feature extractor backbones. For each evaluation metric, the mean and standard deviation across five cross-validation folds are reported. \textbf{Bold} and \underline{Underline} indicate the highest and second highest performance.}
\label{classification_result}
\resizebox{\textwidth}{!}{
\begin{tabular}{llccccc|ccccc}
\toprule
\multirow{2}{*}{\textbf{Approach}} &
\multirow{2}{*}{\begin{tabular}[c]{@{}c@{}}\textbf{Feature} \\ \textbf{Extractor}\end{tabular}} &
\multicolumn{5}{c}{\textbf{Skin SCC}} &
\multicolumn{5}{c}{\textbf{Lung SCC}} \\
\cmidrule(lr){3-7}\cmidrule(lr){8-12}
&& F1 & Precision & Recall & AUC & MCC
& F1 & Precision & Recall & AUC & MCC
\\
\midrule
\multicolumn{12}{l}{\textbf{Non-contextual}} \\
\midrule
\multirow{2}{*}{Max Pooling}
& SCC-DINO
& 0.761 {\tiny 0.039}
& 0.770 {\tiny 0.040}
& 0.776 {\tiny 0.044}
& 0.942 {\tiny 0.011}
& 0.762 {\tiny 0.053}
& 0.778 {\tiny 0.075}
& 0.788 {\tiny 0.072}
& 0.784 {\tiny 0.066}
& 0.934 {\tiny 0.026}
& 0.770 {\tiny 0.090}
\\
& PathGen-CLIP & 0.771 \tiny 0.009 & 0.769 \tiny 0.019 & 0.788 \tiny 0.022 & 0.953 \tiny 0.009 & 0.768 \tiny 0.019 & 0.749 \tiny 0.084& 0.758 \tiny 0.082& 0.761 \tiny 0.089& 0.937 \tiny 0.024& 0.739 \tiny 0.091 \\
\midrule

\multirow{2}{*}{Mean Pooling}
& SCC-DINO
& 0.758 {\tiny 0.027}
& 0.756 {\tiny 0.030}
& 0.771 {\tiny 0.030}
& 0.943 {\tiny 0.014}
& 0.754 {\tiny 0.041}
& \textbf{0.797} {\tiny 0.041}
& \underline{0.796} {\tiny 0.040}
& \textbf{0.809} {\tiny 0.047}
& \underline{0.942} {\tiny 0.012}
& \textbf{0.779} {\tiny 0.055}
\\
& PathGen-CLIP & 0.752 \tiny0.023 & 0.758 \tiny 0.021 & 0.763 \tiny 0.036 & 0.941 \tiny 0.012 & 0.746 \tiny 0.037 & 0.762 \tiny 0.066& 0.773 \tiny 0.067& 0.776 \tiny 0.079& 0.941 \tiny 0.019& 0.748 \tiny 0.083 \\
\midrule
 
\multirow{2}{*}{ABMIL}
& SCC-DINO
& \underline{0.782} {\tiny 0.027}
& 0.774 {\tiny 0.035}
& \underline{0.805} {\tiny 0.012}
& 0.954 {\tiny 0.009}
& \underline{0.781} {\tiny 0.042}
& 0.767 {\tiny 0.060}
& 0.771 {\tiny 0.063}
& 0.787 {\tiny 0.072}
& 0.945 {\tiny 0.026}
& 0.746 {\tiny 0.075}
\\
& PathGen-CLIP
& 0.780 {\tiny 0.041}
& 0.772 {\tiny 0.036}
& 0.807 {\tiny 0.050}
& 0.952 {\tiny 0.019}
& 0.758 {\tiny 0.051}
& 0.763 \tiny 0.071& 0.767 \tiny 0.069& 0.773 \tiny 0.073& \underline{0.946} \tiny 0.023& 0.747 \tiny 0.081 \\
\midrule

\multirow{2}{*}{GABMIL}
& SCC-DINO
& 0.770 {\tiny 0.021}
& 0.770 {\tiny 0.026}
& 0.787 {\tiny 0.017}
& \underline{0.957} {\tiny 0.010}
& 0.780 {\tiny 0.020}
& 0.749 \tiny 0.046 &	0.754 \tiny 0.045 &	0.765 \tiny 0.056 &	0.942 \tiny 0.026 &	0.724 \tiny 0.063 
\\
& PathGen-CLIP & 0.775 \tiny 0.029	& 0.768 \tiny 0.023 & 0.803 \tiny 0.042 &	0.953 \tiny 0.013 &	0.757 \tiny 0.038 & 0.749 \tiny 0.066 & 0.757 \tiny 0.063 &	0.766 \tiny 0.072 & \textbf{0.947} \tiny 0.025 & 0.732 \tiny 0.078 \\
\midrule

\multirow{2}{*}{CLAM-MB}
& SCC-DINO
& 0.778 {\tiny 0.024}
& \underline{0.775} {\tiny 0.030}
& 0.789 {\tiny 0.017)}
& \underline{0.954} {\tiny 0.011}
& \underline{0.781} {\tiny 0.030}
& 0.742 \tiny 0.069 &	0.749 \tiny 0.066 &	0.758 \tiny 0.071 &	0.938 \tiny 0.029 &	0.721 \tiny 0.079
\\
& PathGen-CLIP & \underline{0.793} \tiny 0.045 & 0.786 \tiny 0.040 & \underline{0.811} \tiny 0.055 & \textbf{0.958 \tiny 0.012} & 0.779 \tiny 0.054 & 0.733 \tiny 0.073& 0.749 \tiny 0.080& 0.752 \tiny 0.083& 0.942 \tiny 0.020& 0.717 \tiny 0.086 \\
\midrule
\multicolumn{12}{l}{\textbf{Contextual}} \\
\midrule

\multirow{2}{*}{DSMIL}
& SCC-DINO
& 0.761 {\tiny 0.042}
& 0.762 {\tiny 0.038}
& 0.774 {\tiny 0.052}
& 0.950 {\tiny 0.015}
& 0.770 {\tiny 0.040}
& 0.762 \tiny 0.048 & 0.773 \tiny 0.052 & 0.780 \tiny 0.050 & \textbf{0.943} \tiny 0.022 & 0.739 \tiny 0.054
\\
& PathGen-CLIP
& 0.779 {\tiny 0.025}
& 0.776 {\tiny 0.019}
& 0.793 {\tiny 0.038}
& 0.954 {\tiny 0.011}
& 0.772 {\tiny 0.031}
& \underline{0.775} \tiny 0.044& \underline{0.780} \tiny 0.041& \underline{0.790} \tiny 0.057& 0.944 \tiny 0.025& 0.758 \tiny 0.058 \\
\midrule

\multirow{2}{*}{PatchGCN}
& SCC-DINO
& 0.772 {\tiny 0.019}
& 0.762 {\tiny 0.023}
& 0.799 {\tiny 0.027}
& 0.950 {\tiny 0.010}
& 0.766 {\tiny 0.042}
& 0.761 \tiny 0.089 
& 0.766 \tiny 0.086 
& 0.775 \tiny 0.092 
& 0.937 \tiny 0.029 
& 0.740 \tiny 0.099
\\
& PathGen-CLIP
& 0.775 {\tiny 0.045}
& 0.789 {\tiny 0.041}
& 0.787 {\tiny 0.053}
& 0.956 {\tiny 0.012}
& 0.772 {\tiny 0.042}
& 0.728 \tiny 0.085& 0.735 \tiny 0.082& 0.740 \tiny 0.087& 0.937 \tiny 0.021& 0.704 \tiny 0.096 \\
\midrule

\multirow{2}{*}{HEAT}
& SCC-DINO & 0.764 \tiny 0.042 & 0.761 \tiny 0.043 & 0.785 \tiny 0.057 & 0.953 \tiny 0.011 & 0.755 \tiny 0.047 & 0.734 \tiny 0.020 & 0.738 \tiny 0.025 & 0.737 \tiny 0.016 & 0.931 \tiny 0.014 & 0.705 \tiny 0.027\\
& PathGen-CLIP & 0.775 \tiny 0.036 & 0.779 \tiny 0.027 & 0.784 \tiny 0.055 & 0.945 \tiny 0.015 & 0.777 \tiny 0.043 & 0.735 \tiny 0.061 & 0.745 \tiny 0.060 & 0.761 \tiny 0.051 & 0.932 \tiny 0.021 & 0.703 \tiny 0.064  \\
\midrule

\multirow{2}{*}{WiKG}
& SCC-DINO
& 0.763 {\tiny 0.007}
& 0.770 {\tiny 0.014}
& 0.772 {\tiny 0.017}
& 0.940 {\tiny 0.011}
& 0.768 {\tiny 0.019}
& 0.754 \tiny 0.088 & 0.760 \tiny 0.094 & 0.757 \tiny 0.084 & 0.938 \tiny 0.032 & 0.735 \tiny 0.100
\\
& PathGen-CLIP
& 0.730 {\tiny 0.025}
& 0.739 {\tiny 0.020}
& 0.735 {\tiny 0.048}
& 0.935 {\tiny 0.012}
& 0.743 {\tiny 0.026}
& 0.747 \tiny 0.071& 0.754 \tiny 0.070& 0.751 \tiny 0.069& 0.935 \tiny 0.025& 0.731 \tiny 0.099 \\
\midrule

\multirow{2}{*}{CAMIL}
& SCC-DINO & 0.770 \tiny 0.027 & 0.769 \tiny 0.034 & 0.780 \tiny 0.017 & 0.951 \tiny 0.012 & 0.768 \tiny 0.044 & \underline{0.791} \tiny 0.049 & \textbf{0.797} \tiny 0.051	& 0.804 \tiny 0.055 & 0.943 \tiny 0.026	& \underline{0.771} \tiny 0.060 \\
& PathGen-CLIP & 0.767 \tiny 0.029 & 0.765 \tiny 0.024 & 0.781 \tiny 0.042 & \underline{0.956} \tiny 0.014 & 0.763 \tiny 0.033 & 0.722 \tiny 0.086 & 0.736 \tiny 0.077& 0.744 \tiny 0.093& 0.940 \tiny 0.023& 0.709 \tiny 0.103 \\
\midrule

\multirow{2}{*}{TransMIL}
& SCC-DINO
& 0.718 {\tiny 0.041}
& 0.728 {\tiny 0.047}
& 0.731 {\tiny 0.047}
& 0.919 {\tiny 0.015}
& 0.727 {\tiny 0.049}
& 0.736 \tiny 0.083	& 0.749 \tiny 0.086 &	0.757 \tiny 0.079 &	0.940 \tiny 0.023 &	0.724 \tiny 0.079
\\
& PathGen-CLIP
& 0.756 {\tiny 0.025}
& 0.752 {\tiny 0.025}
& 0.774 {\tiny 0.037}
& 0.944 {\tiny 0.017}
& 0.746 {\tiny 0.040}
& 0.748 \tiny 0.071& 0.763 \tiny 0.079& 0.763 \tiny 0.074& 0.934 \tiny 0.016& 0.735 \tiny 0.080 \\
\midrule

\multirow{2}{*}{LongMIL}
& SCC-DINO
& 0.765 {\tiny 0.037}
& 0.766 {\tiny 0.034}
& 0.773 {\tiny 0.040}
& 0.945 {\tiny 0.011}
& 0.770 {\tiny 0.044}
& 0.748 {\tiny 0.048}
& 0.752 {\tiny 0.052}
& 0.751 {\tiny 0.045}
& 0.934 {\tiny 0.024}
& 0.726 {\tiny 0.047}
\\
& PathGen-CLIP & 0.780 \tiny 0.028 & 0.786 \tiny 0.030 & 0.789 \tiny 0.034 & 0.956 \tiny 0.005 & \underline{0.791} \tiny 0.028 & 0.758 \tiny 0.062 & 0.769 \tiny 0.065 & 0.763 \tiny 0.063	& 0.940 \tiny 0.021 & \underline{0.762} \tiny 0.037 \\
\midrule

\multirow{2}{*}{GTP}
& SCC-DINO
& 0.728 {\tiny 0.034}
& 0.737 {\tiny 0.042}
& 0.732 {\tiny 0.035}
& 0.934 {\tiny 0.019}
& 0.725 {\tiny 0.058}
& 0.732 \tiny 0.058 &	0.738 \tiny 0.061 &	0.738 \tiny 0.063 &	0.930 \tiny 0.022 &	0.710 \tiny 0.066
\\
& PathGen-CLIP
& 0.769 {\tiny 0.034}
& \underline{0.793} {\tiny 0.032}
& 0.767 {\tiny 0.042}
& 0.945 {\tiny 0.017}
& 0.781 {\tiny 0.036}
& 0.716 \tiny 0.090& 0.726 \tiny 0.094& 0.729 \tiny 0.095& 0.927 \tiny 0.035& 0.691 \tiny 0.105 \\
\midrule

\multirow{2}{*}{RACR-MIL}
& SCC-DINO
& \textbf{0.807} {\tiny 0.013}
& \textbf{0.796} {\tiny 0.019}
& \textbf{0.831} {\tiny 0.020}
& \textbf{0.958} {\tiny 0.007}
& \textbf{0.801} {\tiny 0.027}
& 0.790 {\tiny 0.070} & 0.790 {\tiny 0.068} & \underline{0.806} {\tiny 0.081} & \underline{0.942} {\tiny 0.026} & \underline{0.771} {\tiny 0.088}
\\
& PathGen-CLIP
& \textbf{0.805} \tiny 0.025 & \textbf{0.801} \tiny 0.019 & \textbf{0.824} \tiny 0.043 & 0.955 \tiny 0.011 & \textbf{0.797} \tiny 0.027
& \textbf{0.786} \tiny 0.049	& \textbf{0.796} \tiny 0.042 & \textbf{0.805} \tiny 0.047 & 0.938 \tiny 0.024 & \textbf{0.775} \tiny 0.045 \\

\bottomrule
\end{tabular}
}

\end{table*}
\subsubsection{Enhanced grading performance on skin SCC}
We first evaluate RACR-MIL on the internal skin SCC cohort and observe consistent improvements over state-of-the-art MIL baselines across all metrics (Table \ref{classification_result} and Table \ref{classification_result_conch}) (Appendix). Notably, the gains are consistent across all feature extractor backbones, highlighting the model's robustness on skin SCC. RACR-MIL outperforms non-contextual methods with 1-5\% higher F1 and 2-5\% higher MCC scores. It improves over contextual approaches by 2-9\% in F1 (up to 8\% in MCC) and outperforms transformer-based models (e.g., TransMIL, GTP, CAMIL). This underscores the benefit of explicitly encoding semantic relationships using graphs rather than relying on full pairwise attention. Relative to contextual methods that model either latent (HEAT, DSMIL) or spatial (PatchGCN, WiKG) dependencies, our dual graph formulation yields a further 2-5\% F1 improvement, indicating more effective contextual feature aggregation. Consistent with prior weakly-supervised grading studies (\cite{dominguez2024systematic}), ABMIL proved to be a strong baseline, outperforming most contextual methods across both feature extractors, except RACR-MIL. This suggests that explicit tissue context modeling with consistent prioritization of relevant regions can enable MIL to surpass non-contextual pooling. While larger patch encoders (e.g., $448 \times 448$) can partially compensate for limited context, modeling broader tissue interactions remains important for achieving optimal grading performance under weak supervision. RACR-MIL further shows lower variance across all metrics compared to the next-best baselines (ABMIL, CLAM-MB), supporting its robustness across feature spaces.

\subsubsection{Generalizability across diverse SCCs}
\textit{Lung SCC}: As shown in Table \ref{classification_result} and Table \ref{classification_result_conch} (Appendix), RACR-MIL demonstrates competitive performance on lung SCC, particularly with PathGen-CLIP features where it achieves the best overall results. It outperforms graph-based methods by more than 4\% in F1 and MCC scores and shows a clear margin over DSMIL, the most competitive contextual baseline. Notably, RACR-MIL also exhibits lower variability than graph- and transformer-based baselines, indicating a more stable learning across folds. Using CONCH-derived features, RACR-MIL outperforms graph-based contextual baselines by 3-6\%. With SCC-DINO features, mean pooling outperforms RACR-MIL slightly, but the \emph{latent-graph} variant with ranking constraint achieves the best F1 score on lung SCC exceeding all baselines (Table \ref{semantic_spatial}). This suggests that lung SCC benefits more from modeling non-local dependency between spatially dispersed tumor regions, and that the ranking constraint helps prioritize the highest-grade evidence during WSI-level aggregation.


\begin{table*}
\centering
\small
\setlength{\tabcolsep}{4pt}

\caption{Comparison of MIL methods for head and neck SCC grading under in-distribution (ID) and out-of-distribution (OOD) evaluation. Results are reported as mean and standard deviation over five folds. Bold and \underline{underline} denote the best and second-best methods, respectively.}

\resizebox{\textwidth}{!}{

\begin{tabular}{lccccc|ccccc}

\toprule

\multirow{2}{*}{\textbf{Approach}}

& \multicolumn{5}{c}{\textbf{Head and Neck SCC (ID : SCC-DINO)}}
& \multicolumn{5}{|c}{\textbf{Head and Neck SCC (OOD : PathGen-CLIP)}}

\\

\cmidrule(lr){2-6}
\cmidrule(lr){7-11}

& F1 & Prec & Rec & AUC & MCC
& F1 & Prec & Rec & AUC & MCC

\\

\midrule
\multicolumn{11}{l}{\textbf{Non-contextual}} \\
\midrule

Max Pooling
& 0.510 \tiny 0.026 & 0.529 \tiny 0.022 & 0.571 \tiny 0.028 & 0.781 \tiny 0.019 & 0.328 \tiny 0.022 & 0.483 \tiny 0.029&0.477 \tiny 0.019&0.542 \tiny 0.037&0.721 \tiny 0.015&0.205 \tiny 0.035

\\

Mean Pooling
 & 0.541 \tiny 0.033 & 0.543 \tiny 0.041 & 0.592 \tiny 0.042 & 0.800 \tiny 0.025 &	0.363 \tiny 0.061 &0.485 \tiny 0.044&0.517 \tiny 0.021&0.534 \tiny 0.022&0.721 \tiny 0.009&0.240 \tiny 0.031

\\

ABMIL
& 0.574 \tiny 0.059 &	0.569 \tiny 0.056 & 0.613 \tiny 0.065 & 0.819 \tiny 0.020 & 0.395 \tiny 0.088 &0.456 \tiny 0.017&0.463 \tiny 0.028&0.562 \tiny 0.039&0.710 \tiny 0.019&0.207 \tiny 0.050

\\

GABMIL
& \underline{0.578 \tiny 0.036} & \underline{0.572 \tiny 0.034} & \underline{0.628 \tiny 0.042} & \underline{0.822 \tiny 0.020} & \underline{0.407 \tiny 0.047} &0.474 \tiny 0.031&0.481 \tiny 0.040&0.578 \tiny 0.033& \textbf{0.728 \tiny 0.019} & 0.230 \tiny 0.048
\\

CLAM-MB
& 0.576 \tiny 0.050 &	0.573 \tiny 0.047 &	0.622 \tiny 0.066 &	0.814 \tiny 0.024 &	0.400 \tiny 0.075 & 0.463 \tiny 0.039 &	0.472 \tiny 0.040 &	\textbf{0.581 \tiny 0.036} & 0.718 \tiny 0.017 &	0.221 \tiny 0.057
\\

\midrule
\multicolumn{11}{l}{\textbf{Contextual}} \\
\midrule

PatchGCN
& 0.562 \tiny 0.011 &	0.561 \tiny 0.017 &	0.587 \tiny 0.027 &	0.807 \tiny 0.018 &	0.398 \tiny 0.046 & \underline{0.488 \tiny 0.046} &0.502 \tiny 0.060&0.565 \tiny 0.038&0.707 \tiny 0.048&0.239 \tiny 0.058
\\


WiKG
& 0.488 \tiny 0.014 &	0.472 \tiny 0.016 &	0.531 \tiny 0.016	& 0.748 \tiny 0.027 &	0.310 \tiny 0.026 & 0.448 \tiny 0.020 &	0.451 \tiny 0.016 & 0.479 \tiny 0.047 & 0.685 \tiny 0.021 & 0.136 \tiny 0.045

\\

CAMIL
& 0.558 \tiny 0.048 &	0.557 \tiny 0.050 &	0.608 \tiny 0.046 &	0.811 \tiny 0.017	& 0.381 \tiny 0.070 & 0.487 \tiny 0.024 & 0.492 \tiny 0.043	& 0.565 \tiny 0.051 & \underline{0.708 \tiny 0.022} & 0.226 \tiny 0.036
\\

TransMIL
& 0.535 \tiny 0.030 & 0.539 \tiny 0.031 & 0.590 \tiny 0.032 & 0.794 \tiny 0.011 & 0.365 \tiny 0.056 & 0.452 \tiny 0.035 &	0.457 \tiny 0.037 & 0.520 \tiny 0.033 &	0.700 \tiny 0.022 & 0.175 \tiny 0.041
\\

LongMIL
& 0.559 \tiny 0.042 & 0.555 \tiny 0.055 & 0.594 \tiny 0.044 & 0.792 \tiny 0.017 & 0.389 \tiny 0.064 & 0.436 \tiny 0.021 & 0.447 \tiny 0.021 & 0.514 \tiny 0.062 & 0.686 \tiny 0.017 & 0.170 \tiny 0.050 \\

GTP
& 0.545 \tiny 0.040 &	0.534 \tiny 0.035 &	0.573 \tiny 0.053 &	0.793 \tiny 0.025 &	0.372 \tiny 0.072 &0.474 \tiny 0.040& \textbf{0.507 \tiny 0.053} & 0.505 \tiny 0.038 & 0.673 \tiny 0.042&0.188 \tiny 0.057
\\

DSMIL
& 0.574 \tiny 0.012 &	0.569 \tiny 0.010 &	0.629 \tiny 0.032 &	0.823 \tiny 0.016 &	0.398 \tiny 0.032 & 0.469 \tiny 0.039	& 0.476 \tiny 0.049 &	0.518 \tiny 0.031	& 0.699 \tiny 0.020 & 0.182 \tiny 0.059

\\
RACR-MIL
& \textbf{0.585 \tiny 0.021} & \textbf{0.588 \tiny 0.014} & \textbf{0.643 \tiny 0.027} & \textbf{0.823 \tiny 0.020}	& \textbf{0.413 \tiny 0.045} & \textbf{0.510 \tiny 0.051} & \underline{0.505 \tiny 0.044} & \underline{0.575 \tiny 0.049} & 0.705 \tiny 0.036 & \textbf{0.257 \tiny 0.076}

\\
\bottomrule

\end{tabular}
}
\label{table:hnsc_grading}
\end{table*}

\textit{Head and Neck SCC}: Table~\ref{table:hnsc_grading} reports both in-distribution (SCC-DINO) and out-of-distribution (PathGen-CLIP) evaluation on head and neck SCC. RACR-MIL achieves the best F1 and MCC scores under both settings and shows a larger advantage under OOD shift where most strong MIL baselines degrade significantly. Interestingly, while GTP has higher precision in OOD evaluation, its recall is lower by 7\% compared to RACR-MIL. These strong results highlight our model's ability to learn generalizable contextual representations which are more robust to distribution shift. Overall performance on head and neck SCC is lower than for other anatomies, likely due to greater histologic heterogeneity and class imbalance. Approximately 50\% of cases are moderately differentiated, which is the most challenging grade class to separate from adjacent grades.

\begin{figure*}[h!]
\includegraphics[width=\textwidth]{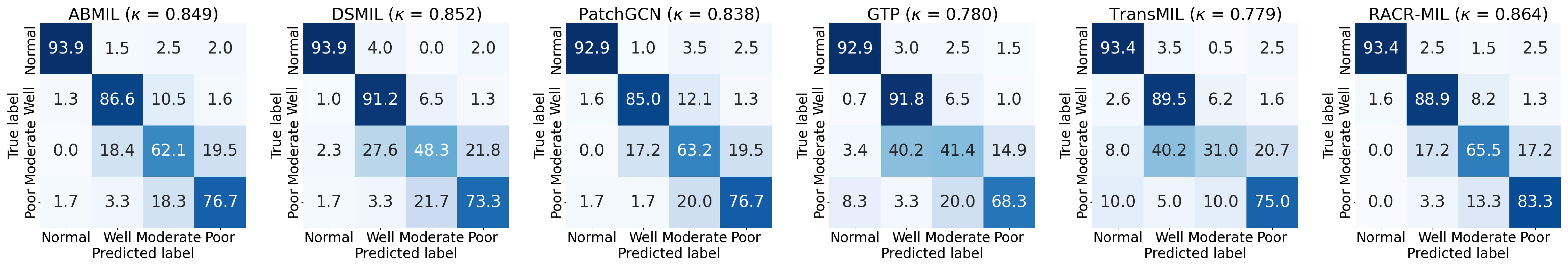}
 \caption{Confusion matrices for skin SCC: RACR-MIL achieves highest $\kappa$ score and classification accuracy on challenging worse-grade tumors.}
 \label{cf_cscc}
\end{figure*}

\subsubsection{Class-wise Performance}
The performance on each grade class for RACR-MIL and baseline models on skin SCC dataset is shown in Figure \ref{cf_cscc}. Our model consistently outperforms baselines across all tumor grades, achieving an average precision of $0.842$ and $\kappa$ score of $0.864$ (Figure \ref{pr_cscc}, Appendix), improving over existing contextual methods by up-to $8\%$. In particular, RACR-MIL achieves the highest accuracy for higher-risk moderate- and poorly-differentiated cases, with improvements of up to $28\%$ and $15\%$ over baselines, respectively. By incorporating ranking and cosine-softmax classifier, our approach remains robust to class imbalance, enhancing performance on the minority worse-grade classes while maintaining comparable accuracy on the dominant well-differentiated class. We also observe that transformer-based contextual methods tend to overfit on the dominant class at the expense of minority classes. Graph-based methods such as PatchGCN and RACR-MIL achieve more balanced performance, with our approach demonstrating superior overall results due to the improved dual graph formulation.

Figure \ref{cf_hnsc} highlights the class-wise performance of RACR-MIL and baseline methods on head and neck SCC. While RACR-MIL has a lower accuracy on the dominant `Moderate' class (53\% cases), it is robust to class imbalance and achieves a balanced performance across moderately- and poorly-differentiated cases. In contrast, existing contextual methods tend to overfit to the dominant classes, leading to lower $\kappa$ scores. Our model outperforms DSMIL on the more challenging worse-grade tumors while achieving a comparable $\kappa$ score. In comparison to ABMIL, RACR-MIL is less prone to overfitting and significantly improves on the minority well-differentiated class (12\% cases). 

On lung SCC (Figure \ref{cf_lscc}), RACR-MIL achieves the highest $\kappa$ score as well as grading accuracy on poorly-differentiated tumors compared to contextual methods. It significantly outperforms the next best contextual approach (PatchGCN) by more than $11\%$ on moderate-grade tumors. Our approach's balanced performance across higher-risk tumors as well as the highest average precision of its latent variant across tumor classes further underscores the robustness of our ranking formulation (Figure \ref{pr_lscc}, Appendix). We also observe that across skin and lung SCCs, ours is the only contextual approach that outperforms ABMIL on $\kappa$ score, achieving better accuracy on the challenging Moderately-differentiated cases. 
\subsection{Tumor Localization}
In addition to grade classification, we evaluate the tumor localization ability of RACR-MIL using both qualitative heatmap inspection and quantitative localization analysis. Following the experimental design of \cite{tourniaire2024whario}, we measure the overlap between model-identified tumor regions and the diagnostically relevant regions annotated by pathologists.

\begin{table*}[h!]
\caption{Localization accuracy of RACR-MIL compared to best performing non-contextual (ABMIL) and contextual (PatchGCN) baselines for clinically relevant tumor regions identified by three expert pathologists (skin SCC). Model was trained using SCC-DINO features. `Sens.' denotes Sensitivity and `Sal.' denotes Saliency.}
\label{iou}
\centering
\resizebox{\textwidth}{!}{
\begin{tabular}{|l|rrrr|rrrr|rrrr|}
\hline
\multirow{3}{*}{\textbf{Approach}} & \multicolumn{4}{c|}{\textbf{Pathologist 1}} & \multicolumn{4}{c|}{\textbf{Pathologist 2}} & \multicolumn{4}{c|}{\textbf{Pathologist 3}} \\ \cline{2-13} 
 & \multicolumn{2}{c}{\textbf{Correct Grade}} & \multicolumn{2}{c|}{\textbf{Incorrect Grade}} & \multicolumn{2}{c}{\textbf{Correct Grade}} & \multicolumn{2}{c}{\textbf{Incorrect Grade}} & \multicolumn{2}{|c}{\textbf{Correct Grade}} & \multicolumn{2}{c|}{\textbf{Incorrect Grade}} \\ \cline{2-13} 
& \multicolumn{1}{c}{\textbf{Sens.}} 
& \multicolumn{1}{c}{\textbf{Sal.}} 
& \multicolumn{1}{c}{\textbf{Sens.}} 
& \multicolumn{1}{c|}{\textbf{Sal.}} 
& \multicolumn{1}{c}{\textbf{Sens.}} 
& \multicolumn{1}{c}{\textbf{Sal.}} 
& \multicolumn{1}{c}{\textbf{Sens.}} 
& \multicolumn{1}{c}{\textbf{Sal.}} 
& \multicolumn{1}{|c}{\textbf{Sens.}} 
& \multicolumn{1}{c}{\textbf{Sal.}} 
& \multicolumn{1}{c}{\textbf{Sens.}} 
& \multicolumn{1}{c|}{\textbf{Sal.}} \\ \hline
ABMIL & 
0.557 & 0.539 & 0.564 & 0.441 & 
0.680 & 0.597 & 0.541 & 0.476 & 
0.785 & \textbf{0.645} & 0.691 & 0.575\\ \hline

PatchGCN & 
0.525 & 0.507 & 0.446 & 0.449 &
0.684 & 0.606 & 0.451 & 0.444 & 
0.686 & 0.623 & 0.285 & 0.289 \\ \hline

\begin{tabular}[c]{@{}c@{}}RACR-MIL\\(no ranking)\end{tabular} & 
0.455 & 0.531 & 0.358 & 0.424 & 
0.553 & 0.575 & 0.494 & 0.506 & 
0.450 & 0.566 & 0.434 & 0.426 \\ \hline

\textbf{RACR-MIL} & 
\textbf{0.604} & \textbf{0.540} & 0.543 & 0.428 & 
\textbf{0.774} & \textbf{0.656} & 0.618 & 0.466 &
\textbf{0.791} & 0.637 & 0.867 & 0.604  \\ \hline
\end{tabular}
}
\end{table*}

\subsubsection{Improved coverage of relevant tumor regions}
A random sample of 19 WSIs was selected for review and annotation by three senior pathologists comprising 5 well-, 8 moderately-, and 6 poorly-differentiated cases. These WSIs formed the test set for localization analysis, while the remaining cases were used to train RACR-MIL using the optimal hyperparameters for SCC-DINO feature extractor (Table \ref{tab:hyperparameters}). Each pathologist reviewed all 19 WSIs and annotated up to seven tumor regions per WSI that were most relevant for grade diagnosis. Each annotated region was assigned a grade by the clinician based on its dominant histologic feature.

The annotations were then used to evaluate the overlap with the model's top-predicted regions of interest (ROIs). To reconcile the resolution difference between pixel-level annotations and patch-level model outputs, overlap was computed at the pixel level. Model ROIs were defined from patch-level predictions by first selecting the most important patches according to the attention weights. Following prior work (\cite{sun2025label}), we selected the top 25\% of patches in each WSI based on attention scores $w_n$, where the 25\% threshold was defined relative to the total number of tissue patches in that WSI. This threshold is also supported empirically, as pathologist-annotated regions occupied a median of 25.5\% of tissue area across 19 cases (Table \ref{tab:case_comparison}). Among the selected patches, we retain only those with grade-specific probability $p_n^c > 0.5$ as positive grade-relevant patches. Contiguous positive patches are grouped into ROIs, and dilated using a $3 \times 3$ kernel to generate a smooth binary model mask at $5X$ resolution. Pathologist annotations are similarly downsampled to $5X$ resolution to generate binary reference masks. Tumor localization is then quantified using pixel-level intersection-over-union between the model and annotation masks. 
Since the annotations captured diagnostically relevant regions rather than the full tumor extent, we used \emph{Sensitivity} as a primary metric to assess the model's ability to localize critical tumor regions. Sensitivity was computed on a per-WSI basis as the average overlap proportion between annotated regions and model's ROIs. Model's overall localization performance was computed by averaging sensitivity across slides, thus avoiding overweighing WSIs with more annotations. We also computed \emph{Saliency}, defined as the average model confidence (patch-level predicted class probability) within each annotated region, to assess the model's prediction confidence. Both sensitivity and saliency were further stratified into \emph{Correct Grade} and \emph{Incorrect Grade} groups based on whether the predicted ROI grade matched the annotated region grade for each WSI. To reduce instability due to very small regions, localization analysis only included annotations with a minimum total area equivalent to any set of 9 patches of size $448 \times 448$ at $20X$ resolution, regardless of their spatial arrangement. As shown in Table \ref{iou}, RACR-MIL consistently achieves higher sensitivity than ABMIL and PatchGCN across all three pathologists, indicating better coverage of diagnostically relevant tumor regions. RACR-MIL also shows higher saliency for correctly classified regions for two of the three pathologists, suggesting more confident and reliable prioritization when predictions are accurate. Moreover, for incorrectly classified regions, RACR-MIL attains the highest sensitivity for two pathologists, further supporting its superior tumor localization performance. Additionally, incorporating the ranking loss substantially improves localization performance across all three pathologists, highlighting its role in prioritizing diagnostically relevant worse-grade regions.

\begin{figure}[t!]
\centering
\includegraphics[width=0.9\linewidth]{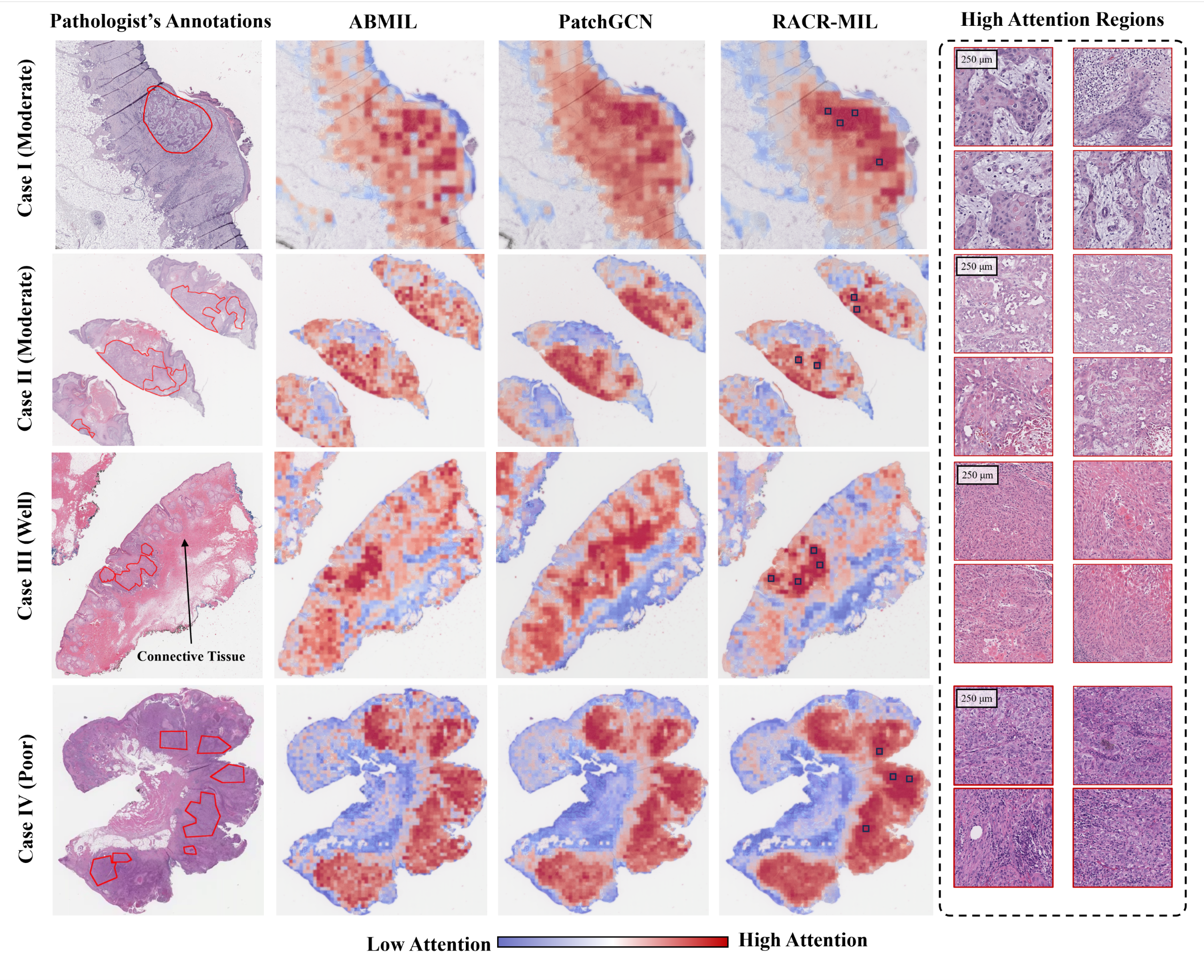}
\caption{Normalized attention heatmaps highlighting the tumor ROIs prioritized by different models. Patch-level tumor grade annotations are highlighted with a `Red' outline in the leftmost panel. In the heatmaps, 'Red' and Blue indicate high and low attention regions, respectively.}
\label{diff_heatmap}
\end{figure}

\subsubsection{Improved alignment with pathologist's annotations} \label{qualitative_analysis}
We highlight the impact of the proposed innovations on localization of diagnostically important tumor regions by examining normalized attention heatmaps from four representative WSIs (Figure \ref{diff_heatmap}). These qualitative examples were selected to illustrate common differences in region prioritization across models. They are intended to show how each model prioritizes pathologist-identified, grading-relevant tumor regions, rather than to delineate the complete tumor tissue in each WSI. For visualization, patch-level attention weights $w_n$ were normalized independently within each WSI as $a_n = (w_n - w_n^{min})/(w_n^{max} - w_n^{min})$. Hence, the heatmaps reflect relative prioritization within a WSI rather than absolute importance.
Overall, RACR-MIL shows closer qualitative agreement with pathologist-annotated diagnostically relevant tumor regions than the two best-performing baselines on skin SCC. In Cases I, II and IV, the dual graph design allows RACR-MIL to capture a larger portion of the relevant tumor while avoiding the overly diffused attention pattern observed in PatchGCN. In Case III, the combination of contextual modeling and ranking reduces false positives by assigning higher importance to the tumor region compared to adjacent normal tissues. These examples further illustrate the value of combining ranking with contextual features, as PatchGCN assigns similar importance to connective tissue and tumor (Case III) while ABMIL overlooks important tumor subregions in Cases I and II. PatchGCN also produces more diffuse attention maps, particularly in Case I, consistent with oversmoothing and reduced sensitivity to fine-grained morphological differences. Case IV further shows that RACR-MIL assigns consistent high attention to poorly-differentiated regions. Figure \ref{diff_heatmap} also highlights representative top-attended patches, showing that RACR-MIL prioritizes diagnostically relevant regions. In Case IV, the top-attended patches show marked pleomorphism and lymphocytic infiltration, consistent with high-grade SCC while in Case III, the top-attended patches show keratinization and distinct cell borders, which are characteristic of lower-grade SCC.

\begin{figure}[t!]
\centering
\includegraphics[width=0.9\linewidth]{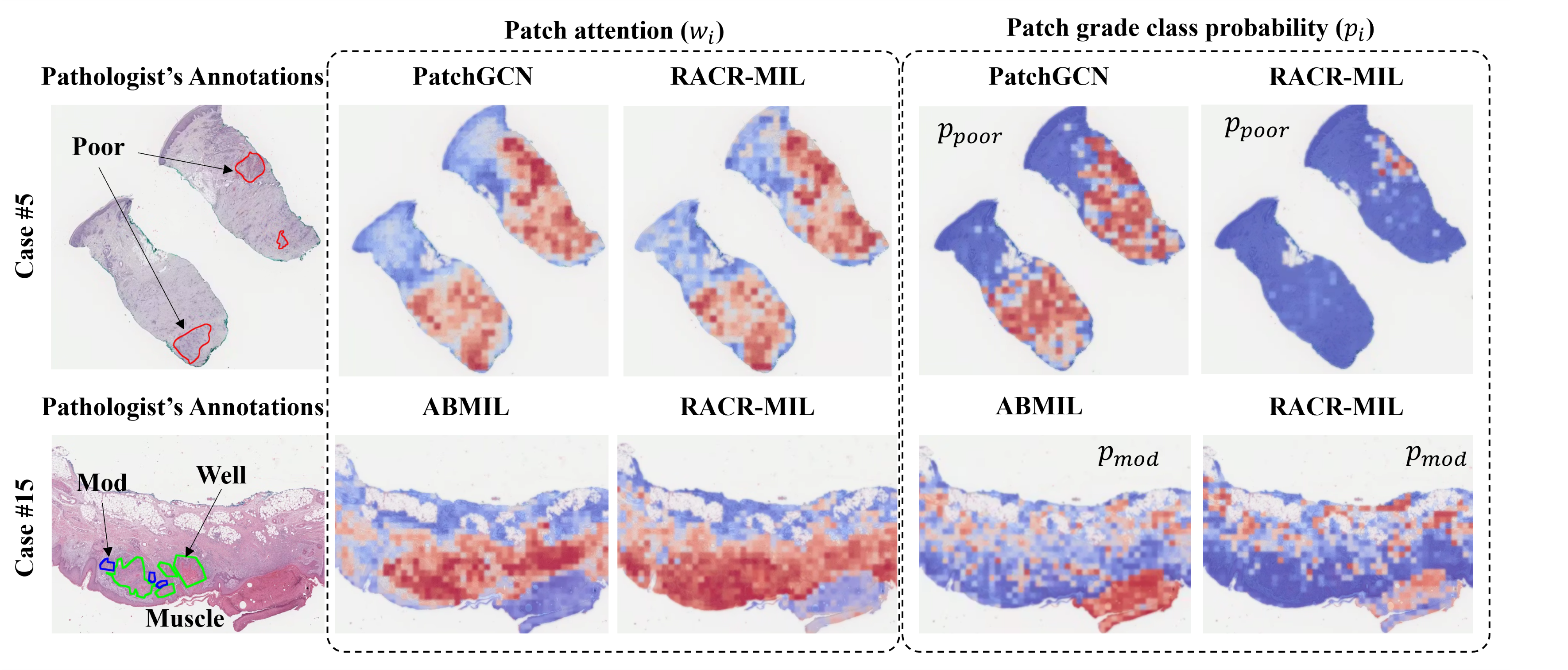}
\caption{Cases highlighting examples where ABMIL and PatchGCN perform better than RACR-MIL in localizing the highest-grade tumor.}
\label{failure_heatmap}
\end{figure}

\subsubsection{Failure Cases}
We further examine cases in which RACR-MIL underperforms relative to ABMIL or PatchGCN. Table \ref{tab:case_comparison} (Appendix) summarizes the grading outcomes across all 19 WSIs. RACR-MIL disagreed with the majority-consensus slide-level grade in three cases (Cases 5, 9, and 18). In four additional cases (Cases 2, 11, 12, 15), it failed to capture the highest-grade region annotated by at least one pathologist, although its slide-level prediction remained consistent with the majority consensus. Case 18 was clinically ambiguous, as RACR-MIL predicted poor differentiation in agreement with the regional assessments of two pathologists despite a moderate consensus slide-level grade.
Figure \ref{failure_heatmap} presents two representative failure cases. In Case 5, RACR-MIL and PatchGCN showed similar attention patterns, indicating comparable prioritization of tumor regions. However, RACR-MIL underestimated the extent of poorly-differentiated tumor, assigning high poor-grade probability to a subset of the lesion while classifying most of the tumor as moderately-differentiated. This might be due to the patch selection strategy in RACR-MIL, wherein only high-confidence patches contribute to the ranking objective, potentially limiting supervision for ambiguous or boundary regions. In Case 15, RACR-MIL failed to identify a small moderately-differentiated tumor region as evident in the probability heatmap, likely because it was spatially limited and surrounded by well-differentiated tumor making it susceptible to contextual smoothing. However, unlike ABMIL, RACR-MIL did not spuriously assign moderate grade predictions to adjacent muscle tissue, suggesting that the ranking constraint helps suppress such false positives.

\subsection{Ablation Study}

\begin{table*}[!t]
\setlength{\tabcolsep}{6pt}
\centering
\caption{Impact of ranking constraints, diversity loss, and dual graph representations on SCC grading performance. Results are reported as mean $\pm$ standard deviation across five folds. Values in parentheses denote absolute changes in the mean relative to the corresponding graph-only baseline: $G_{spa}$, $G_{lat}$, and $G_{dual}$. Ranking improves F1 by 1–2\% across all SCC cohorts while dual-graph enhances performance over spatial-only and latent-only graphs in majority cases.}
\label{semantic_spatial}
\resizebox{\textwidth}{!}{
\begin{tabular}{llcccccccccc}
\toprule

\multirow{2}{*}{\begin{tabular}[c]{@{}c@{}}\textbf{Graph}\\ \textbf{Variant}\end{tabular}} &
\multirow{2}{*}{\textbf{Feature}} &
\multicolumn{3}{c}{\textbf{Skin SCC}} &
\multicolumn{3}{c}{\textbf{Head \& Neck SCC}} &
\multicolumn{3}{c}{\textbf{Lung SCC}} \\

\\

\cmidrule(lr){3-5}
\cmidrule(lr){6-8}
\cmidrule(lr){9-11}

& &
\textbf{F1} & \textbf{MCC} & \textbf{$\kappa$} &
\textbf{F1} & \textbf{MCC} & \textbf{$\kappa$} &
\textbf{F1} & \textbf{MCC} & \textbf{$\kappa$}

\\
\midrule






\multirow{2}{*}{$G_{\mathrm{spa}}$}
& SCC-DINO
& 0.787 \tiny 0.040 & 0.792 \tiny 0.042 & 0.860 \tiny 0.033
& 0.554 {\tiny 0.032} & 0.373 {\tiny 0.032} & 0.660 \tiny 0.039
& 0.762 \tiny 0.072 & 0.735 \tiny 0.088 & 0.855 \tiny 0.053
\\
& PathGen
& 0.757 {\tiny 0.029} & 0.754 {\tiny 0.042} & 0.850 {\tiny 0.029}
& 0.472 {\tiny 0.030} & 0.212 {\tiny 0.053} & 0.330 {\tiny 0.033}
& 0.735 {\tiny 0.070} & 0.715 {\tiny 0.075} & 0.861 {\tiny 0.026}
\\

\midrule
\multirow{2}{*}{$G_{\mathrm{lat}}$}
& SCC-DINO
& 0.790 {\tiny 0.014} & 0.795 {\tiny 0.030} & 0.862 {\tiny 0.023}
& 0.561 {\tiny 0.014} & 0.378 {\tiny 0.034} & 0.652 {\tiny 0.027}
& 0.792 {\tiny 0.076} & 0.771 {\tiny 0.087} & 0.877 {\tiny 0.042}
\\
& PathGen
& 0.765 {\tiny 0.035} & 0.762 {\tiny 0.036} & 0.847 {\tiny 0.033}
& 0.472 {\tiny 0.040} & 0.208 {\tiny 0.063} & 0.331 {\tiny 0.039}
& 0.767 {\tiny 0.065} & 0.752 {\tiny 0.069} & 0.871 {\tiny 0.027}
\\

\midrule

\multirow{2}{*}{$G_{\mathrm{spa}} + L_{\mathrm{rank}}$}
& SCC-DINO
& \begin{tabular}{c}0.795 \tiny 0.027\\ \scriptsize (\textcolor{blue}{+0.008})\end{tabular}
& \begin{tabular}{c}0.795 \tiny 0.032\\ \scriptsize (\textcolor{blue}{+0.003})\end{tabular}
& \begin{tabular}{c}0.852 \tiny 0.044\\ \scriptsize (\textcolor{red}{-0.008})\end{tabular}
& \begin{tabular}{c}0.569 \tiny 0.043\\ \scriptsize (\textcolor{blue}{+0.015})\end{tabular}
& \begin{tabular}{c}0.396 \tiny 0.064\\ \scriptsize (\textcolor{blue}{+0.023})\end{tabular}
& \begin{tabular}{c}0.679 \tiny 0.025\\ \scriptsize (\textcolor{blue}{+0.019})\end{tabular}
& \begin{tabular}{c}0.792 \tiny 0.060\\ \scriptsize (\textcolor{blue}{+0.030})\end{tabular}
& \begin{tabular}{c}0.773 \tiny 0.067\\ \scriptsize (\textcolor{blue}{+0.038})\end{tabular}
& \begin{tabular}{c}0.879 \tiny 0.031\\ \scriptsize (\textcolor{blue}{+0.024})\end{tabular}
\\

& PathGen
& \begin{tabular}{c}0.776 \tiny 0.027\\ \scriptsize (\textcolor{blue}{+0.019})\end{tabular}
& \begin{tabular}{c}0.766 \tiny 0.035\\ \scriptsize (\textcolor{blue}{+0.012})\end{tabular}
& \begin{tabular}{c}0.857 \tiny 0.028\\ \scriptsize (\textcolor{blue}{+0.007})\end{tabular}
& \begin{tabular}{c}0.476 \tiny 0.045\\ \scriptsize (\textcolor{blue}{+0.004})\end{tabular}
& \begin{tabular}{c}0.216 \tiny 0.071\\ \scriptsize (\textcolor{blue}{+0.004})\end{tabular}
& \begin{tabular}{c}0.342 \tiny 0.040\\ \scriptsize (\textcolor{blue}{+0.012})\end{tabular}
& \begin{tabular}{c}0.759 \tiny 0.081\\ \scriptsize (\textcolor{blue}{+0.024})\end{tabular}
& \begin{tabular}{c}0.743 \tiny 0.084\\ \scriptsize (\textcolor{blue}{+0.032})\end{tabular}
& \begin{tabular}{c}0.875 \tiny 0.031\\ \scriptsize (\textcolor{blue}{+0.014})\end{tabular}
\\

\midrule

\multirow{2}{*}{$G_{\mathrm{lat}} + L_{\mathrm{rank}}$}
& SCC-DINO
& \begin{tabular}{c}0.794 {\tiny 0.020}\\ \scriptsize \textcolor{blue}{(+0.004)}\end{tabular}
& \begin{tabular}{c}0.797 {\tiny 0.028}\\ \scriptsize \textcolor{blue}{(+0.002)}\end{tabular}
& \begin{tabular}{c}\textbf{0.873} {\tiny 0.035}\\ \scriptsize \textcolor{blue}{(+0.011)}\end{tabular}
& \begin{tabular}{c}0.575 {\tiny 0.013}\\ \scriptsize \textcolor{blue}{(+0.014)}\end{tabular}
& \begin{tabular}{c}0.393 {\tiny 0.039}\\ \scriptsize \textcolor{blue}{(+0.015)}\end{tabular}
& \begin{tabular}{c}0.698 {\tiny 0.040}\\ \scriptsize \textcolor{blue}{(+0.046)}\end{tabular}
& \begin{tabular}{c}\textbf{0.800 \tiny 0.068}\\ \scriptsize \textcolor{blue}{(+0.008)}\end{tabular}
& \begin{tabular}{c}\textbf{0.779 \tiny 0.078}\\ \scriptsize \textcolor{blue}{(+0.008)}\end{tabular}
& \begin{tabular}{c}\textbf{0.883 \tiny 0.039}\\ \scriptsize \textcolor{blue}{(+0.006)}\end{tabular}
\\
& PathGen
& \begin{tabular}{c}0.776 {\tiny 0.022}\\ \scriptsize \textcolor{blue}{(+0.011)}\end{tabular}
& \begin{tabular}{c}0.767 {\tiny 0.019}\\ \scriptsize \textcolor{blue}{(+0.005)}\end{tabular}
& \begin{tabular}{c}0.858 {\tiny 0.022}\\ \scriptsize \textcolor{blue}{(+0.011)}\end{tabular}
& \begin{tabular}{c}0.475 {\tiny 0.033}\\ \scriptsize \textcolor{blue}{(+0.003)}\end{tabular}
& \begin{tabular}{c}0.205 {\tiny 0.058}\\ \scriptsize \textcolor{red}{(-0.003)}\end{tabular}
& \begin{tabular}{c}0.330 {\tiny 0.045}\\ \scriptsize \textcolor{red}{(-0.001)}\end{tabular}
& \begin{tabular}{c}0.775 {\tiny 0.068}\\ \scriptsize \textcolor{blue}{(+0.008)}\end{tabular}
& \begin{tabular}{c}0.760 {\tiny 0.073}\\ \scriptsize \textcolor{blue}{(+0.008)}\end{tabular}
& \begin{tabular}{c}0.878 {\tiny 0.025}\\ \scriptsize \textcolor{blue}{(+0.007)}\end{tabular}
\\
\midrule
\midrule

\multirow{2}{*}{$G_{\mathrm{dual}}$}
& SCC-DINO
& 0.788 {\tiny 0.019} & 0.788 {\tiny 0.019} & 0.862 {\tiny 0.032}
& 0.576 {\tiny 0.044} & 0.386 {\tiny 0.057} & 0.699 {\tiny 0.043}
& 0.775 {\tiny 0.065} & 0.754 {\tiny 0.086} & 0.851 {\tiny 0.076}
\\
& PathGen
& 0.785 {\tiny 0.025} & 0.785 {\tiny 0.026} & 0.866 {\tiny 0.028}
& 0.503 {\tiny 0.048} & 0.247 {\tiny 0.074} & 0.377 {\tiny 0.064}
& 0.775 {\tiny 0.058} & 0.765 {\tiny 0.054} & 0.878 {\tiny 0.025}
\\

\midrule

\multirow{2}{*}{$G_{\mathrm{dual}} + L_{\mathrm{rank}}$}
& SCC-DINO
& \begin{tabular}{c}0.800 {\tiny 0.026}\\ \scriptsize \textcolor{blue}{(+0.012)}\end{tabular}
& \begin{tabular}{c}0.797 {\tiny 0.034}\\ \scriptsize \textcolor{blue}{(+0.009)}\end{tabular}
& \begin{tabular}{c}0.869 {\tiny 0.025}\\ \scriptsize \textcolor{blue}{(+0.007)}\end{tabular}
& \begin{tabular}{c}0.579 {\tiny 0.032}\\ \scriptsize \textcolor{blue}{(+0.003)}\end{tabular}
& \begin{tabular}{c}0.396 {\tiny 0.046}\\ \scriptsize \textcolor{blue}{(+0.010)}\end{tabular}
& \begin{tabular}{c}\textbf{0.713} {\tiny 0.035}\\ \scriptsize \textcolor{blue}{(+0.014)}\end{tabular}
& \begin{tabular}{c}0.790 {\tiny 0.078}\\ \scriptsize \textcolor{blue}{(+0.015)}\end{tabular}
& \begin{tabular}{c}0.771 {\tiny 0.098}\\ \scriptsize \textcolor{blue}{(+0.017)}\end{tabular}
& \begin{tabular}{c}0.876 {\tiny 0.051}\\ \scriptsize \textcolor{blue}{(+0.025)}\end{tabular}
\\
& PathGen
& \begin{tabular}{c}0.796 {\tiny 0.035}\\ \scriptsize \textcolor{blue}{(+0.011)}\end{tabular}
& \begin{tabular}{c}0.786 {\tiny 0.035}\\ \scriptsize \textcolor{blue}{(+0.001)}\end{tabular}
& \begin{tabular}{c}0.872 {\tiny 0.027}\\ \scriptsize \textcolor{blue}{(+0.006)}\end{tabular}
& \begin{tabular}{c}0.508 {\tiny 0.052}\\ \scriptsize \textcolor{blue}{(+0.005)}\end{tabular}
& \begin{tabular}{c}0.253 {\tiny 0.079}\\ \scriptsize \textcolor{blue}{(+0.006)}\end{tabular}
& \begin{tabular}{c}0.389 {\tiny 0.055}\\ \scriptsize \textcolor{blue}{(+0.012)}\end{tabular}
& \begin{tabular}{c}0.779 {\tiny 0.055}\\ \scriptsize \textcolor{blue}{(+0.004)}\end{tabular}
& \begin{tabular}{c}0.769 {\tiny 0.054}\\ \scriptsize \textcolor{blue}{(+0.004)}\end{tabular}
& \begin{tabular}{c}0.871 {\tiny 0.039}\\ \scriptsize \textcolor{red}{(-0.007)}\end{tabular}
\\
\midrule
\multirow{2}{*}{$G_{dual} + L_{\mathrm{div}}$}
& SCC-DINO
& \begin{tabular}{c}0.793 {\tiny 0.023}\\ \scriptsize \textcolor{blue}{(+0.005)}\end{tabular}
& \begin{tabular}{c}0.793 {\tiny 0.024}\\ \scriptsize \textcolor{blue}{(+0.005)}\end{tabular}
& \begin{tabular}{c}0.864 {\tiny 0.030}\\ \scriptsize \textcolor{blue}{(+0.002)}\end{tabular}
& \begin{tabular}{c}0.575 {\tiny 0.044}\\ \scriptsize \textcolor{red}{(-0.001)}\end{tabular}
& \begin{tabular}{c}0.386 {\tiny 0.056}\\ \scriptsize \textcolor{blue}{(+0.000)}\end{tabular}
& \begin{tabular}{c}0.699 {\tiny 0.045}\\ \scriptsize \textcolor{blue}{(+0.000)}\end{tabular}
& \begin{tabular}{c}0.775 {\tiny 0.073}\\ \scriptsize \textcolor{blue}{(+0.000)}\end{tabular}
& \begin{tabular}{c}0.754 {\tiny 0.096}\\ \scriptsize \textcolor{blue}{(+0.000)}\end{tabular}
& \begin{tabular}{c}0.851 {\tiny 0.084}\\ \scriptsize \textcolor{blue}{(+0.000)}\end{tabular}
\\
& PathGen
& \begin{tabular}{c}0.790 {\tiny 0.013}\\ \scriptsize \textcolor{blue}{(+0.005)}\end{tabular}
& \begin{tabular}{c}0.787 {\tiny 0.009}\\ \scriptsize \textcolor{blue}{(+0.002)}\end{tabular}
& \begin{tabular}{c}0.876 {\tiny 0.024}\\ \scriptsize \textcolor{blue}{(+0.010)}\end{tabular}
& \begin{tabular}{c}0.506 {\tiny 0.050}\\ \scriptsize \textcolor{blue}{(+0.003)}\end{tabular}
& \begin{tabular}{c}0.253 {\tiny 0.075}\\ \scriptsize \textcolor{blue}{(+0.006)}\end{tabular}
& \begin{tabular}{c}0.382 {\tiny 0.061}\\ \scriptsize \textcolor{blue}{(+0.005)}\end{tabular}
& \begin{tabular}{c}0.775 {\tiny 0.058}\\ \scriptsize \textcolor{blue}{(+0.000)}\end{tabular}
& \begin{tabular}{c}0.765 {\tiny 0.054}\\ \scriptsize \textcolor{blue}{(+0.000)}\end{tabular}
& \begin{tabular}{c}0.878 {\tiny 0.025}\\ \scriptsize \textcolor{blue}{(+0.000)}\end{tabular}
\\
\midrule

\multirow{2}{*}{\begin{tabular}[c]{@{}c@{}}
$G_{\mathrm{dual}} + L_{\mathrm{div}}$ \\
$+\, L_{\mathrm{rank}}$
\end{tabular}} 
& SCC-DINO
& \begin{tabular}{c}\textbf{0.807} {\tiny 0.016}\\ \scriptsize \textcolor{blue}{(+0.019)}\end{tabular}
& \begin{tabular}{c}\textbf{0.801} {\tiny 0.032}\\ \scriptsize \textcolor{blue}{(+0.013)}\end{tabular}
& \begin{tabular}{c}0.864 {\tiny 0.031}\\ \scriptsize \textcolor{blue}{(+0.002)}\end{tabular}
& \begin{tabular}{c}\textbf{0.585} {\tiny 0.018}\\ \scriptsize \textcolor{blue}{(+0.009)}\end{tabular}
& \begin{tabular}{c}\textbf{0.413} {\tiny 0.040}\\ \scriptsize \textcolor{blue}{(+0.027)}\end{tabular}
& \begin{tabular}{c}0.705 {\tiny 0.021}\\ \scriptsize \textcolor{blue}{(+0.006)}\end{tabular}
& \begin{tabular}{c}0.790 {\tiny 0.070}\\ \scriptsize \textcolor{blue}{(+0.015)}\end{tabular}
& \begin{tabular}{c}0.771 {\tiny 0.088}\\ \scriptsize \textcolor{blue}{(+0.017)}\end{tabular}
& \begin{tabular}{c}0.877 {\tiny 0.046}\\ \scriptsize \textcolor{blue}{(+0.026)}\end{tabular}
\\

& PathGen
& \begin{tabular}{c}\textbf{0.805} {\tiny 0.025}\\ \scriptsize \textcolor{blue}{(+0.020)}\end{tabular}
& \begin{tabular}{c}\textbf{0.797} {\tiny 0.027}\\ \scriptsize \textcolor{blue}{(+0.012)}\end{tabular}
& \begin{tabular}{c}\textbf{0.885} {\tiny 0.025}\\ \scriptsize \textcolor{blue}{(+0.019)}\end{tabular}
& \begin{tabular}{c}\textbf{0.510} {\tiny 0.051}\\ \scriptsize \textcolor{blue}{(+0.007)}\end{tabular}
& \begin{tabular}{c}\textbf{0.257} {\tiny 0.076}\\ \scriptsize \textcolor{blue}{(+0.010)}\end{tabular}
& \begin{tabular}{c}\textbf{0.393} {\tiny 0.053}\\ \scriptsize \textcolor{blue}{(+0.016)}\end{tabular}
& \begin{tabular}{c}\textbf{0.786} {\tiny 0.049}\\ \scriptsize \textcolor{blue}{(+0.011)}\end{tabular}
& \begin{tabular}{c}\textbf{0.775} {\tiny 0.045}\\ \scriptsize \textcolor{blue}{(+0.010)}\end{tabular}
& \begin{tabular}{c}\textbf{0.873} {\tiny 0.040}\\ \scriptsize \textcolor{red}{(-0.005)}\end{tabular}
\\

\bottomrule
\end{tabular}
}
\end{table*}

\begin{table}[t!]
\centering
\caption{Impact of inter- and intra-grade ranking on skin SCC grading accuracy using dual-graph with diversity regularization.}
\label{tab:dual_rank_skin}
\resizebox{0.7\textwidth}{!}{
\begin{tabular}{lccc|ccc}
\toprule
\multirow{2}{*}{\textbf{Graph Variant}} &
\multicolumn{3}{c|}{\textbf{SCC-DINO}} &
\multicolumn{3}{c}{\textbf{PathGen-CLIP}} \\
\cmidrule(lr){2-4} \cmidrule(lr){5-7}
& \textbf{F1} & \textbf{MCC} & \textbf{$\kappa$}
& \textbf{F1} & \textbf{MCC} & \textbf{$\kappa$} \\
\midrule

$G_{\mathrm{dual}} + L_{\mathrm{div}}$
& 0.793 \tiny 0.023 & 0.793 \tiny 0.025 & 0.864 \tiny 0.030 
& 0.790 \tiny 0.013 & 0.787 \tiny 0.009 & 0.876 \tiny 0.024
\\
$G_{\mathrm{dual}} + L_{\mathrm{div}} + L_{\mathrm{inter}}$
& 0.802 \tiny 0.019 & 0.799 \tiny 0.035 & 0.873 \tiny 0.030 
& 0.797 \tiny 0.028 & 0.796 \tiny 0.033 & 0.876 \tiny 0.031
\\
$G_{\mathrm{dual}} + L_{\mathrm{div}} + L_{\mathrm{intra}}$
& 0.785 \tiny 0.022 & 0.778 \tiny 0.036	& 0.851 \tiny 0.027
& 0.794 \tiny 0.025 & 0.788 \tiny 0.029 & 0.886 \tiny 0.028
\\
$G_{\mathrm{dual}} + L_{\mathrm{div}} + L_{\mathrm{inter}} + L_{\mathrm{intra}}$
& 0.807 \tiny 0.016 & 0.801 \tiny 0.032 & 0.864 \tiny 0.031
& 0.805 \tiny 0.025 & 0.797 \tiny 0.027 & 0.885 \tiny 0.025 \\

\bottomrule
\end{tabular}}

\end{table}

All innovations in our framework contribute to improved grading performance. We evaluated eight model variants: 1) graph only architectures without rank-constrained attention, including latent graph only ($G_{lat}$), spatial graph only ($G_{spa}$) and latent + spatial graphs ($G_{dual}$); 2) these same graph variants with ranking constraints: $G_{lat} + \mathcal{L}_{rank}, G_{spa} + \mathcal{L}_{rank}, G_{dual} + \mathcal{L}_{rank}$; 3) dual graph with both ranking constraints and diversity regularization: $G_{dual}$ + $\mathcal{L}_{rank}$ + $\mathcal{L}_{div}$; 4) dual graph with diversity regularization only: $G_{dual}$ + $\mathcal{L}_{div}$ . All variants were evaluated using both SCC-DINO and PathGen-CLIP feature extractors across all SCC datasets, and the results are reported in Table~\ref{semantic_spatial}. With SCC-DINO features, the dual-graph model with ranking and diversity regularization achieves the largest F1 improvements in most settings, except on lung SCC. With PathGen-CLIP feature backbone, each of our contributions improves performance across the three datasets. 

\subsubsection{Ablation Study of Graph Components}

\begin{figure*}[!h]
\centering
\includegraphics[width=\linewidth]{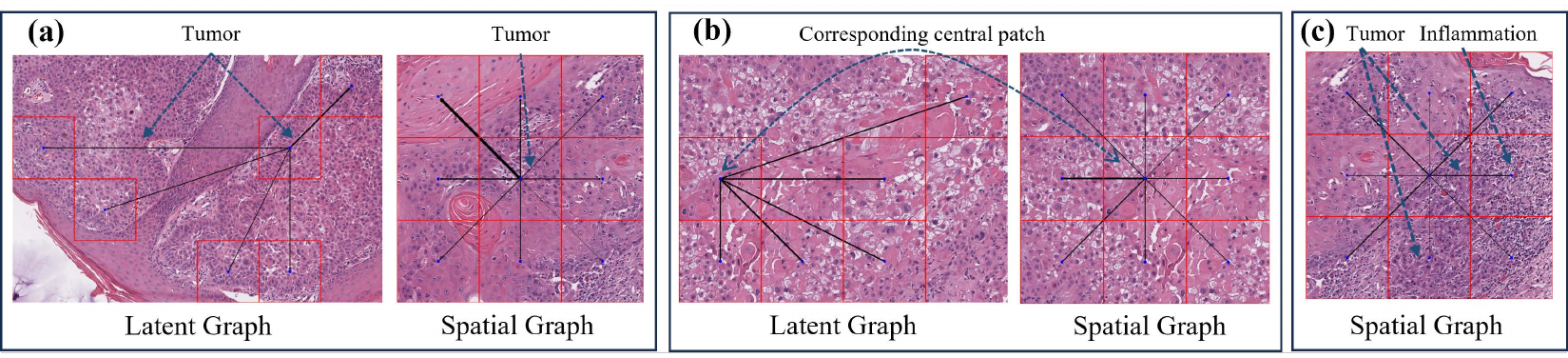}
\caption{Contextual relationships learned by RACR-MIL. Edge thickness indicates the learned self-attention weight (thicker = higher attention). (a) The latent graph emphasizes morphologically similar tumor regions (including non-adjacent areas), while the spatial graph assigns higher weight to non-cellular keratinization region. (b) Even in a relatively homogeneous tumor–stroma region, the spatial graph prioritizes heterogeneous local neighbors, whereas the latent graph links regions comprising tumor and collagen. (c) The spatial graph selectively highlights inflammation-associated neighbors around the central tumor patch, capturing microenvironment heterogeneity}
\label{ablation_graph}
\end{figure*}

We find that the dual-graph formulation generally outperforms single-graph variants, particularly when combined with ranking constraints (Table \ref{semantic_spatial}). The increase in performance is most consistent for head and neck SCC, where combining spatial and latent graphs improves F1 by 1-3\%. On skin SCC, the dual graph also improves performance, with the largest gains observed when diversity regularization is added. In particular, combining the diversity regularization with the dual graph improves classification performance by 3\% for moderately- and poorly-differentiated classes (Figure \ref{cf_racr}, Appendix). With PathGen-CLIP features, the dual graph is consistently stronger and outperforms both $G_{lat}$ and $G_{spa}$ across all SCC cohorts. These results support our claim that the two graphs provide complementary information rather than redundant structure.
We also visualize the learned self-attention edge weights in the latent and spatial graphs in Figure \ref{ablation_graph} to better interpret the contribution of each graph branch. In Figure \ref{ablation_graph}A, the latent graph connects morphologically similar regions including distant tumor areas, while the spatial graph emphasizes heterogeneous local context by assigning higher edge weights to keratin near the tumor boundary. In Figure \ref{ablation_graph}B, where the central patch lies in a relatively homogeneous tumor-stroma region, the spatial graph prioritizes heterogeneous neighbors and assigns higher edge attention to neighboring patches with higher tumor burden and lower eosinophilic content (on the left). The latent graph connects the central patch with non-adjacent regions having similar tumor and collagen composition. Figure \ref{ablation_graph}C shows the spatial graph emphasizing the tumor-inflammation interface while down-weighing a nearby tumor patch, suggesting sensitivity to tumor microenvironment heterogeneity. Together, these examples are qualitatively consistent with the latent and spatial graphs learning complementary relationships and capturing semantic similarity along with local tumor microenvironment context.

\subsubsection{Ablation Study of Ranking Module}
\begin{figure*}[!t]
\centering
\includegraphics[width=\textwidth]{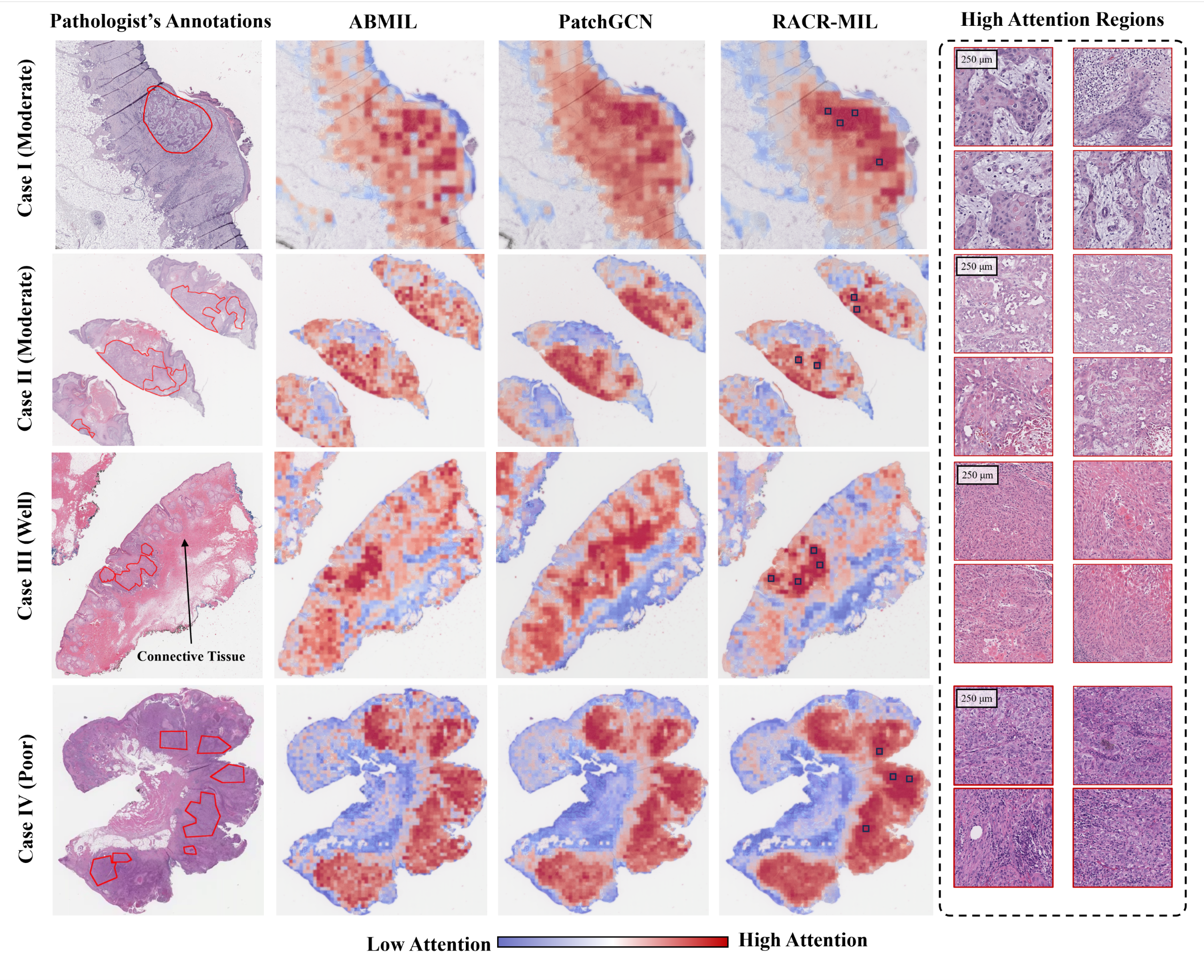}
\caption{Attention heatmaps highlighting the impact of inter-grade ranking in assigning higher importance to tumor regions with worse grades (Pathologist Annotations - Red: Poorly-differentiated, Blue: Moderately-differentiated, Green: Well-differentiated)}
\label{fig_rank_heatmap}
\end{figure*}

\begin{figure*}[!t]
\centering
\includegraphics[width=0.9\textwidth]{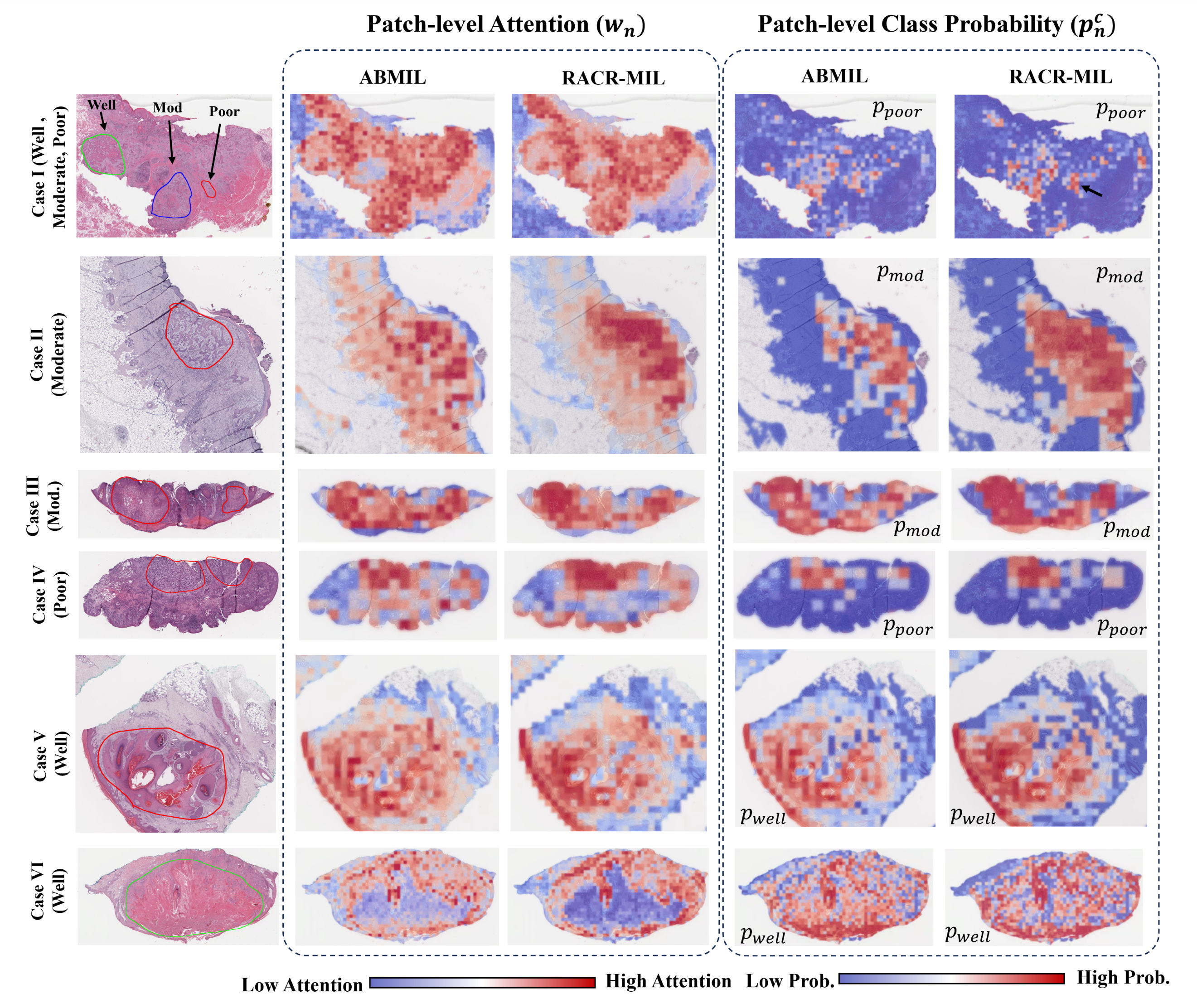}
\caption{Probability heatmaps illustrating the class likelihood of tumor ROIs identified by our model. Tumor annotations are outlined in the leftmost panel. For case I, tumor annotations are colored according to the grade. Heatmaps show patch-level attention and the predicted probability of the ground-truth grade. `Mod.' represents Moderately-differentiated}
\label{fig_prob_heatmap}
\end{figure*}

Table \ref{semantic_spatial} shows that the proposed ranking constraints enable RACR-MIL to address intra-WSI tumor heterogeneity better and achieve up to 2\% higher F1-score across all graph variants. The ranking constraints help the model to consistently assign higher attention to worse tumor regions, thus improving classification performance on higher-grade tumors (Figure \ref{cf_racr}, Appendix). As shown in Table \ref{tab:dual_rank_skin}, the inter- and intra-grade constraints are complementary and achieve the highest F1 and MCC when applied jointly. Using either ranking constraint yields suboptimal grading performance on skin SCC, with the effect of inter-grade ranking being more pronounced in terms of F1 for both feature extractors. Intra-grade ranking alone orders patches by within-grade confidence, but without inter-grade constraint, attention separation across grades can reduce, leading to weaker prioritization of worse tumors. On the other hand, inter-grade ranking alone enforces ordinal alignment across grades but does not calibrate patch-level confidence, making it susceptible to noisy grade-specific probabilities. Combining both constraints addresses these limitations by enforcing reliable inter-grade ordering and stabilizing within-grade confidence calibration of patch-level probability. 

Figure \ref{fig_rank_heatmap} further illustrates the benefit of ranking contraints by comparing attention heatmaps of RACR-MIL with ABMIL, which does not use ranking. In heterogeneous WSIs comprising multiple tumor grades, inter-grade ranking allows the model to consistently focus on the highest-grade tumor region. For example, in Case I ( Fig. \ref{fig_rank_heatmap}), moderately-differentiated areas receive higher attention relative to well-differentiated areas. In Case II (Fig. \ref{fig_rank_heatmap}), poorly differentiated regions are consistently emphasized across all WSI sections. Figure \ref{fig_prob_heatmap} highlights the joint effect of inter- and intra-grade ranking on patch-level confidence and attention allocation. In Case I (Fig.  \ref{fig_prob_heatmap}), which comprises tumor regions from all three grades, higher-grade regions are prioritized over well-differentiated areas. The intra-grade ranking constraint encourages consistency between attention and patch-level class probability, enabling the model to identify a small poorly-differentiated region with higher confidence. In WSIs comprising a single tumor grade (cases II - VI), intra-grade ranking further increases patch-level class likelihood as evident in the probability heatmaps across grades (Figure \ref{fig_prob_heatmap}, column 5). In addition, the dual graph formulation learns improved contextual features, allowing the model to assign consistent importance to morphologically similar and spatially neighbouring tumor patches. This leads to broader and more coherent delineation of tumor regions than ABMIL, consistent with the attention heatmaps.

\begin{figure}[!t]
\centering
\includegraphics[width=0.98\textwidth]{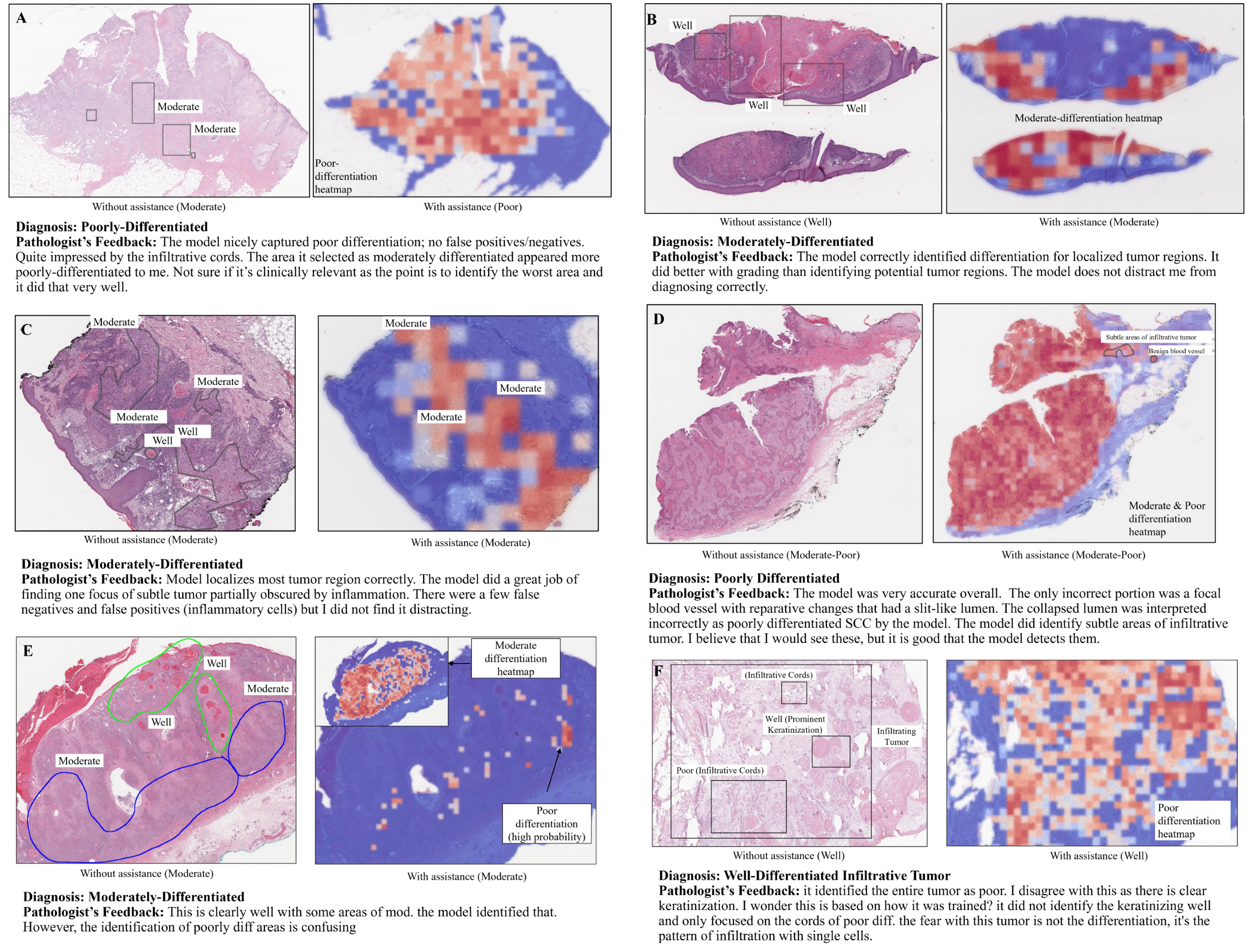}
\caption{Model-guided diagnosis and expert feedback using RACR-MIL for skin SCC: (A, B) RACR-MIL highlighted higher-grade tumor regions missed initially, leading to diagnostic revision. (C, D) The model accurately localized pathologist-annotated regions and identified subtle infiltrative components. (E,F) Clinically distracting cases: (E) RACR-MIL highlights a small portion of the infiltrative tumor front as poorly differentiated while correctly classifying the majority of the tumor as moderately differentiated.
(F) Although RACR-MIL detects poorly differentiated infiltrative tumor regions with good sensitivity, it misclassifies or deprioritizes parts of keratinized areas which represents a limitation from clinician's perspective. Heatmaps indicate predicted class likelihood of the diagnosed grade (`Red' and `Blue' indicate high and low probabilities, respectively).}
\label{fig_review}
\end{figure}
\subsection{Evaluating efficacy as a diagnostic assistant} 
We evaluate RACR-MIL's effectiveness as a diagnostic tool using a small-scale pilot study with two pathologists. We examined the model's impact on pathologists' grading accuracy and its ability to guide them to diagnostically relevant tumor regions. Qualitative feedback from pathologists indicates that the model accurately identified tumor regions and correctly localized the highest-grade tumor ROI(s) in 11 out of 12 cases. Across multiple WSIs, the model identified both small and large-scale tumor regions and successfully highlighted subtle infiltrative patterns, including nuclear atypia.  Model assistance prompted grading revisions in multiple cases, including grade upgrades in three of six slides reviewed by one pathologist.
Representative examples in which the model improved pathologists' accuracy are shown in Figure \ref{fig_review} (A, B). In both these cases, RACR-MIL correctly highlights higher-grade tumor foci, prompting the pathologist to update their diagnosis. In Figure \ref{fig_review}B, it delineates moderately-differentiated regions surrounding a well-differentiated tumor that were initially overlooked. In Figures \ref{fig_review}C and \ref{fig_review}D, the model identified subtle discriminative features such as nuclear atypia dominated by inflammatory cells and infiltrative patterns at the tumor's edge. Overall, pathologists reported that the model improved diagnostic efficiency in 7 out of 12 cases, while in 2 other cases, the model’s output was considered distracting by a clinician (Figures \ref{fig_review}E and \ref{fig_review}F). In Figure \ref{fig_review}F, the model deprioritizes well-differentiated regions due to its training objective, which encourages prioritization of the highest-grade poorly-differentiated tumor region. From the clinician’s perspective, the failure to appropriately recognize and prioritize keratinization is a key limitation. In Figure \ref{fig_review}E, RACR-MIL falsely identifies small sections of a tumor invasive front as poorly-differentiated while correctly classifying the majority of the moderately-differentiated regions. These findings demonstrate the model's potential for translation as an AI-assisted tool to support more accurate and efficient SCC grading.

\subsection{Hyperparameter selection}
To select the hyperparameters for the ranking losses, diversity regularization, and latent graph construction, we evaluated validation performance using the same five-fold cross-validation setup. Table~\ref{hyper_val} summarizes the results on skin SCC for the PathGen-CLIP and SCC-DINO feature extractors across different values of the ranking-loss weight \((\lambda_1)\), diversity-loss weight \((\lambda_2)\), and latent graph neighborhood size \(K\).
To reduce the number of free hyperparameters, we used the same weight for both the inter-grade and intra-grade ranking losses, which performed well in practice. For PathGen-CLIP, the validation F1 score remained stable across all tested configurations, with the best performance obtained at \(\lambda_1 = 0.05\), \(\lambda_2 = 0.001\), and \(K=5\). The corresponding configuration with \(K=8\) gave a slightly lower F1. Similarly, SCC-DINO features showed stable behavior, with the highest validation F1 achieved at \(\lambda_1 = 0.20\) under both \((\lambda_2 = 0.100, K=5)\) and \((\lambda_2 = 0.100, K=8)\). Since these settings resulted in identical performance, we selected \(K=5\) to reduce the risk of oversmoothing in a denser latent graph. To further limit dataset-specific tuning, we used the same ranking-loss weight across all datasets. We also used the same diversity-loss weight \((\lambda_2 = 0.001)\) for all PathGen-CLIP experiments. For SCC-DINO, \(\lambda_2 = 0.1\) was used for skin SCC, while a smaller value of \(\lambda_2 = 0.01\) gave better validation performance for lung SCC and head and neck SCC. For latent graph diffusion, we set the Personalized PageRank parameter \(\alpha=0.25\) to preserve locality and avoid introducing overly distant neighbors from longer diffusion walks. Overall, these results indicate that the proposed framework is not highly sensitive to hyperparameter selection. The use of shared ranking-loss weights and largely consistent settings across datasets suggests that the observed gains are robust and not dependent on heavy dataset-specific tuning.

We also evaluated the sensitivity of RACR-MIL to the patch-selection probability threshold $\tau$ used in the ranking formulation (Appendix~\ref{classification_result_threshold}). We set the minimum threshold to $\tau=0.6$ to exclude low-confidence patch predictions with greater grade uncertainty while retaining enough patches for ranking. Our model's performance remains stable across thresholds for both Skin and Lung SCCs, demonstrating RACR-MIL's robustness to patch selection. The threshold of $0.6$ achieved the highest overall F1-score across five folds, and RACR-MIL outperforms the MIL baselines at all evaluated thresholds for PathGen-CLIP features. To minimize overtuning to a specific dataset or feature extractor, we used the lowest threshold that provided sufficient class confidence ($\tau=0.6$) across all feature extractors and datasets in the main experiments.

\begin{table*}[!t]
\centering
\caption{Hyperparameter selection based on validation accuracy across ranking-loss weight, diversity-loss weight, and latent graph nearest neighbours count ($K$) for PathGen-CLIP and SCC-DINO feature extractors.}
\label{hyper_val}

\resizebox{0.48\textwidth}{!}{
\begin{tabular}{c c c c c c}
\hline
\multicolumn{6}{c}{\textbf{PathGen-CLIP}} \\
\hline
\textbf{$\lambda_{1}$} &
\textbf{$\lambda_{2}$} &
\textbf{$K$} &
\textbf{Val F1} &
\textbf{Val MCC} &
\textbf{Val $\kappa$} \\
\hline
\textbf{0.05} & \textbf{0.001} & \textbf{5} & \textbf{0.800} & 0.786 & 0.868 \\
0.10 & 0.001 & 5 & 0.785 & 0.774 & 0.855 \\
0.20 & 0.001 & 5 & 0.796 & 0.788 & 0.860 \\
0.05 & 0.010 & 5 & 0.797 & 0.781 & 0.859 \\
0.10 & 0.010 & 5 & 0.789 & 0.775 & 0.852 \\
0.20 & 0.010 & 5 & 0.794 & 0.785 & 0.859 \\
\midrule
0.05 & 0.001 & 8 & 0.799 & 0.786 & 0.872 \\
0.05 & 0.010 & 8 & 0.798 & 0.785 & 0.866 \\
\hline
\end{tabular}
}
\hspace{0.01\textwidth}
\resizebox{0.48\textwidth}{!}{
\begin{tabular}{c c c c c c}
\hline
\multicolumn{6}{c}{\textbf{SCC-DINO}} \\
\hline
\textbf{$\lambda_{1}$} &
\textbf{$\lambda_{2}$} &
\textbf{$K$} &
\textbf{Val F1} &
\textbf{Val MCC} &
\textbf{Val $\kappa$} \\
\hline
0.05 & 0.010 & 5 & 0.797 & 0.792 & 0.841 \\
0.10 & 0.010 & 5 & 0.798 & 0.796 & 0.848 \\
0.20 & 0.010 & 5 & 0.801 & 0.800 & 0.840 \\
0.05 & 0.100 & 5 & 0.802 & 0.800 & 0.846 \\
0.10 & 0.100 & 5 & 0.800 & 0.797 & 0.848 \\
\textbf{0.20} & \textbf{0.100} & \textbf{5} & \textbf{0.803} & 0.803 & 0.841 \\
\midrule
0.20 & 0.010 & 8 & 0.803 & 0.803 & 0.841 \\
0.20 & 0.100 & 8 & 0.803 & 0.803 & 0.841 \\
\hline
\end{tabular}
}
\end{table*}

\section{Conclusion}
We present RACR-MIL, a novel multiple instance learning framework for weakly-supervised SCC tumor grading in whole slide images that integrates graph-based contextual learning with rank-ordering constrained pooling. Drawing on clinical domain knowledge, RACR-MIL employs a dual-graph formulation to capture multi-level tissue context: (1) local tumor and microenvironment context, including keratinization and stromal organization, and (2) phenotypically similar tumor regions, whether spatially close or distant, to reflect tumor growth structure. By leveraging distinct spatial and latent graphs and incorporating self-attention and diversity regularization, our model enhances the extraction of complementary contextual information. RACR-MIL consistently outperforms existing methods across multiple anatomies, with strong improvements in identifying high-risk tumors as well as tumor localization. Each of our proposed innovations contributes to improved grading performance, as validated through comprehensive ablation studies. Beyond grading, RACR-MIL’s ranking-guided approach can be extended to outcome prediction tasks involving ordinal outcomes such as disease severity, tumor stage or tumor burden estimation. Additionally, its accurate identification of diagnostically relevant tumor regions as well as positive clinician feedback in a pilot study highlights its potential as an AI-assisted tool to support and enhance pathologists' workflow. 

There are several limitations of this study that we will address in future work. (i) MIL approaches were compared using the mean and standard deviation across cross-validation folds, without statistical significance testing. Although RACR-MIL showed consistent improvements over the baselines across the three datasets, rigorous statistical validation would require larger patient cohorts and repeated runs with multiple random seeds, which were beyond the scope of the current experiments. (ii) The model can confuse non-tumor patches with tumor patches, resulting in false positives. Improved feature and contextual information extraction might help in separating different cell types in the feature space. (iii) The use of relatively larger patches provides grading-relevant cellular and tissue-level context but may reduce tumor localization granularity. Smaller patches could improve boundary precision, particularly for small tumor regions, but may provide less context for assessing differentiation. Future work should evaluate multi-scale patch representations and spatial auxiliary tasks, such as depth prediction, to improve tumor localization along with grade classification. (iv) Clinical translation of the proposed method requires quantification of the uncertainty associated with the prediction to build trust. (v) Finally, the clinical utility of the proposed approach in reducing inter-pathologist variability needs to be evaluated on a larger cohort. Our future work would also focus on extending and evaluating the performance of RACR-MIL's generalizable framework to survival prediction and other histologic cancer types, such as adenocarcinoma.

\section{CRediT authorship contribution statement}
\textbf{Anirudh Choudhary}: Conceptualization, Methodology, Software, Validation, Formal analysis, Investigation, Data curation, Writing – original draft, Writing – review \& editing, Visualization. \textbf{Mosbah Aouad}: Writing – review \& editing, Validation. \textbf{Krishnakant Saboo}: Writing - original draft, Validation. \textbf{Angelina Hwang}: Data curation, Resources. \textbf{Jacob Kechter}: Data curation, Resources. \textbf{Blake Bordeaux}: Data curation, Validation. \textbf{Puneet Bhullar}: Data curation, Resources. \textbf{David DiCaudo}: Validation, Investigation, Data curation. \textbf{Steven Nelson}: Validation, Investigation, Data Curation. \textbf{Nneka Comfere}: Validation, Data curation. \textbf{Emma Johnson}: Validation, Data curation. \textbf{Jason Sluzevich}: Validation, Data curation. \textbf{Olayemi Sokumbi}: Validation, Data curation, Investigation. \textbf{Leah Swanson}: Data curation. \textbf{Dennis Murphree}: Funding acquisition, Project administration, Resources. \textbf{Aaron Mangold}: Conceptualization, Funding acquisition, Project administration, Resources, Supervision, Writing – review \& editing. \textbf{Ravishankar Iyer}: Funding acquisition, Project administration, Resources, Supervision, Writing – review \& editing. 

\section{Declaration of Competing Interests}
The authors state the following competing interests: Aaron Mangold has consulted for Phlecs BV, Kyowa, Eli Lilly, Momenta, UCB, and Regeneron in the past, greater than 24 months ago. He has consulted for Incyte, Soligenix, Clarivate, Argenyx, and Bristol Myers Squibb in the past, less than 12 months ago. He consults for Nuvig, Tourmaline Bio, Janssen, Boehringer Ingelheim currently. He consults for Regeneron and Pfizer currently with payments to the institution. He has grant support from Kyowa, Miragen, Regeneron, Corbus, Pfizer, Incyte, Eli Lilly, Argenx, Palvella, Abbvie, Priovant, Bristol Myers Squibb, Merck in the last 24 months. Beyond 24 months, grant support has come from Sun Pharma, Elorac, Novartis, and Janssen. He has received royalties from Adelphi Values and Clarivate. Aaron Mangold, Ravishankar Iyer, Anirudh Choudhary and Krishnakant Saboo have patent $PCT/US2024/062101$ pending to Mayo Foundation for Medical Education and Research and University of Illinois Urbana-Champaign. All other authors have no known competing financial interests or personal relationships that could have appeared to influence the work reported in this paper.

\section{Data Statement}
Whole-slide image datasets from the TCGA Research Network and the Clinical Proteomic Tumor Analysis Consortium, are publicly available through the NCI Genomic Data Commons Portal (\href{https://portal.gdc.cancer.gov/}{https://portal.gdc.cancer.gov/}) and the Cancer Imaging Archive (\href{https://www.cancerimagingarchive.net/}{https://www.cancerimagingarchive.net/}), respectively. These datasets were used to evaluate the performance of RACR-MIL for SCC grading in lung, and head and neck specimens. The in-house skin SCC dataset analyzed in this study is not publicly available due to patient privacy considerations but may be made accessible upon reasonable request to the corresponding author. Access requires submission of a brief research proposal and agreement to a data use policy limited to non-commercial academic research.

\section*{Acknowledgments}
This work was supported by the Mayo Clinic and Illinois Alliance Fellowship for Technology-based Healthcare Research. This research utilized resources from the National Center for Supercomputing Applications (NCSA), supported by the University of Illinois Urbana-Champaign. We thank Chang Hu, Alysia Hughes, Xing Li, Alyssa Stockard, and Zachary Leibovit-Reiben for their valuable feedback. Data used in this publication were generated by the National Cancer Institute Clinical Proteomic Tumor Analysis Consortium (CPTAC) and the TCGA Research Network. 
\bibliographystyle{elsarticle-harv} 
\bibliography{main}

\begin{thebibliography}{46}
\expandafter\ifx\csname natexlab\endcsname\relax\def\natexlab#1{#1}\fi
\providecommand{\url}[1]{\texttt{#1}}
\providecommand{\href}[2]{#2}
\providecommand{\path}[1]{#1}
\providecommand{\DOIprefix}{doi:}
\providecommand{\ArXivprefix}{arXiv:}
\providecommand{\URLprefix}{URL: }
\providecommand{\Pubmedprefix}{pmid:}
\providecommand{\doi}[1]{\href{http://dx.doi.org/#1}{\path{#1}}}
\providecommand{\Pubmed}[1]{\href{pmid:#1}{\path{#1}}}
\providecommand{\bibinfo}[2]{#2}
\ifx\xfnm\relax \def\xfnm[#1]{\unskip,\space#1}\fi
\bibitem[{Alfasly et~al.(2024)Alfasly, Shafique, Nejat, Khan, Alsaafin, Alabtah and Tizhoosh}]{alfasly2024rotation}
\bibinfo{author}{Alfasly, S.}, \bibinfo{author}{Shafique, A.}, \bibinfo{author}{Nejat, P.}, \bibinfo{author}{Khan, J.}, \bibinfo{author}{Alsaafin, A.}, \bibinfo{author}{Alabtah, G.}, \bibinfo{author}{Tizhoosh, H.R.}, \bibinfo{year}{2024}.
\newblock \bibinfo{title}{Rotation-agnostic image representation learning for digital pathology}, in: \bibinfo{booktitle}{Proceedings of the IEEE/CVF Conference on Computer Vision and Pattern Recognition}, pp. \bibinfo{pages}{11683--11693}.
\bibitem[{Behzadi et~al.(2024)Behzadi, Madani, Wang, Bai, Bhardwaj, Tarakanova, Yamase, Nam and Nabavi}]{behzadi2024weakly}
\bibinfo{author}{Behzadi, M.M.}, \bibinfo{author}{Madani, M.}, \bibinfo{author}{Wang, H.}, \bibinfo{author}{Bai, J.}, \bibinfo{author}{Bhardwaj, A.}, \bibinfo{author}{Tarakanova, A.}, \bibinfo{author}{Yamase, H.}, \bibinfo{author}{Nam, G.H.}, \bibinfo{author}{Nabavi, S.}, \bibinfo{year}{2024}.
\newblock \bibinfo{title}{Weakly-supervised deep learning model for prostate cancer diagnosis and gleason grading of histopathology images}.
\newblock \bibinfo{journal}{Biomedical Signal Processing and Control} \bibinfo{volume}{95}, \bibinfo{pages}{106351}.
\bibitem[{Burges et~al.(2005)Burges, Shaked, Renshaw, Lazier, Deeds, Hamilton and Hullender}]{10.1145/1102351.1102363}
\bibinfo{author}{Burges, C.}, \bibinfo{author}{Shaked, T.}, \bibinfo{author}{Renshaw, E.}, \bibinfo{author}{Lazier, A.}, \bibinfo{author}{Deeds, M.}, \bibinfo{author}{Hamilton, N.}, \bibinfo{author}{Hullender, G.}, \bibinfo{year}{2005}.
\newblock \bibinfo{title}{Learning to rank using gradient descent}, in: \bibinfo{booktitle}{Proceedings of the 22nd International Conference on Machine Learning}, \bibinfo{publisher}{Association for Computing Machinery}, \bibinfo{address}{New York, NY, USA}. p. \bibinfo{pages}{89–96}.
\newblock \URLprefix \url{https://doi.org/10.1145/1102351.1102363}, \DOIprefix\doi{10.1145/1102351.1102363}.
\bibitem[{Caron et~al.(2021)Caron, Touvron, Misra, J{\'e}gou, Mairal, Bojanowski and Joulin}]{caron2021emerging}
\bibinfo{author}{Caron, M.}, \bibinfo{author}{Touvron, H.}, \bibinfo{author}{Misra, I.}, \bibinfo{author}{J{\'e}gou, H.}, \bibinfo{author}{Mairal, J.}, \bibinfo{author}{Bojanowski, P.}, \bibinfo{author}{Joulin, A.}, \bibinfo{year}{2021}.
\newblock \bibinfo{title}{Emerging properties in self-supervised vision transformers}, in: \bibinfo{booktitle}{Proceedings of the IEEE/CVF international conference on computer vision}, pp. \bibinfo{pages}{9650--9660}.
\bibitem[{Chan et~al.(2023)Chan, Cendra, Ma, Yin and Yu}]{chan2023histopathology}
\bibinfo{author}{Chan, T.H.}, \bibinfo{author}{Cendra, F.J.}, \bibinfo{author}{Ma, L.}, \bibinfo{author}{Yin, G.}, \bibinfo{author}{Yu, L.}, \bibinfo{year}{2023}.
\newblock \bibinfo{title}{Histopathology whole slide image analysis with heterogeneous graph representation learning}, in: \bibinfo{booktitle}{Proceedings of the IEEE/CVF conference on computer vision and pattern recognition}, pp. \bibinfo{pages}{15661--15670}.
\bibitem[{Chen et~al.(2021)Chen, Lu, Shaban, Chen, Chen, Williamson and Mahmood}]{chen2021whole}
\bibinfo{author}{Chen, R.J.}, \bibinfo{author}{Lu, M.Y.}, \bibinfo{author}{Shaban, M.}, \bibinfo{author}{Chen, C.}, \bibinfo{author}{Chen, T.Y.}, \bibinfo{author}{Williamson, D.F.}, \bibinfo{author}{Mahmood, F.}, \bibinfo{year}{2021}.
\newblock \bibinfo{title}{Whole slide images are 2d point clouds: Context-aware survival prediction using patch-based graph convolutional networks}, in: \bibinfo{booktitle}{International Conference on Medical Image Computing and Computer-Assisted Intervention}, \bibinfo{organization}{Springer}. pp. \bibinfo{pages}{339--349}.
\bibitem[{Cui et~al.(2019)Cui, Jia, Lin, Song and Belongie}]{cui2019class}
\bibinfo{author}{Cui, Y.}, \bibinfo{author}{Jia, M.}, \bibinfo{author}{Lin, T.Y.}, \bibinfo{author}{Song, Y.}, \bibinfo{author}{Belongie, S.}, \bibinfo{year}{2019}.
\newblock \bibinfo{title}{Class-balanced loss based on effective number of samples}, in: \bibinfo{booktitle}{Proceedings of the IEEE/CVF conference on computer vision and pattern recognition}, pp. \bibinfo{pages}{9268--9277}.
\bibitem[{Diede et~al.(2024)Diede, Walker, Carr and Shahwan}]{diede2024grading}
\bibinfo{author}{Diede, C.}, \bibinfo{author}{Walker, T.}, \bibinfo{author}{Carr, D.R.}, \bibinfo{author}{Shahwan, K.T.}, \bibinfo{year}{2024}.
\newblock \bibinfo{title}{Grading differentiation in cutaneous squamous cell carcinoma: a review of the literature}.
\newblock \bibinfo{journal}{Archives of Dermatological Research} \bibinfo{volume}{316}, \bibinfo{pages}{434}.
\bibitem[{Dominguez-Morales et~al.(2024)Dominguez-Morales, Duran-Lopez, Marini, Vicente-Diaz, Linares-Barranco, Atzori and M{\"u}ller}]{dominguez2024systematic}
\bibinfo{author}{Dominguez-Morales, J.P.}, \bibinfo{author}{Duran-Lopez, L.}, \bibinfo{author}{Marini, N.}, \bibinfo{author}{Vicente-Diaz, S.}, \bibinfo{author}{Linares-Barranco, A.}, \bibinfo{author}{Atzori, M.}, \bibinfo{author}{M{\"u}ller, H.}, \bibinfo{year}{2024}.
\newblock \bibinfo{title}{A systematic comparison of deep learning methods for gleason grading and scoring}.
\newblock \bibinfo{journal}{Medical Image Analysis} \bibinfo{volume}{95}, \bibinfo{pages}{103191}.
\bibitem[{Early et~al.(2023)Early, Cheung, Cutajar, Xie, Kandola and Twomey}]{early2024inherently}
\bibinfo{author}{Early, J.}, \bibinfo{author}{Cheung, G.K.}, \bibinfo{author}{Cutajar, K.}, \bibinfo{author}{Xie, H.}, \bibinfo{author}{Kandola, J.}, \bibinfo{author}{Twomey, N.}, \bibinfo{year}{2023}.
\newblock \bibinfo{title}{Inherently interpretable time series classification via multiple instance learning}.
\newblock \bibinfo{journal}{arXiv preprint arXiv:2311.10049} .
\bibitem[{Edwards et~al.(2015)Edwards, Oberti, Thangudu, Cai, McGarvey, Jacob, Madhavan and Ketchum}]{edwards2015cptac}
\bibinfo{author}{Edwards, N.J.}, \bibinfo{author}{Oberti, M.}, \bibinfo{author}{Thangudu, R.R.}, \bibinfo{author}{Cai, S.}, \bibinfo{author}{McGarvey, P.B.}, \bibinfo{author}{Jacob, S.}, \bibinfo{author}{Madhavan, S.}, \bibinfo{author}{Ketchum, K.A.}, \bibinfo{year}{2015}.
\newblock \bibinfo{title}{The cptac data portal: a resource for cancer proteomics research}.
\newblock \bibinfo{journal}{Journal of proteome research} \bibinfo{volume}{14}, \bibinfo{pages}{2707--2713}.
\bibitem[{Fourkioti et~al.(2023)Fourkioti, De~Vries, Jin, Alexander and Bakal}]{fourkioti2023camil}
\bibinfo{author}{Fourkioti, O.}, \bibinfo{author}{De~Vries, M.}, \bibinfo{author}{Jin, C.}, \bibinfo{author}{Alexander, D.C.}, \bibinfo{author}{Bakal, C.}, \bibinfo{year}{2023}.
\newblock \bibinfo{title}{Camil: Context-aware multiple instance learning for cancer detection and subtyping in whole slide images}.
\newblock \bibinfo{journal}{arXiv preprint arXiv:2305.05314} .
\bibitem[{Gao et~al.(2013)Gao, Aksoy, Dogrusoz, Dresdner, Gross, Sumer, Sun, Jacobsen, Sinha, Larsson et~al.}]{gao2013integrative}
\bibinfo{author}{Gao, J.}, \bibinfo{author}{Aksoy, B.A.}, \bibinfo{author}{Dogrusoz, U.}, \bibinfo{author}{Dresdner, G.}, \bibinfo{author}{Gross, B.}, \bibinfo{author}{Sumer, S.O.}, \bibinfo{author}{Sun, Y.}, \bibinfo{author}{Jacobsen, A.}, \bibinfo{author}{Sinha, R.}, \bibinfo{author}{Larsson, E.}, et~al., \bibinfo{year}{2013}.
\newblock \bibinfo{title}{Integrative analysis of complex cancer genomics and clinical profiles using the cbioportal}.
\newblock \bibinfo{journal}{Science signaling} \bibinfo{volume}{6}, \bibinfo{pages}{pl1--pl1}.
\bibitem[{Gasteiger et~al.(2019)Gasteiger, Wei{\ss}enberger and G{\"u}nnemann}]{gasteiger2019diffusion}
\bibinfo{author}{Gasteiger, J.}, \bibinfo{author}{Wei{\ss}enberger, S.}, \bibinfo{author}{G{\"u}nnemann, S.}, \bibinfo{year}{2019}.
\newblock \bibinfo{title}{Diffusion improves graph learning}.
\newblock \bibinfo{journal}{Advances in neural information processing systems} \bibinfo{volume}{32}.
\bibitem[{Geijs et~al.(2024)Geijs, Dooper, Aswolinskiy, Hillen, Amir and Litjens}]{GEIJS2024103063}
\bibinfo{author}{Geijs, D.J.}, \bibinfo{author}{Dooper, S.}, \bibinfo{author}{Aswolinskiy, W.}, \bibinfo{author}{Hillen, L.M.}, \bibinfo{author}{Amir, A.L.}, \bibinfo{author}{Litjens, G.}, \bibinfo{year}{2024}.
\newblock \bibinfo{title}{Detection and subtyping of basal cell carcinoma in whole-slide histopathology using weakly-supervised learning}.
\newblock \bibinfo{journal}{Medical Image Analysis} \bibinfo{volume}{93}, \bibinfo{pages}{103063}.
\newblock \URLprefix \url{https://www.sciencedirect.com/science/article/pii/S1361841523003237}, \DOIprefix\doi{https://doi.org/10.1016/j.media.2023.103063}.
\bibitem[{Guan et~al.(2022)Guan, Zhang, Tian, Yang, Dong, Xiang, Yang, Huang, Zhang and Han}]{guan2022node}
\bibinfo{author}{Guan, Y.}, \bibinfo{author}{Zhang, J.}, \bibinfo{author}{Tian, K.}, \bibinfo{author}{Yang, S.}, \bibinfo{author}{Dong, P.}, \bibinfo{author}{Xiang, J.}, \bibinfo{author}{Yang, W.}, \bibinfo{author}{Huang, J.}, \bibinfo{author}{Zhang, Y.}, \bibinfo{author}{Han, X.}, \bibinfo{year}{2022}.
\newblock \bibinfo{title}{Node-aligned graph convolutional network for whole-slide image representation and classification}, in: \bibinfo{booktitle}{Proceedings of the IEEE/CVF Conference on Computer Vision and Pattern Recognition}, pp. \bibinfo{pages}{18813--18823}.
\bibitem[{Hu et~al.(2020)Hu, Dong, Wang and Sun}]{hu2020heterogeneous}
\bibinfo{author}{Hu, Z.}, \bibinfo{author}{Dong, Y.}, \bibinfo{author}{Wang, K.}, \bibinfo{author}{Sun, Y.}, \bibinfo{year}{2020}.
\newblock \bibinfo{title}{Heterogeneous graph transformer}, in: \bibinfo{booktitle}{Proceedings of the web conference 2020}, pp. \bibinfo{pages}{2704--2710}.
\bibitem[{Ilse et~al.(2018)Ilse, Tomczak and Welling}]{ilse2018attention}
\bibinfo{author}{Ilse, M.}, \bibinfo{author}{Tomczak, J.}, \bibinfo{author}{Welling, M.}, \bibinfo{year}{2018}.
\newblock \bibinfo{title}{Attention-based deep multiple instance learning}, in: \bibinfo{booktitle}{International conference on machine learning}, \bibinfo{organization}{PMLR}. pp. \bibinfo{pages}{2127--2136}.
\bibitem[{Kanavati et~al.(2020)Kanavati, Toyokawa, Momosaki, Rambeau, Kozuma, Shoji, Yamazaki, Takeo, Iizuka and Tsuneki}]{kanavati2020weakly}
\bibinfo{author}{Kanavati, F.}, \bibinfo{author}{Toyokawa, G.}, \bibinfo{author}{Momosaki, S.}, \bibinfo{author}{Rambeau, M.}, \bibinfo{author}{Kozuma, Y.}, \bibinfo{author}{Shoji, F.}, \bibinfo{author}{Yamazaki, K.}, \bibinfo{author}{Takeo, S.}, \bibinfo{author}{Iizuka, O.}, \bibinfo{author}{Tsuneki, M.}, \bibinfo{year}{2020}.
\newblock \bibinfo{title}{Weakly-supervised learning for lung carcinoma classification using deep learning}.
\newblock \bibinfo{journal}{Scientific reports} \bibinfo{volume}{10}, \bibinfo{pages}{9297}.
\bibitem[{Li et~al.(2021)Li, Li and Eliceiri}]{li2021dual}
\bibinfo{author}{Li, B.}, \bibinfo{author}{Li, Y.}, \bibinfo{author}{Eliceiri, K.W.}, \bibinfo{year}{2021}.
\newblock \bibinfo{title}{Dual-stream multiple instance learning network for whole slide image classification with self-supervised contrastive learning}, in: \bibinfo{booktitle}{Proceedings of the IEEE/CVF conference on computer vision and pattern recognition}, pp. \bibinfo{pages}{14318--14328}.
\bibitem[{Li et~al.(2020)Li, Xiong, Thabet and Ghanem}]{li2020deepergcn}
\bibinfo{author}{Li, G.}, \bibinfo{author}{Xiong, C.}, \bibinfo{author}{Thabet, A.}, \bibinfo{author}{Ghanem, B.}, \bibinfo{year}{2020}.
\newblock \bibinfo{title}{Deepergcn: All you need to train deeper gcns}.
\newblock \bibinfo{journal}{arXiv preprint arXiv:2006.07739} .
\bibitem[{Li et~al.(2023)Li, Zhang, Zhu, Cai, Zheng and Yang}]{li2023longmil}
\bibinfo{author}{Li, H.}, \bibinfo{author}{Zhang, Y.}, \bibinfo{author}{Zhu, C.}, \bibinfo{author}{Cai, J.}, \bibinfo{author}{Zheng, S.}, \bibinfo{author}{Yang, L.}, \bibinfo{year}{2023}.
\newblock \bibinfo{title}{Long-mil: scaling long contextual multiple instance learning for histopathology whole slide image analysis}.
\newblock \bibinfo{journal}{arXiv preprint arXiv:2311.12885} .
\bibitem[{Li et~al.(2024)Li, Chen, Chu, Sun, Guan, Han and He}]{li2024dynamic}
\bibinfo{author}{Li, J.}, \bibinfo{author}{Chen, Y.}, \bibinfo{author}{Chu, H.}, \bibinfo{author}{Sun, Q.}, \bibinfo{author}{Guan, T.}, \bibinfo{author}{Han, A.}, \bibinfo{author}{He, Y.}, \bibinfo{year}{2024}.
\newblock \bibinfo{title}{Dynamic graph representation with knowledge-aware attention for histopathology whole slide image analysis}, in: \bibinfo{booktitle}{Proceedings of the IEEE/CVF Conference on Computer Vision and Pattern Recognition}, pp. \bibinfo{pages}{11323--11332}.
\bibitem[{Loshchilov and Hutter(2018)}]{loshchilov2018decoupled}
\bibinfo{author}{Loshchilov, I.}, \bibinfo{author}{Hutter, F.}, \bibinfo{year}{2018}.
\newblock \bibinfo{title}{Decoupled weight decay regularization}.
\newblock \bibinfo{journal}{arXiv preprint arXiv:1711.05101} .
\bibitem[{Lu et~al.(2024)Lu, Chen, Williamson, Chen, Liang, Ding, Jaume, Odintsov, Le, Gerber et~al.}]{lu2024visual}
\bibinfo{author}{Lu, M.Y.}, \bibinfo{author}{Chen, B.}, \bibinfo{author}{Williamson, D.F.}, \bibinfo{author}{Chen, R.J.}, \bibinfo{author}{Liang, I.}, \bibinfo{author}{Ding, T.}, \bibinfo{author}{Jaume, G.}, \bibinfo{author}{Odintsov, I.}, \bibinfo{author}{Le, L.P.}, \bibinfo{author}{Gerber, G.}, et~al., \bibinfo{year}{2024}.
\newblock \bibinfo{title}{A visual-language foundation model for computational pathology}.
\newblock \bibinfo{journal}{Nature medicine} \bibinfo{volume}{30}, \bibinfo{pages}{863--874}.
\bibitem[{Lu et~al.(2021)Lu, Williamson, Chen, Chen, Barbieri and Mahmood}]{lu2021data}
\bibinfo{author}{Lu, M.Y.}, \bibinfo{author}{Williamson, D.F.}, \bibinfo{author}{Chen, T.Y.}, \bibinfo{author}{Chen, R.J.}, \bibinfo{author}{Barbieri, M.}, \bibinfo{author}{Mahmood, F.}, \bibinfo{year}{2021}.
\newblock \bibinfo{title}{Data-efficient and weakly supervised computational pathology on whole-slide images}.
\newblock \bibinfo{journal}{Nature Biomedical Engineering} \bibinfo{volume}{5}, \bibinfo{pages}{555--570}.
\bibitem[{Mun et~al.(2021)Mun, Paik, Shin, Kwak and Chang}]{mun2021yet}
\bibinfo{author}{Mun, Y.}, \bibinfo{author}{Paik, I.}, \bibinfo{author}{Shin, S.J.}, \bibinfo{author}{Kwak, T.Y.}, \bibinfo{author}{Chang, H.}, \bibinfo{year}{2021}.
\newblock \bibinfo{title}{Yet another automated gleason grading system (yaaggs) by weakly supervised deep learning}.
\newblock \bibinfo{journal}{npj Digital Medicine} \bibinfo{volume}{4}, \bibinfo{pages}{99}.
\bibitem[{Myronenko et~al.(2021)Myronenko, Xu, Yang, Roth and Xu}]{myronenko2021accounting}
\bibinfo{author}{Myronenko, A.}, \bibinfo{author}{Xu, Z.}, \bibinfo{author}{Yang, D.}, \bibinfo{author}{Roth, H.R.}, \bibinfo{author}{Xu, D.}, \bibinfo{year}{2021}.
\newblock \bibinfo{title}{Accounting for dependencies in deep learning based multiple instance learning for whole slide imaging}, in: \bibinfo{booktitle}{International Conference on Medical Image Computing and Computer-Assisted Intervention}, \bibinfo{organization}{Springer}. pp. \bibinfo{pages}{329--338}.
\bibitem[{Ot{\'a}lora et~al.(2021)Ot{\'a}lora, Marini, M{\"u}ller and Atzori}]{otalora2021combining}
\bibinfo{author}{Ot{\'a}lora, S.}, \bibinfo{author}{Marini, N.}, \bibinfo{author}{M{\"u}ller, H.}, \bibinfo{author}{Atzori, M.}, \bibinfo{year}{2021}.
\newblock \bibinfo{title}{Combining weakly and strongly supervised learning improves strong supervision in gleason pattern classification}.
\newblock \bibinfo{journal}{BMC Medical Imaging} \bibinfo{volume}{21}, \bibinfo{pages}{1--14}.
\bibitem[{Prezzano et~al.(2021)Prezzano, Scott, Smith, Mannava and Ibrahim}]{prezzano2021concordance}
\bibinfo{author}{Prezzano, J.C.}, \bibinfo{author}{Scott, G.A.}, \bibinfo{author}{Smith, F.L.}, \bibinfo{author}{Mannava, K.A.}, \bibinfo{author}{Ibrahim, S.F.}, \bibinfo{year}{2021}.
\newblock \bibinfo{title}{Concordance of squamous cell carcinoma histologic grading among dermatopathologists and mohs surgeons}.
\newblock \bibinfo{journal}{Dermatologic Surgery} \bibinfo{volume}{47}, \bibinfo{pages}{1433--1437}.
\bibitem[{Qin et~al.(2011)Qin, Gammeter, Bossard, Quack and Van~Gool}]{qin2011hello}
\bibinfo{author}{Qin, D.}, \bibinfo{author}{Gammeter, S.}, \bibinfo{author}{Bossard, L.}, \bibinfo{author}{Quack, T.}, \bibinfo{author}{Van~Gool, L.}, \bibinfo{year}{2011}.
\newblock \bibinfo{title}{Hello neighbor: Accurate object retrieval with k-reciprocal nearest neighbors}, in: \bibinfo{booktitle}{CVPR 2011}, \bibinfo{organization}{IEEE}. pp. \bibinfo{pages}{777--784}.
\bibitem[{Rogers et~al.(2015)Rogers, Weinstock, Feldman and Coldiron}]{rogers2015incidence}
\bibinfo{author}{Rogers, H.W.}, \bibinfo{author}{Weinstock, M.A.}, \bibinfo{author}{Feldman, S.R.}, \bibinfo{author}{Coldiron, B.M.}, \bibinfo{year}{2015}.
\newblock \bibinfo{title}{Incidence estimate of nonmelanoma skin cancer (keratinocyte carcinomas) in the us population, 2012}.
\newblock \bibinfo{journal}{JAMA dermatology} \bibinfo{volume}{151}, \bibinfo{pages}{1081--1086}.
\bibitem[{Shao et~al.(2021)Shao, Bian, Chen, Wang, Zhang, Ji et~al.}]{shao2021transmil}
\bibinfo{author}{Shao, Z.}, \bibinfo{author}{Bian, H.}, \bibinfo{author}{Chen, Y.}, \bibinfo{author}{Wang, Y.}, \bibinfo{author}{Zhang, J.}, \bibinfo{author}{Ji, X.}, et~al., \bibinfo{year}{2021}.
\newblock \bibinfo{title}{Transmil: Transformer based correlated multiple instance learning for whole slide image classification}.
\newblock \bibinfo{journal}{Advances in neural information processing systems} \bibinfo{volume}{34}, \bibinfo{pages}{2136--2147}.
\bibitem[{Silva-Rodriguez et~al.(2021)Silva-Rodriguez, Colomer, Dolz and Naranjo}]{silva2021self}
\bibinfo{author}{Silva-Rodriguez, J.}, \bibinfo{author}{Colomer, A.}, \bibinfo{author}{Dolz, J.}, \bibinfo{author}{Naranjo, V.}, \bibinfo{year}{2021}.
\newblock \bibinfo{title}{Self-learning for weakly supervised {Gleason} grading of local patterns}.
\newblock \bibinfo{journal}{IEEE Journal of Biomedical and Health Informatics} \bibinfo{volume}{25}, \bibinfo{pages}{3094--3104}.
\bibitem[{Sun et~al.(2025a)Sun, Tessier, Meeuwsen, Grisi, van Midden, Litjens and Baumgartner}]{sun2025label}
\bibinfo{author}{Sun, S.}, \bibinfo{author}{Tessier, L.}, \bibinfo{author}{Meeuwsen, F.}, \bibinfo{author}{Grisi, C.}, \bibinfo{author}{van Midden, D.}, \bibinfo{author}{Litjens, G.}, \bibinfo{author}{Baumgartner, C.F.}, \bibinfo{year}{2025}a.
\newblock \bibinfo{title}{Label-free concept based multiple instance learning for gigapixel histopathology}.
\newblock \bibinfo{journal}{arXiv preprint arXiv:2501.02922} .
\bibitem[{Sun et~al.(2025b)Sun, Zhang, Si, Zhu, Zhang, Shui, Li, Gong, LYU, Lin and Yang}]{ICLR2025_ebf8764e}
\bibinfo{author}{Sun, Y.}, \bibinfo{author}{Zhang, Y.}, \bibinfo{author}{Si, Y.}, \bibinfo{author}{Zhu, C.}, \bibinfo{author}{Zhang, K.}, \bibinfo{author}{Shui, Z.}, \bibinfo{author}{Li, J.}, \bibinfo{author}{Gong, X.}, \bibinfo{author}{LYU, X.}, \bibinfo{author}{Lin, T.}, \bibinfo{author}{Yang, L.}, \bibinfo{year}{2025}b.
\newblock \bibinfo{title}{Pathgen-1.6m: 1.6 million pathology image-text pairs generation through multi-agent collaboration}, in: \bibinfo{editor}{Yue, Y.}, \bibinfo{editor}{Garg, A.}, \bibinfo{editor}{Peng, N.}, \bibinfo{editor}{Sha, F.}, \bibinfo{editor}{Yu, R.} (Eds.), \bibinfo{booktitle}{International Conference on Learning Representations}, pp. \bibinfo{pages}{94611--94653}.
\newblock \URLprefix \url{https://proceedings.iclr.cc/paper_files/paper/2025/file/ebf8764ecf0688cdd9fe1e5a9c525d0d-Paper-Conference.pdf}.
\bibitem[{Tokez et~al.(2020)Tokez, Hollestein, Louwman, Nijsten and Wakkee}]{tokez2020incidence}
\bibinfo{author}{Tokez, S.}, \bibinfo{author}{Hollestein, L.}, \bibinfo{author}{Louwman, M.}, \bibinfo{author}{Nijsten, T.}, \bibinfo{author}{Wakkee, M.}, \bibinfo{year}{2020}.
\newblock \bibinfo{title}{Incidence of multiple vs first cutaneous squamous cell carcinoma on a nationwide scale and estimation of future incidences of cutaneous squamous cell carcinoma}.
\newblock \bibinfo{journal}{JAMA dermatology} \bibinfo{volume}{156}, \bibinfo{pages}{1300--1306}.
\bibitem[{Tomczak et~al.(2015)Tomczak, Czerwi{\'n}ska and Wiznerowicz}]{tomczak2015review}
\bibinfo{author}{Tomczak, K.}, \bibinfo{author}{Czerwi{\'n}ska, P.}, \bibinfo{author}{Wiznerowicz, M.}, \bibinfo{year}{2015}.
\newblock \bibinfo{title}{Review the cancer genome atlas (tcga): an immeasurable source of knowledge}.
\newblock \bibinfo{journal}{Contemporary Oncology/Wsp{\'o}{\l}czesna Onkologia} \bibinfo{volume}{2015}, \bibinfo{pages}{68--77}.
\bibitem[{Tourniaire et~al.(2024)Tourniaire, Ilie, Mazi{\`e}res, Vigier, Ghiringhelli, Piton, Sabourin, Bibeau, Hofman, Ayache et~al.}]{tourniaire2024whario}
\bibinfo{author}{Tourniaire, P.}, \bibinfo{author}{Ilie, M.}, \bibinfo{author}{Mazi{\`e}res, J.}, \bibinfo{author}{Vigier, A.}, \bibinfo{author}{Ghiringhelli, F.}, \bibinfo{author}{Piton, N.}, \bibinfo{author}{Sabourin, J.C.}, \bibinfo{author}{Bibeau, F.}, \bibinfo{author}{Hofman, P.}, \bibinfo{author}{Ayache, N.}, et~al., \bibinfo{year}{2024}.
\newblock \bibinfo{title}{Whario: whole-slide-image-based survival analysis for patients treated with immunotherapy}.
\newblock \bibinfo{journal}{Journal of Medical Imaging} \bibinfo{volume}{11}, \bibinfo{pages}{037502--037502}.
\bibitem[{Wu et~al.(2024)Wu, Ke, Jiang, Wu, Kong and Shao}]{wu2024leveraging}
\bibinfo{author}{Wu, J.}, \bibinfo{author}{Ke, X.}, \bibinfo{author}{Jiang, X.}, \bibinfo{author}{Wu, H.}, \bibinfo{author}{Kong, Y.}, \bibinfo{author}{Shao, L.}, \bibinfo{year}{2024}.
\newblock \bibinfo{title}{Leveraging tumor heterogeneity: Heterogeneous graph representation learning for cancer survival prediction in whole slide images}.
\newblock \bibinfo{journal}{Advances in Neural Information Processing Systems} \bibinfo{volume}{37}, \bibinfo{pages}{64312--64337}.
\bibitem[{Xiang et~al.(2023)Xiang, Wang, Wang, Zhang, Yang, Yang, Han and Liu}]{xiang2023automatic}
\bibinfo{author}{Xiang, J.}, \bibinfo{author}{Wang, X.}, \bibinfo{author}{Wang, X.}, \bibinfo{author}{Zhang, J.}, \bibinfo{author}{Yang, S.}, \bibinfo{author}{Yang, W.}, \bibinfo{author}{Han, X.}, \bibinfo{author}{Liu, Y.}, \bibinfo{year}{2023}.
\newblock \bibinfo{title}{Automatic diagnosis and grading of prostate cancer with weakly supervised learning on whole slide images}.
\newblock \bibinfo{journal}{Computers in Biology and Medicine} \bibinfo{volume}{152}, \bibinfo{pages}{106340}.
\bibitem[{Xiong et~al.(2021)Xiong, Zeng, Chakraborty, Tan, Fung, Li and Singh}]{xiong2021nystromformer}
\bibinfo{author}{Xiong, Y.}, \bibinfo{author}{Zeng, Z.}, \bibinfo{author}{Chakraborty, R.}, \bibinfo{author}{Tan, M.}, \bibinfo{author}{Fung, G.}, \bibinfo{author}{Li, Y.}, \bibinfo{author}{Singh, V.}, \bibinfo{year}{2021}.
\newblock \bibinfo{title}{Nystr{\"o}mformer: A nystr{\"o}m-based algorithm for approximating self-attention}, in: \bibinfo{booktitle}{Proceedings of the AAAI conference on artificial intelligence}, pp. \bibinfo{pages}{14138--14148}.
\bibitem[{Yacob et~al.(2023)Yacob, Siarov, Villiamsson, Suvilehto, Sj{\"o}blom, Kjellberg and Neittaanm{\"a}ki}]{yacob2023weakly}
\bibinfo{author}{Yacob, F.}, \bibinfo{author}{Siarov, J.}, \bibinfo{author}{Villiamsson, K.}, \bibinfo{author}{Suvilehto, J.T.}, \bibinfo{author}{Sj{\"o}blom, L.}, \bibinfo{author}{Kjellberg, M.}, \bibinfo{author}{Neittaanm{\"a}ki, N.}, \bibinfo{year}{2023}.
\newblock \bibinfo{title}{Weakly supervised detection and classification of basal cell carcinoma using graph-transformer on whole slide images}.
\newblock \bibinfo{journal}{Scientific Reports} \bibinfo{volume}{13}, \bibinfo{pages}{7555}.
\bibitem[{Zhang et~al.(2021)Zhang, Ma, Van~Arnam, Gupta, Saltz, Vakalopoulou and Samaras}]{zhang2021joint}
\bibinfo{author}{Zhang, J.}, \bibinfo{author}{Ma, K.}, \bibinfo{author}{Van~Arnam, J.}, \bibinfo{author}{Gupta, R.}, \bibinfo{author}{Saltz, J.}, \bibinfo{author}{Vakalopoulou, M.}, \bibinfo{author}{Samaras, D.}, \bibinfo{year}{2021}.
\newblock \bibinfo{title}{A joint spatial and magnification based attention framework for large scale histopathology classification}, in: \bibinfo{booktitle}{Proceedings of the IEEE/CVF Conference on Computer Vision and Pattern Recognition}, pp. \bibinfo{pages}{3776--3784}.
\bibitem[{Zhang et~al.(2022)Zhang, Zhang, Zhao, Chen, Arik and Pfister}]{zhang2022nested}
\bibinfo{author}{Zhang, Z.}, \bibinfo{author}{Zhang, H.}, \bibinfo{author}{Zhao, L.}, \bibinfo{author}{Chen, T.}, \bibinfo{author}{Arik, S.{\"O}.}, \bibinfo{author}{Pfister, T.}, \bibinfo{year}{2022}.
\newblock \bibinfo{title}{Nested hierarchical transformer: Towards accurate, data-efficient and interpretable visual understanding}, in: \bibinfo{booktitle}{Proceedings of the AAAI Conference on Artificial Intelligence}, pp. \bibinfo{pages}{3417--3425}.
\bibitem[{Zheng et~al.(2022)Zheng, Gindra, Green, Burks, Betke, Beane and Kolachalama}]{zheng2022graph}
\bibinfo{author}{Zheng, Y.}, \bibinfo{author}{Gindra, R.H.}, \bibinfo{author}{Green, E.J.}, \bibinfo{author}{Burks, E.J.}, \bibinfo{author}{Betke, M.}, \bibinfo{author}{Beane, J.E.}, \bibinfo{author}{Kolachalama, V.B.}, \bibinfo{year}{2022}.
\newblock \bibinfo{title}{A graph-transformer for whole slide image classification}.
\newblock \bibinfo{journal}{IEEE transactions on medical imaging} \bibinfo{volume}{41}, \bibinfo{pages}{3003--3015}.

\end{thebibliography}

\newpage
\appendix
\section{Appendix}
\subsection{Model hyperparameter configuration}
We summarize the RACR-MIL hyperparameters for the three SCC datasets in Table \ref{tab:hyperparameters}. We use the cosine softmax classifier for skin SCC, lung SCC and head and neck SCC. We used 5 nearest neighbors for the latent graph and 8 nearest neighbors for the spatial graph. A single self-attention layer with two heads, one for each graph, was employed. The temperature parameter for the cosine-softmax classifier was optimized separately for each dataset and feature extractor. All models were trained using the AdamW optimizer (\cite{loshchilov2018decoupled}) with a learning rate of $1 \times 10^{-4}$ and a batch size of 16. We used $L_2$ regularization with weight of $1 \times 10^{-3}$ and introduced the ranking losses after a 10 epoch warm-up phase. Specifically, we set $\lambda_1 = \lambda_2 = 0$ during warmup so that the model can first learn stable patch-level probabilities, reducing the influence of noisy pseudo-grades assigned to patches under weak supervision. 
\begin{table}[h!]
\centering
\caption{Hyperparameter values for models trained using SCC-DINO, PathGen-CLIP, and CONCHv1.5 feature extractors}
\label{tab:hyperparameters}
\resizebox{\textwidth}{!}{
\begin{tabular}{p{4.2cm} p{4.2cm} p{4.2cm} p{4.2cm}}
\hline
\textbf{Hyperparameter} & \textbf{SCC-DINO} & \textbf{PathGen-CLIP} & \textbf{CONCHv1.5} \\
\hline

K - Latent Graph 
& 5 
& 5
& 5 \\

K - Spatial Graph 
& 8 
& 8
& 8 \\

\# Att-GNN layers 
& 1 
& 1
& 1 \\

$\alpha$ (Diffusion)
& 0.25
& 0.25
& 0.25 \\

Hidden Dimensions 
& 64 
& 128
& 128 \\

Dropout
& 0.5 
& 0.5
& 0.5 \\

$t$ (cosine classifier)
& 0.1 (CSCC), 0.1 (HNSC), 0.1 (LSCC)
& 0.1 (CSCC), 0.1 (HNSC), 0.3 (LSCC)
& 0.1 (CSCC), 0.3 (LSCC) \\

Weight Decay
& $10^{-3}$ 
& $10^{-3}$
& $10^{-3}$ \\

Batch Size
& 16 
& 16
& 16 \\

Learning Rate 
& $10^{-4}$
& $10^{-4}$
& $10^{-4}$ \\

Warmup Epochs (Ranking)
& 10 
& 10
& 10 \\

Early Stopping Epochs 
& 9 
& 9
& 9 \\

Total Epochs
& 60 (CSCC), 100 (HNSC), 100 (LSCC)
& 60 (CSCC), 100 (HNSC), 100 (LSCC)
& 60 (CSCC), 100 (LSCC) \\

Model Selection Criterion 
& Validation F1-score 
& Validation F1-score
& Validation F1-score \\

\hline
\end{tabular}
}
\end{table}

\subsection{Precision-recall curves and confusion matrices for class-wise performance}

\begin{figure*}[h!]
\includegraphics[width=\textwidth]{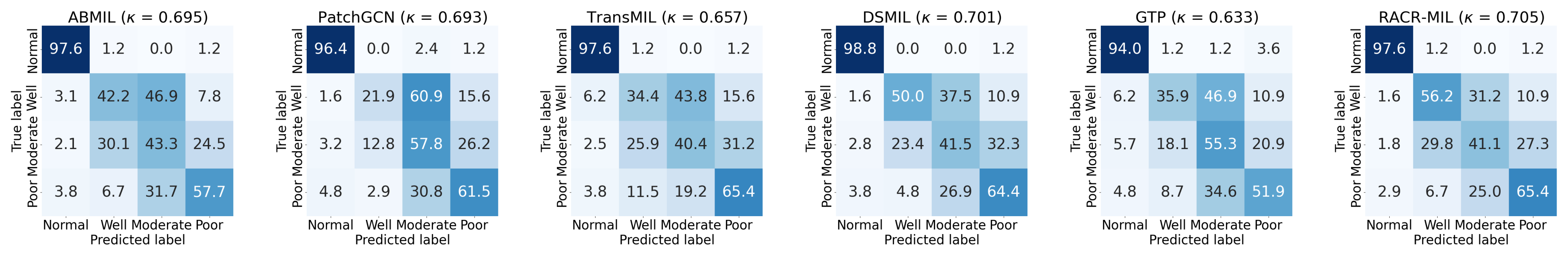}
 \caption{Confusion matrices for head and neck SCC: RACR-MIL achieves balanced performance across grades with highest $\kappa$ score across contextual approaches.}
 \label{cf_hnsc}
\end{figure*}

\begin{figure*}[h!]
\includegraphics[width=\textwidth]{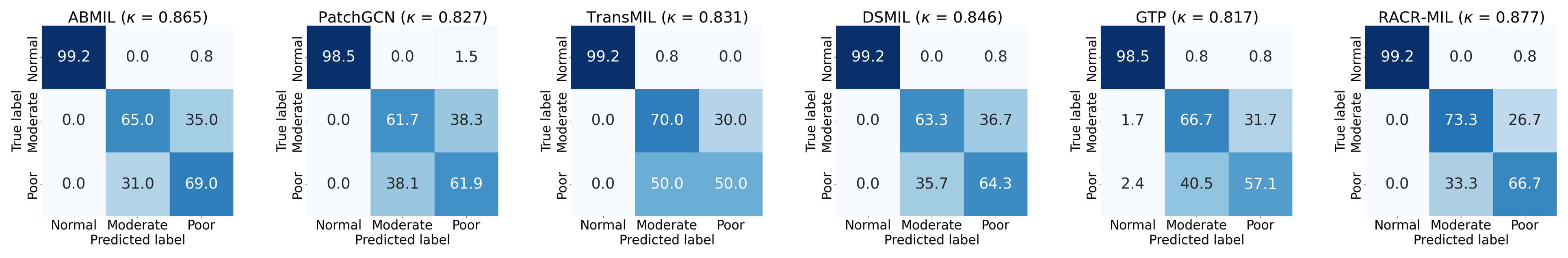}
 \caption{Confusion matrices for lung SCC: RACR-MIL achieves highest $\kappa$ score outperforming all approaches on Poorly-differentiated tumors.}
 \label{cf_lscc}
\end{figure*}

\begin{figure*}[h!]
\includegraphics[width=\textwidth]{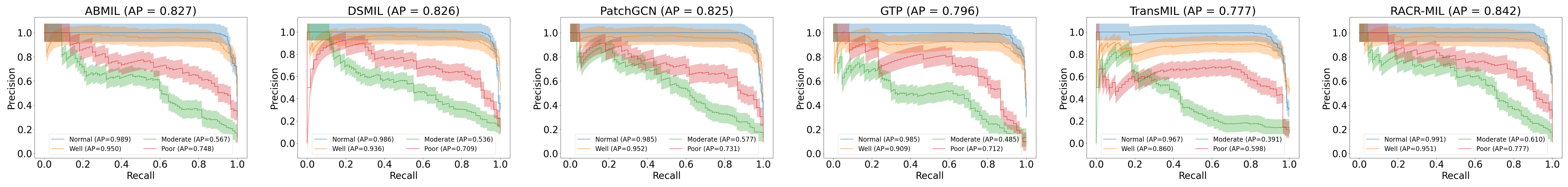}
 \caption{Precision-recall curves for skin SCC: RACR-MIL achieves the highest average precision across all grades, highlighting its robustness.}
 \label{pr_cscc}
\end{figure*}


\begin{figure*}[h!]
\includegraphics[width=\textwidth]{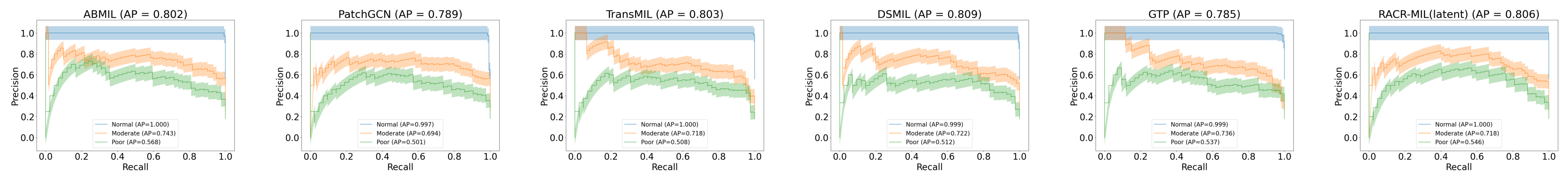}
 \caption{Precision-recall curves for lung SCC grading: RACR-MIL's latent graph variant achieves the second highest average precision across all classes.}
 \label{pr_lscc}
\end{figure*}

\begin{figure*}[h!]
\centering
\includegraphics[width=0.6\textwidth]{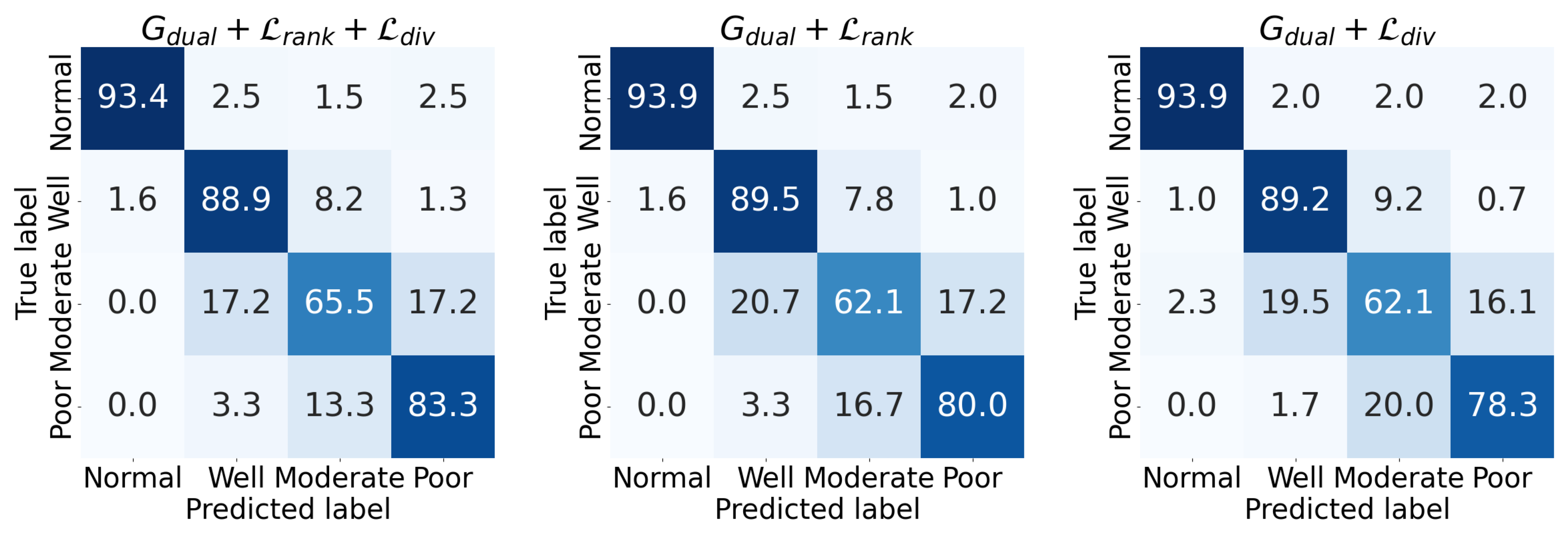}
 \caption{Confusion matrices for RACR-MIL variants: Incorporating the diversity loss and ranking losses with dual graph formulation leads to the highest accuracy on higher-risk grades.}
 \label{cf_racr}
\end{figure*}

\clearpage
\subsection{Evaluation using CONCHv1.5 features}

\begin{table*}[!ht]
\centering
\small
\setlength{\tabcolsep}{4pt}
\caption{Comparison of RACR-MIL with state-of-the-art MIL approaches using CONCHv1.5 feature encoder. Mean and standard deviation across five folds are reported. \textbf{Bold} and \underline{Underline} indicate the best and second best performance.}
\label{classification_result_conch}
\resizebox{\textwidth}{!}{
\begin{tabular}{lccccc|ccccc}
\toprule
\multirow{2}{*}{\textbf{Approach}} &
\multicolumn{5}{c}{\textbf{Skin SCC}} &
\multicolumn{5}{c}{\textbf{Lung SCC}} \\
\cmidrule(lr){2-6}\cmidrule(lr){7-11}
& F1 & Precision & Recall & AUC & MCC
& F1 & Precision & Recall & AUC & MCC \\
\midrule
\multicolumn{11}{l}{\textbf{Non-contextual}} \\
\midrule

Max Pooling
& 0.738 \tiny 0.035
& 0.740 \tiny 0.027	
& 0.760 \tiny 0.044	
& 0.949 \tiny 0.006	
& 0.739 \tiny 0.056

& 0.757 {\tiny 0.102}
& 0.779 {\tiny 0.097}
& 0.776 {\tiny 0.100}
& 0.935 {\tiny 0.024}
& 0.748 {\tiny 0.105}
\\
\midrule

Mean Pooling
& 0.735 {\tiny 0.041}
& 0.741 {\tiny 0.034}
& 0.742 {\tiny 0.054}
& 0.939 {\tiny 0.017}
& 0.724 {\tiny 0.057}
& 0.735 {\tiny 0.081}
& 0.750 {\tiny 0.084}
& 0.756 {\tiny 0.077}
& \underline{0.947} {\tiny 0.020}
& 0.714 {\tiny 0.091}
\\
\midrule

ABMIL
& 0.767 {\tiny 0.018}
& 0.763 {\tiny 0.022}
& 0.791 {\tiny 0.029}
& 0.952 {\tiny 0.014}
& 0.759 {\tiny 0.034}
& 0.764 {\tiny 0.092}
& 0.778 {\tiny 0.094}
& 0.778 {\tiny 0.081}
& 0.943 {\tiny 0.021}
& \underline{0.751} {\tiny 0.091}
\\
\midrule

GABMIL
& 0.777 \tiny 0.021	
& 0.773 \tiny 0.019	
& 0.794 \tiny 0.031	
& \underline{0.956} \tiny 0.014
& 0.768 \tiny 0.035

& \textbf{0.780} {\tiny 0.075}
& \textbf{0.787} {\tiny 0.074}
& \underline{0.790} {\tiny 0.065}
& 0.946 {\tiny 0.020}
& \textbf{0.766} {\tiny 0.073}
\\
\midrule

CLAM-MB
& 0.768 {\tiny 0.028}
& 0.764 {\tiny 0.020}
& 0.790 {\tiny 0.045}
& \underline{0.956} {\tiny 0.015}
& 0.764 {\tiny 0.046}
& 0.761 {\tiny 0.066}
& 0.779 {\tiny 0.079}
& 0.778 {\tiny 0.062}
& 0.944 {\tiny 0.025}
& 0.747 {\tiny 0.073}
\\
\midrule

\multicolumn{11}{l}{\textbf{Contextual}} \\
\midrule

DSMIL
& 0.769 {\tiny 0.040}
& 0.763 {\tiny 0.037}
& 0.790 {\tiny 0.051}
& 0.954 {\tiny 0.018}
& 0.756 {\tiny 0.056}
& 0.765 {\tiny 0.076}
& 0.774 {\tiny 0.075}
& 0.781 {\tiny 0.079}
& \textbf{0.953} {\tiny 0.024}
& 0.749 {\tiny 0.091}
\\
\midrule

PatchGCN
& 0.760 {\tiny 0.033}
& 0.755 {\tiny 0.029}
& 0.789 {\tiny 0.044}
& 0.949 {\tiny 0.017}
& 0.749 {\tiny 0.050}
& 0.749 {\tiny 0.104}
& 0.763 {\tiny 0.109}
& 0.762 {\tiny 0.099}
& 0.941 {\tiny 0.028}
& 0.739 {\tiny 0.112}
\\
\midrule

HEAT
& 0.757 {\tiny 0.036}
& 0.759 {\tiny 0.033}
& \underline{0.797} {\tiny 0.051}
& 0.946 {\tiny 0.018}
& 0.724 {\tiny 0.048}
& 0.712 {\tiny 0.111}
& 0.738 {\tiny 0.086}
& 0.739 {\tiny 0.092}
& 0.934 {\tiny 0.034}
& 0.695 {\tiny 0.110}
\\
\midrule

WiKG
& 0.734 {\tiny 0.019}
& 0.741 {\tiny 0.007}
& 0.740 {\tiny 0.039}
& 0.938 {\tiny 0.009}
& 0.726 {\tiny 0.034}
& 0.716 {\tiny 0.065}
& 0.730 {\tiny 0.059}
& 0.730 {\tiny 0.058}
& 0.942 {\tiny 0.017}
& 0.698 {\tiny 0.065}
\\
\midrule

CAMIL
& \underline{0.781} {\tiny 0.050}
& 0.781 {\tiny 0.044}
& \underline{0.797} {\tiny 0.066}
& \textbf{0.957} {\tiny 0.019}
& \underline{0.776} {\tiny 0.057}
& 0.741 {\tiny 0.083}
& 0.754 {\tiny 0.098}
& 0.747 {\tiny 0.079}
& 0.943 {\tiny 0.024}
& 0.730 {\tiny 0.086}
\\
\midrule

TransMIL
& 0.766 {\tiny 0.040}
& 0.764 {\tiny 0.036}
& 0.789 {\tiny 0.051}
& 0.950 {\tiny 0.016}
& 0.757 {\tiny 0.055}
& 0.749 {\tiny 0.107}
& 0.766 {\tiny 0.093}
& 0.766 {\tiny 0.091}
& 0.945 {\tiny 0.025}
& 0.747 {\tiny 0.105}
\\
\midrule

LongMIL
& 0.745 \tiny 0.042
& 0.752 \tiny 0.047
& 0.754 \tiny 0.063
& 0.950 \tiny 0.014	
& 0.742 \tiny 0.042
& 0.715 \tiny 0.052 
& 0.726 \tiny 0.052 
& 0.732 \tiny 0.049
& 0.928 \tiny 0.031	
& 0.692 \tiny 0.060
\\
\midrule

GTP
& 0.758 {\tiny 0.050}
& \underline{0.785} {\tiny 0.045}
& 0.756 {\tiny 0.056}
& 0.949 {\tiny 0.019}
& 0.763 {\tiny 0.046}
& 0.712 {\tiny 0.104}
& 0.738 {\tiny 0.088}
& 0.734 {\tiny 0.084}
& 0.935 {\tiny 0.021}
& 0.687 {\tiny 0.098}
\\
\midrule

RACR-MIL
& \textbf{0.785} {\tiny 0.035}
& \textbf{0.787} {\tiny 0.033}
& \textbf{0.802} {\tiny 0.050}
& \underline{0.956} {\tiny 0.015}
& \textbf{0.784} {\tiny 0.053}
& \underline{0.775} {\tiny 0.048}
& \underline{0.781} {\tiny 0.043}
& \textbf{0.797} {\tiny 0.050}
& 0.932 {\tiny 0.029}
& 0.748 {\tiny 0.048}
\\

\bottomrule
\end{tabular}
}
\end{table*}

\begin{table*}[!h]
\centering
\small
\setlength{\tabcolsep}{4pt}
\caption{Impact of different patch-selection probability thresholds ($\tau$) on RACR-MIL performance using the PathGen-CLIP encoder. Mean and standard deviation across five folds are reported.}
\label{classification_result_threshold}
\resizebox{\textwidth}{!}{
\begin{tabular}{lccccc|ccccc}
\toprule
\multirow{2}{*}{\textbf{Threshold   }} &
\multicolumn{5}{c}{\textbf{Skin SCC}} &
\multicolumn{5}{c}{\textbf{Lung SCC}} \\
\cmidrule(lr){2-6}\cmidrule(lr){7-11}
& F1 & Precision & Recall & AUC & MCC
& F1 & Precision & Recall & AUC & MCC \\
\midrule


$\tau = 0.6$
& \textbf{0.805} {\tiny 0.025}
& \textbf{0.801} {\tiny 0.019}
& \textbf{0.824} {\tiny 0.043}
& 0.955 {\tiny 0.011}
& \textbf{0.797} {\tiny 0.027}
& \textbf{0.786} \tiny 0.049	
& \textbf{0.796} \tiny 0.042 
& \textbf{0.805} \tiny 0.047 
& 0.938 \tiny 0.024 
& \textbf{0.775} \tiny 0.045
\\
\midrule

$\tau = 0.7$
& 0.795 {\tiny 0.020}
& 0.792 {\tiny 0.013}
& 0.809 {\tiny 0.042}
& 0.956 {\tiny 0.013}
& 0.785 {\tiny 0.026}
& 0.778 \tiny 0.057	
& 0.789 \tiny 0.051	
& 0.796 \tiny 0.060	
& \textbf{0.939} \tiny 0.025	
& 0.768 \tiny 0.054

\\
\midrule

$\tau = 0.8$
& 0.798 {\tiny 0.010}
& 0.790 {\tiny 0.011}
& 0.820 {\tiny 0.031}
& \textbf{0.957} {\tiny 0.013}
& 0.784 {\tiny 0.016}
& 0.775 \tiny 0.059	
& 0.783 \tiny 0.055	
& 0.794 \tiny 0.062	
& 0.938 \tiny 0.025	
& 0.762 \tiny 0.060
\\

\bottomrule
\end{tabular}
}
\end{table*}

\clearpage

\subsection{Comparison of model predictions and pathologist-assigned grades in the Tumor Localization study}

\begin{table*}[h!]
\centering
\scriptsize
\setlength{\tabcolsep}{4pt}
\renewcommand{\arraystretch}{1.1}
\caption{Comparison of model predictions with three pathologists across 19 WSIs used in the localization study. Cyan highlights denote cases where the models fail to predict the highest-grade tumor identified by any of the pathologists. Red text indicates cases where model predictions disagree with the majority grade determined from the pathologists’ diagnoses.}
\resizebox{\textwidth}{!}{
\begin{tabular}{c c c c c c c c c c}
\hline
\textbf{Case \#} &
\textbf{\makecell{Training\\labels}} &
\textbf{\makecell{Proportion of \\area labeled}} &
\textbf{\makecell{Number of\\patches}} &
\textbf{Pathologist \#1} &
\textbf{Pathologist \#2} &
\textbf{Pathologist \#3} &
\textbf{RACR-MIL} &
\textbf{ABMIL} &
\textbf{PatchGCN} \\
\hline
1 & Mod  & 20\% & 215  & Mod & Mod & Mod & Mod & Mod & Mod \\
2 & Well & 57\% & 267  & Well-Mod & Well & Well & \cellcolor{cyan!20}Well & {\color{red}Mod} & {\color{red}Mod} \\
3 & Mod  & 24\% & 122  & Poor & Mod-Poor & Poor & Poor & {\color{red}Mod} & {\color{red}Mod} \\
4 & Mod  & 54\% & 1133 & Well-Mod & Well-Mod & Mod & Mod & Mod & Mod \\
5 & Poor & 55\% & 858  & Poor & Poor & Poor & {\color{red}Mod} & \color{red}{Mod} & Poor \\
6 & Mod  & 26\% & 807  & Mod-Poor & Mod-Poor & Poor & Poor & Poor & \color{red}{Mod} \\
7 & Poor & 33\% & 2499 & Well-Mod & Well-Mod & Mod & Mod & Mod & Mod \\
8 & Poor & 86\% & 3698 & Mod & Mod-Poor & Mod-Poor & Mod & Mod & Mod \\
9 & Well & 43\% & 747  & Well-Mod & Mod-Poor & Mod & {\color{red}Poor} & {\color{red}Poor} & {\color{red}Poor} \\
10 & Well & 43\% & 235  & Mod & Well-Mod & Mod & Mod & Mod & Mod \\
11 & Well & 6\%  & 963  & Well & Well-Mod & Well-Mod & \cellcolor{cyan!20}Well & \cellcolor{cyan!20}Well & \cellcolor{cyan!20}Well \\
12 & Mod  & 10\% & 1371 & Mod-Poor & Mod & Mod & \cellcolor{cyan!20}Mod & \cellcolor{cyan!20}Mod & \cellcolor{cyan!20}Mod \\
13 & Mod  & 65\% & 1514 & Mod & Mod & Mod & Mod & Mod & Mod \\
14 & Mod  & 36\% & 3668 & Well-Mod & Well & Mod & Mod & Mod & Mod \\
15 & Well & 13\% & 2267 & Well-Mod & Well & Well & \cellcolor{cyan!20}Well & \cellcolor{cyan!20} Well & {\color{red}Mod} \\
16 & Poor & 22\% & 3310 & Poor & Poor & Poor & Poor & Poor & Poor \\
17 & Mod  & 12\% & 2167 & Poor & Mod-Poor & Mod-Poor & Poor & Poor & Poor \\
18 & Poor & 9\%  & 3430 & Mod-Poor & Mod-Poor & Mod & {\color{red}Poor} & \cellcolor{cyan!20}Mod & \cellcolor{cyan!20}Mod \\
19 & Poor & 19\% & 3595 & Poor & Mod-Poor & Poor & Poor & Poor & Poor \\
\hline
\end{tabular}
}

\label{tab:case_comparison}
\end{table*}

\subsection{UMAP plot highlighting the batch effect in combined HNSC cohort (SCC-DINO feature extractor)}
\begin{figure*}[h!]
\includegraphics[width=\textwidth]{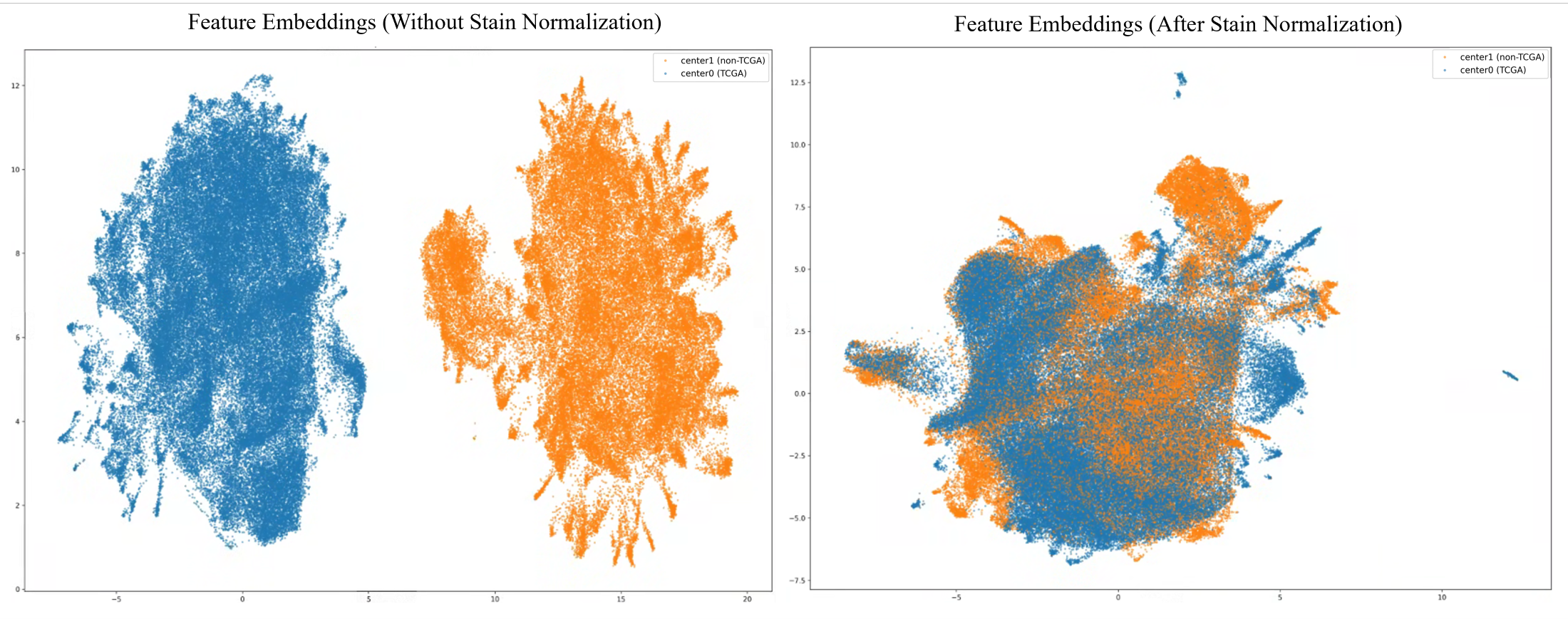}
 \caption{UMAP visualization of patch features from the combined HNSC cohort (CPTAC and TCGA) before and after stain normalization}
 \label{fig:umap_hnsc}
\end{figure*}

\clearpage
\subsection{Example cases highlighting tissue variability in skin SCC dataset}

\begin{figure*}[h!]
    \centering
    
    \begin{subfigure}{0.48\textwidth}
        \centering
        \includegraphics[width=\linewidth]{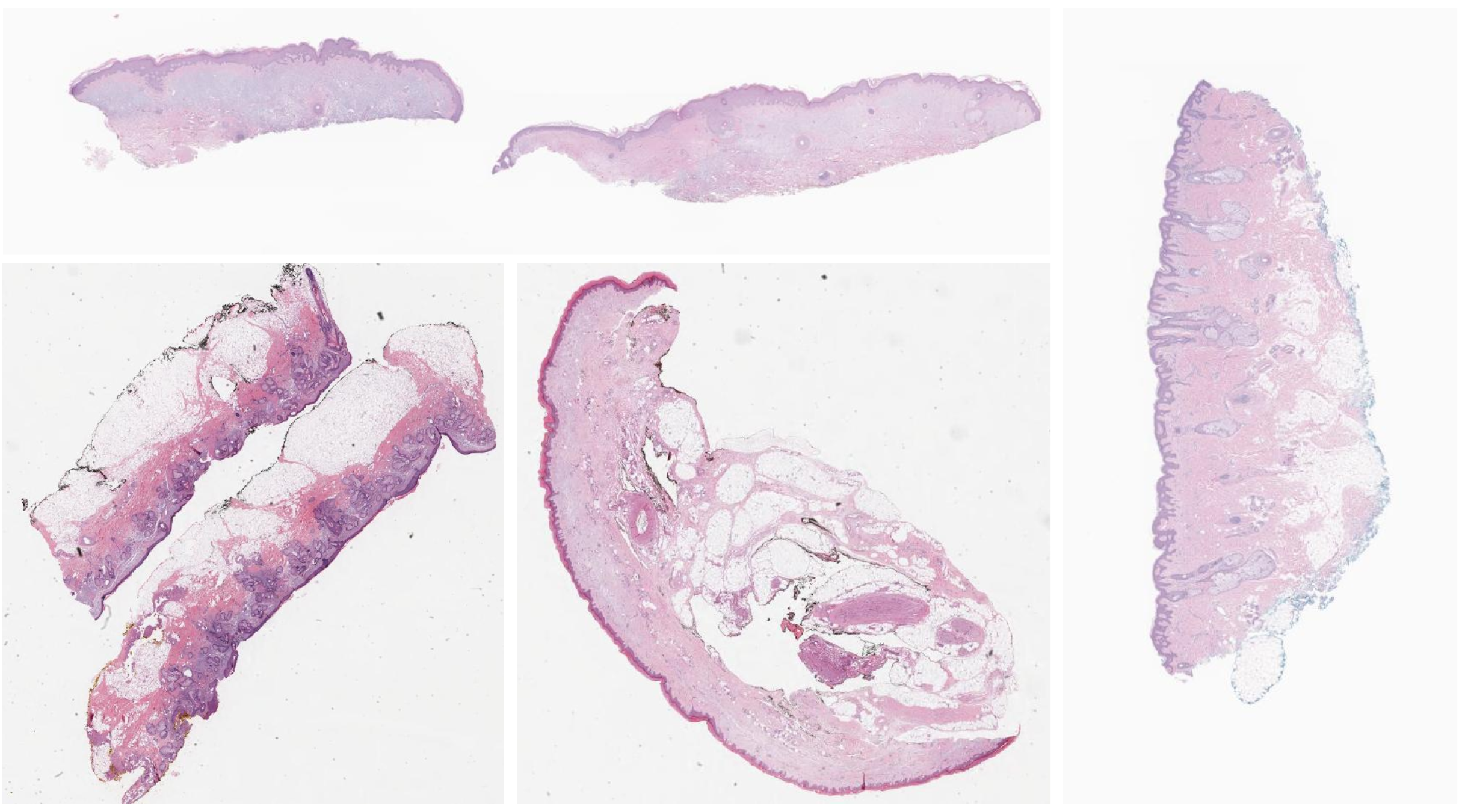}
        \caption{Normal skin tissue}
        \label{fig:normal_cscc}
    \end{subfigure}
    \hfill
    \begin{subfigure}{0.48\textwidth}
        \centering
        \includegraphics[width=\linewidth]{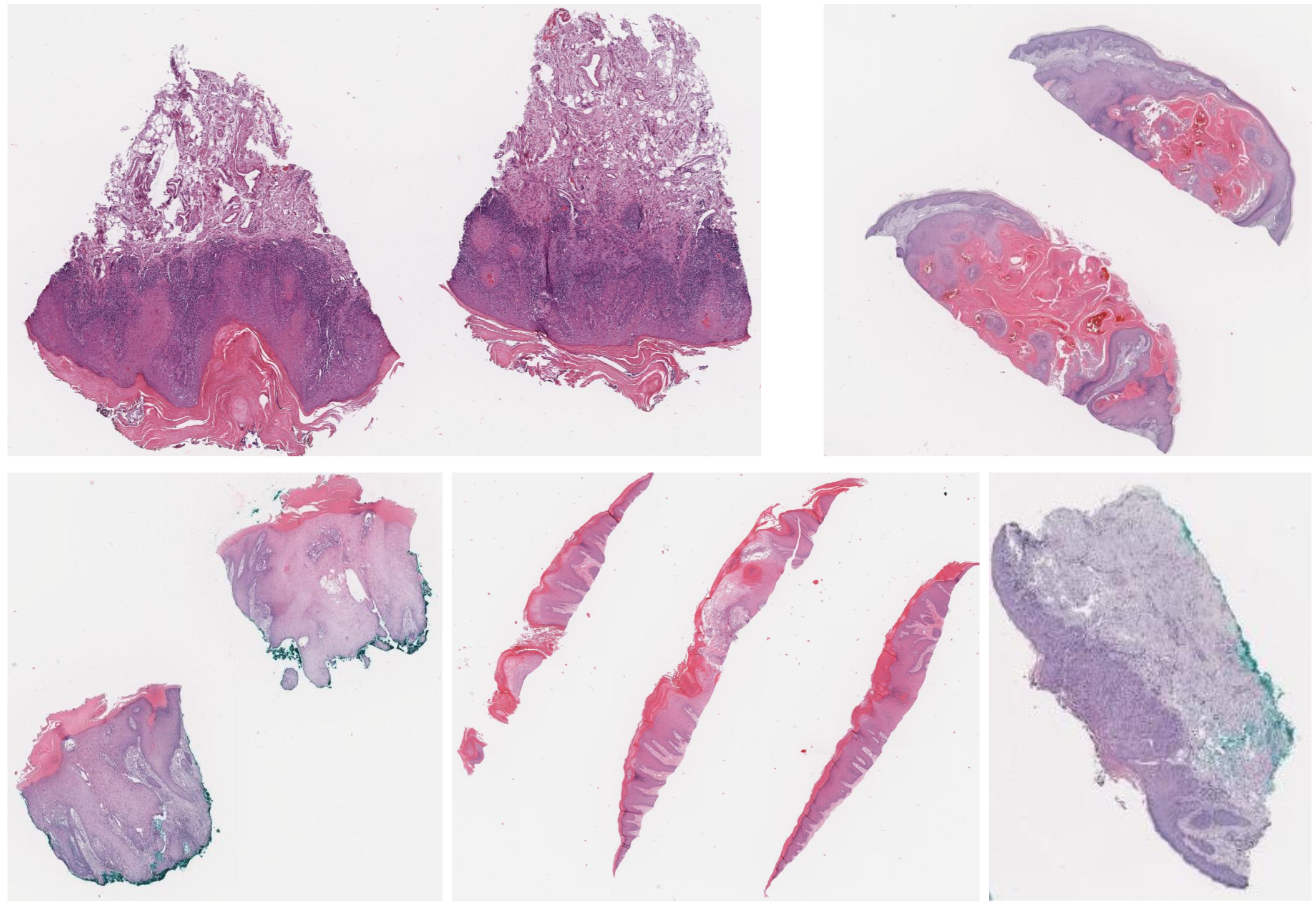}
        \caption{Well-differentiated SCC}
        \label{fig:well_cscc}
    \end{subfigure}
    
    \vspace{0.5cm}
    
    \begin{subfigure}{0.48\textwidth}
        \centering
        \includegraphics[width=\linewidth]{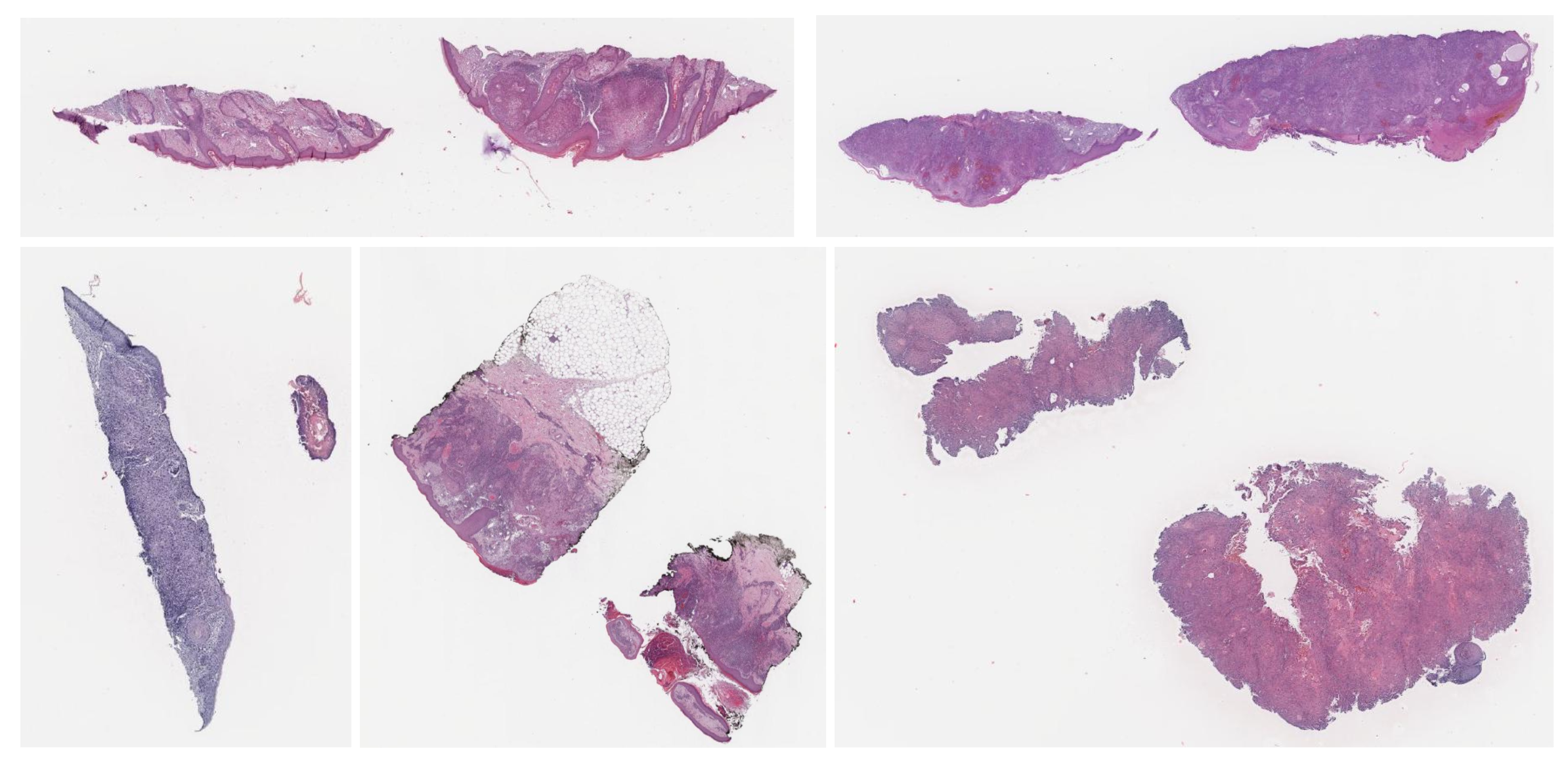}
        \caption{Moderately-differentiated SCC}
        \label{fig:moderate_cscc}
    \end{subfigure}
    \hfill
    \begin{subfigure}{0.48\textwidth}
        \centering
        \includegraphics[width=\linewidth]{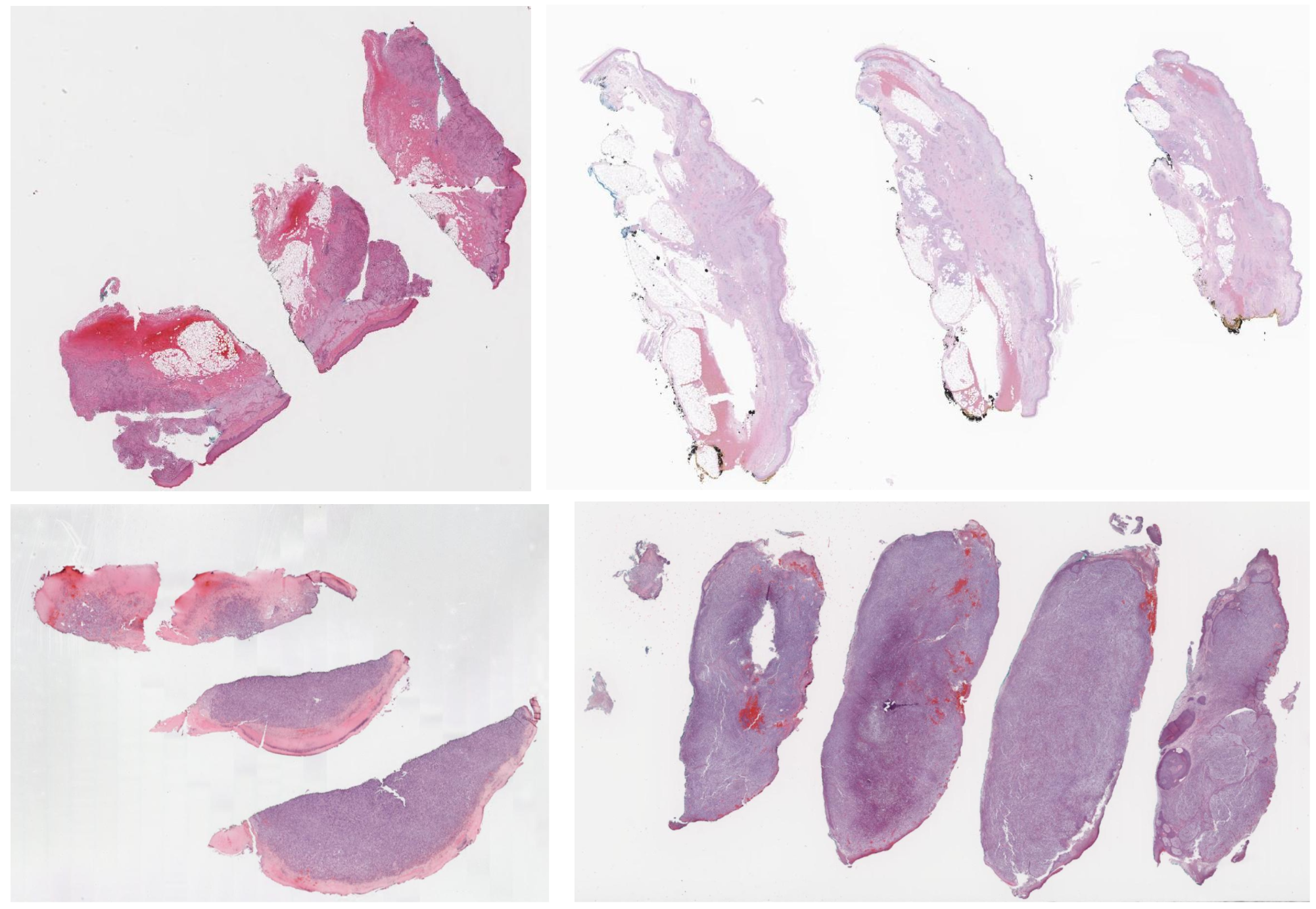}
        \caption{Poorly-differentiated SCC}
        \label{fig:poor_cscc}
    \end{subfigure}
    
    \caption{Representative cases from the skin SCC dataset highlighting real-world variability in tissue morphology, staining characteristics, and sampling artifacts across grade classes.}
    \label{fig:cscc_variability}
\end{figure*}


\end{document}